\documentclass[11pt]{article}

\usepackage{acl}

\usepackage{times}
\usepackage{latexsym}

\usepackage[T1]{fontenc}

\usepackage[utf8]{inputenc}

\usepackage{microtype}

\usepackage{tipa} %

\usepackage{inconsolata}

\usepackage{appendix} %

\usepackage{lscape} %
\usepackage{hyperref} %
\usepackage{graphicx}
\usepackage{adjustbox}    %
\usepackage{multirow}
\usepackage{bbm} %
\usepackage{amsmath} %
\usepackage{csquotes} %
\usepackage{numprint} %
\usepackage{longtable}
\usepackage{booktabs} %
\usepackage{arydshln} %

\newcolumntype{L}{>{\centering\arraybackslash}m{2cm}} %
\newcolumntype{S}{>{\arraybackslash}m{6cm}} %
\newcolumntype{R}{>{\centering\arraybackslash}m{0.6cm}} %
\newcolumntype{T}{>{\centering\arraybackslash}m{0.57cm}} %
\newcolumntype{O}{>{\centering\arraybackslash}m{0.25cm}} %
\newcolumntype{E}{>{\centering\arraybackslash}m{1.3cm}} %
\newcolumntype{N}{>{\centering\arraybackslash}m{2.6cm}} %

\usepackage{algorithm}
\usepackage{algpseudocode}

\makeatletter
\newcounter{phase}[algorithm]
\newlength{\phaserulewidth}
\newcommand{\setphaserulewidth}{\setlength{\phaserulewidth}}

\makeatother

\setphaserulewidth{.7pt}

\title{Interplay of Machine Translation, Diacritics, and Diacritization}

\author{Wei-Rui Chen$^{\lambda}$ ~~~Ife Adebara$^{\lambda}$ ~~~Muhammad Abdul-Mageed$^{\lambda,\gamma,\psi}$\\ 
  $^{\lambda}$Deep Learning \& Natural Language Processing Group,
  The University of British Columbia\\  $^{\gamma}$Department of Natural Language Processing \& Department of Machine Learning, MBZUAI\\ $^{\psi}$ Invertible AI\\
  \tt \{weirui.chen,ife.adebara,muhammad.mageed\}@ubc.ca}

\begin{document}
\maketitle
\begin{abstract}
We investigate two research questions: (1) how do machine translation (MT) and diacritization influence the performance of each other in a multi-task learning setting (2) the effect of keeping (vs. removing) diacritics on MT performance. We examine these two questions in both high-resource (HR) and low-resource (LR) settings across 55 different languages (36 African languages and 19 European languages). For (1), results show that diacritization significantly benefits MT in the LR scenario, doubling or even tripling performance for some languages, but harms MT in the HR scenario. We find that MT harms diacritization in LR but benefits significantly in HR for some languages. For (2), MT performance is similar regardless of diacritics being kept or removed. In addition, we propose two classes of metrics to measure the complexity of a diacritical system, finding these metrics to correlate positively with the performance of our diacritization models. Overall, our work provides insights for developing MT and diacritization systems under different data size conditions and may have implications that generalize beyond the 55 languages we investigate.
\end{abstract}

\section{Introduction}\label{sec:intro}

\begin{figure}[t]
\begin{centering}
\includegraphics[scale=0.206]{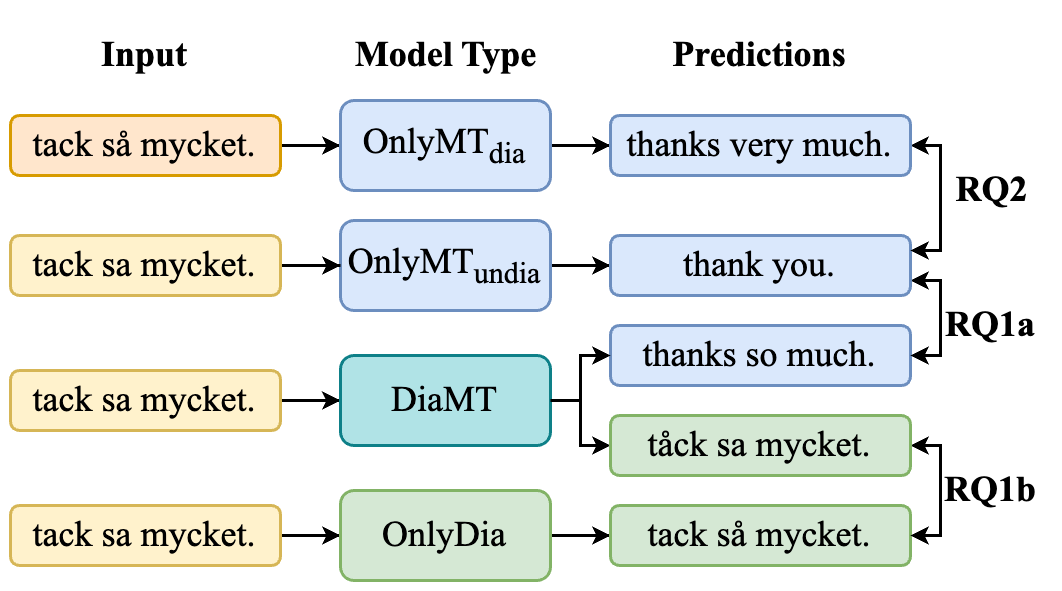}
  \caption{Illustration of our experimental setup, taking a Swedish datapoint `tack s\r{a}
 mycket.' (thank you very much.) as an example. To answer our (\textbf{RQ}s), we develop four types of models: three single-task models \textbf{OnlyMT\textsubscript{dia}} (trained to translate with diacritized source),  \textbf{OnlyMT\textsubscript{undia}} (trained to translate with undiacritized source), and \textbf{OnlyDia} (trained to diacritize); and one multi-task model \textbf{DiaMT} (trained to translate and diacritize simultaneously).}
  \label{fig:setup}
 \end{centering}
\end{figure}

Diacritics are symbols added to a letter to modify its meaning, pronunciation, or phonetic value in an orthographic system \cite{doi:10.1080/10888430903034788, ball_2001, wells2000orthographic}. These symbols can have a lexical or grammatical function \cite{janicki2005reconstruction}. In their lexical function, diacritics distinguish one word from another. For instance in Yor\`{u}b\'{a}, diacritics differentiate meanings in words such as: \texttt{\`{o}g\'{u}n} (\textit{a deity}), \texttt{ogun} (\textit{battle}), \texttt{\`{o}g\`{u}n} (\textit{a river}), \texttt{og\'{u}n} (\textit{number 20 / inheritance}). On the other hand, diacritics also serve a grammatical function by distinguishing one grammatical category from another. For example in Iau, diacritics differentiate past and perfect verbs as in: \texttt{b\'a} \textit{(`came')} and \texttt{ba} (\textit{`has come'}) \cite{hyman2016lexical}. Disregarding diacritics in certain tasks could result in the omission of crucial semantic information.

Despite the important role of diacritics, we are not aware of work that investigates their effect on MT across languages. In this paper, we attempt to fill this knowledge gap by studying the interaction between machine translation (MT), diacritics and diacritization. Diacritization is the task of correctly attaching diacritics to characters. For the interplay between MT and diacritics, we test the effect of keeping and removing diacritics on MT. For the interplay of MT and diacritization, we design a multi-task setting that involves both MT and diacritization. The multi-task models learn to translate and attach diacritics to characters simultaneously. Specifically, we raise two main research questions: in a multi-task setting, whether or not, and if so to what extent does diacritization benefit MT (\textbf{RQ1a.}), and MT benefit diacritization (\textbf{RQ1b.}); and in a single-task setting, whether or not, and if so to what extent, does keeping and removing diacritics affect performance of MT systems (\textbf{RQ2.}). An overview of our experimental setup is shown in Figure~\ref{fig:setup}. We also examine how varying training data sizes, hereafter referred to as \textbf{`train sizes'}, impact the model's performance across various languages.

Our contributions can be summarized as follows: \textbf{(1)} We propose a novel approach to enhance the performance of low-resource machine translation by incorporating diacritization as a multi-task training. \textbf{(2)} We illustrate that, in a single-task setting, the choice of either retaining or omitting diacritics generally has minimal impact on machine translation performance. \textbf{(3)} We propose two categories of language-agnostic metrics designed to assess the complexity of the diacritical system in a language and examine their implications on diacritization performance. To the best of our knowledge, this study represents the most comprehensive analysis of the interplay between diacritics and machine translation. Drawing insights from our experimental findings, we offer practical guidelines for researchers and practitioners involved in developing machine translation or diacritization systems.

This paper is organized as follows. Section~\ref{sec:lit_review} is a literature review. Experimental settings are provided in Section~\ref{sec:exp}. Section~\ref{sec:data} presents information of the data and our proposed language-agnostic complexity metrics. In Section~\ref{sec:results}, we present and discuss our results and key findings. We conclude in Section~\ref{sec:conclusion}.

\section{Related Work}\label{sec:lit_review}

We first review existing literature on MT and diacritics, followed by work on diacritization as a standalone task, and finally we discuss the interplay between diacritization and MT. %

\noindent\textbf{MT and Diacritics.}
There are three primary approaches to handling diacritics in MT: diacritics removal, retention, and restoration. The decision to adopt any of these approaches is motivated by various factors. For example, the inconsistent use of diacritics in a dataset has been identified as a key reason to remove them~\cite{DBLP:journals/corr/SennrichHB16, durrani-etal-2010-hindi}. Removing diacritics may also be useful for addressing data sparsity and/or out-of-vocabulary issues~\cite{williams-etal-2016-edinburghs}. In certain instances, the removal of diacritics has been found to improve BLEU score~\cite{DBLP:journals/corr/SennrichHB16}. While the reasons for diacritics removal are explicit in some cases, other studies have not explicitly stated their motivations~\cite{DBLP:journals/corr/abs-1809-00125}. 
Meanwhile, retaining diacritics can enhance performance for certain languages but may have a detrimental effect on others~\cite{adebara-abdul-mageed-2022-towards}. When to retain or remove diacritics remains an open question that this paper also hopes to address. Finally, restoration of diacritics has positive impact on MT systems in languages like Arabic and Yor\`{u}b\'{a} ~\cite{alqahtani-etal-2016-investigating, adelani-etal-2021-effect}.

\noindent\textbf{Diacritization.} A number of works focus on the task of diacritization. For example,~\citet{belinkov-glass-2015-arabic} employ a Bi-LSTM-based model to create a many to many recurrent neural network to perform diacritization. \citet{mubarak-etal-2019-highly} build a transformer-based sequence-to-sequence framework to train a diacritization model for Arabic. \citet{laki2020automatic} create diacritization models with transformer architecture for $14$ East European languages.

\noindent\textbf{Improving Diacritization with MT.}
\citet{thompson-alshehri-2022-improving} propose an approach for Arabic diacritization that uses MT as an auxialiary task in a multi-task setting. Their findings reveal that incorporating translation improves performance of diacritization. They hypothesize that this improvement stems from the implicit acquisition of semantic knowledge during the training of the MT process. While their experiments focus solely on Arabic, our study expands the scope to cover a broader range of languages, specifically $55$ languages across African and European regions.

\section{Experiments}\label{sec:exp}

\subsection{Setup}\label{sec:exp_setup}
We collect an extensive set of $55$ language pairs where the target language is \textbf{always English} under different train sizes (five sizes for African languages and nine sizes for European languages, detailed in Section~\ref{sec:train_sizes}). For every pair of train size and language pair, e.g. ($125$k, \textit{fr-en}) and ($5$k, \textit{bex-en}) , we build four types of models as illustrated in Figure~\ref{fig:setup}. We list each model type along with the corresponding research question in Table~\ref{tab:model_comparison}. For our single-task setting, there are three types of models: (i) models that perform MT and are trained with undiacritized source (\textbf{OnlyMT\textsubscript{undia}}), (ii) models that perform MT and are trained with diacritized source (\textbf{OnlyMT\textsubscript{dia}}), and (iii) models that perform diacritization (\textbf{OnlyDia}). The only distinction between the two OnlyMT models lies in whether diacritics are incorporated into the source sequences. For the multitask setting, (iv) a \textbf{DiaMT} model is trained to perform both diacritization and translation.

\begin{table}[h]
\centering
\tiny 
 \begin{tabular}[t]{EOEN}
 \toprule
\multicolumn{3}{c}{\textbf{Models Compared}} & \textbf{Research Question} \\
\midrule
DiaMT & vs. & OnlyMT\textsubscript{undia} & Does diacritization benefit MT? (\textbf{RQ1a}) \\
\midrule
DiaMT & vs. & OnlyDia & Does MT benefit diacritization? (\textbf{RQ1b}) \\
\midrule
OnlyMT\textsubscript{dia} & vs. & OnlyMT\textsubscript{undia} & What effect does keeping/removing diacritics have on MT? (\textbf{RQ2})\\
\bottomrule

\end{tabular}
\caption{Models compared and corresponding RQs.}  \label{tab:model_comparison}
\end{table}

\subsection{Evaluation Metrics} 
We use BLEU score~\cite{papineni-etal-2002-bleu} with SACREBLEU implementation~\cite{post-2018-call}\footnote{https://pypi.org/project/sacrebleu/} to measure the performance of MT. For diacritization, we adopt diacritization error rate (DER) and word error rate (WER)~\cite{Abandah2015} with implementation details described in Appendix~\ref{sec:implementations_DER_WER}.

\subsection{Models \& Training}
We adopt transformer architecture~\cite{vaswani2017attention} for all models and train from scratch with the Fairseq library~\cite{ott2019fairseq}, each using a single Nvidia A100 GPU. For train sizes $1$k, $2$k, $3$k, $4$k, $5$k, the number of steps is $30$k. For higher train sizes, we use $100$k steps for $25$k, $500$k steps for $125$k, $1.5$M steps for $625$k, and $3$M steps for $1$M train size. We evaluate our test set on the model with the best performance (lowest loss) on development set. Detailed information about  hyperparamter settings, software version and license are included in Appendix Table~\ref{tab:model_hyperparameters}. %

\section{Data}\label{sec:data}
\subsection{Data Sources}\label{sec:data_source}
\noindent\textbf{African languages.}
To conduct our study, we use a random sample of African languages from the parallel Bible Corpus \cite{mayer-cysouw-2014-creating} which consists of $830$ languages. Specifically, we focus on the subset of $297$ African languages that use diacritics and randomly select $36$ African languages from these. We use the Bible because we assume it will provide correct and consistently diacritized data for our experiments. In Table~\ref{tab:chr_variants_afri}, we present the diacritical systems found in these African languages. The table showcases a diverse range of diacritics with varying levels of complexity. Some languages have simple diacritical systems, where a single diacritic is applied to each character, as seen in languages such as Paasaal (\textit{sig}) and Hdi (\textit{xed}). In contrast, other languages have base characters capable of accommodating multiple diacritics. For instance, in the language Mundani (\textit{mnf}), the character \textipa{\textsubrhalfring{\^a}} carries two diacritics simultaneously.

\noindent\textbf{European Languages.}
We use $19$ European languages from the European Parliament corpus~\cite{koehn-2005-europarl}.\footnote{The data we use is the updated 2012 version which can be accessed at \url{https://www.statmt.org/europarl/}} All of these languages use  diacritics~\cite{10.1007/3-540-45715-1_35, wells2000orthographic} in their orthography. We select this corpus because we assume the diacritics in the document will be correct and consistent, given the domain it is derived from. 

We observed code-switching phenomenon in the dataset. For example, a Spanish sentence may include French word(s). To ensure a clean comparison across these languages, we use fasttext tool~\cite{joulin2016bag, joulin2016fasttext} to identify and remove lines with heavy code-switching.\footnote{lid.176.bin edition of language identification tool with access at \url{https://fasttext.cc/docs/en/language-identification.html}} Specifically, we remove a line if the model prediction of the respective language is lower than $90\%$.\footnote{In spite of this measure, a manual inspection still uncovers a few examples of foreign characters in the data, which we assume have a minimal adverse effect  on our experiments. We show the diacritical system extracted from the data in Table~\ref{tab:chr_variants_euro} which may include foreign characters and diacritics. For African languages, since the domain is the Bible, we assume there are no foreign or code-switched texts. Therefore, we do not carry out any data cleaning for African languages.} Furthermore, we remove overly long and short lines. Specifically, we remove lines with $>500$ or $<6$ characters.

\subsection{Train Sizes}\label{sec:train_sizes}
To determine any interaction between performance and data sizes, we experiment with varying amounts of training data across different experiments. We now provide details of these train sizes for African and European languages. 

\noindent\textbf{African.}
We shuffle the data before we split it into $80\%$ for training (Train), $10\%$ for development (Dev), and $10\%$ for testing (Test). We have $5$ train sizes for African languages ($1$k, $2$k, $3$k, $4$k,  $5$k). Henceforth, the term `$5$k' is used to denote the full training set for each language, reflecting the approximate number of examples in these sets.\footnote{Morokodo (\textit{mgc}) has $2$k as its largest train size as an exception.} The number of examples for each language is listed in Appendix Table~\ref{tab:dataset_split_datapoints_afri}.

\noindent\textbf{European.}\label{sec:data_euro_lang}
We split the data and assign $1,500$ data points to Test, another $1,500$ data points to Dev, and the remaining data as Train. We then subset training data into the $9$ train sizes in the set \{$1$k, $2$k, $3$k, $4$k, $5$k, $25$k, $125$k, $625$k, $1$M\}. The Train/Dev/Test split information is in Appendix Table~\ref{tab:dataset_split_datapoints_euro}.

\subsection{Data Processing}\label{sec:data_processing}

\begin{table}[h]
\centering
\scriptsize
 \begin{tabular}[t]{Tccc}
 \toprule
\multicolumn{2}{c}{\textbf{Model}} & \textbf{Source}  & \textbf{Target} \\
\midrule

\multicolumn{2}{c}{OnlyDia} & t a c k | s a  | m y c k e t & t a c k | s a \r{} | m y c k e t \\
\midrule

\multicolumn{2}{c}{OnlyMT\textsubscript{undia}} & t a c k | s a  | m y c k e t & thank you very much \\
\midrule

\multicolumn{2}{c}{OnlyMT\textsubscript{dia}} & t a c k | s a \r{} | m y c k e t  & thank you very much\\ 
\midrule

\multirow{2}{*}{DiaMT} & Dia & \text{$\varepsilon$} t a c k | s a  | m y c k e t &  t a c k | s a \r{} | m y c k e t \\ 
& MT & \text{$\tau$} t a c k | s a  | m y c k e t & thank you very much \\ 
\bottomrule

\end{tabular}
\caption{An example of source and target for four different types of models. }  \label{tab:data_format}
\end{table}

The format of source and target of the processed data can be seen in Table~\ref{tab:data_format}. We handle non-English (source languages) and English (target language) data differently. For non-English data with diacritics, we (1) decompose every character carrying diacritic(s) into a base character and independent diacritic(s) with NFKD normalization,\footnote{\url{https://unicode.org/reports/tr15/}} (2) replace word-boundary whitespaces with the symbol `|' to maintain information of word boundary after tokenization, (3) insert a whitespace between characters in preparation for whitespace tokenization, and (4) employ whitespace tokenization to build character-level vocabulary which includes characters and diacritics as tokens.\footnote{An exception is the vocabulary for OnlyMT\textsubscript{undia} which has no diacritics because the source side is undiacritized and the target side is English, a language without diacritics~\cite{10.1007/3-540-45715-1_35}.} Decomposing text with NFKD to retrieve independent diacritics and build character-level vocabulary enables better generalization of the model for rare combinations of a base character and diacritic(s). In addition, it helps avoid data sparsity that can occur if word or sub-word tokenization is used. For example, the probability distribution of the variants of `o' in the African language Fon (\textit{fon}) is skewed. The probabilities are about $60.8\%, 38.1\%, 1.1\%$ for  o, \'o, \v{o}, respectively.  Without decomposition, it could be very difficult for the model to learn a decent embedding representation for \v{o} since there is a limited number of examples from which the model can capture its linguistic information. By making each diacritic a token, the model may be able to learn a generalized pattern for diacritic \v{} because it can learn its linguistic behavior in not only \v{o} but also other characters that carry this diacritic in this language, e.g., \v{e}, \v{\i}.

For English data, we tokenize it with whitespace to form word-level tokens. We strive to minimize the introduction of uncontrolled variables by utilizing word-level tokenization. Unlike word-level tokenization, BPE~\cite{sennrich-etal-2016-neural} and BPE-related implementations of subword tokenization can introduce additional uncontrolled variables to the experiments. In particular, the frequency component in BPE renders this method dependent on the corpus. The sampling and language model components in SentencePiece~\cite{kudo-richardson-2018-sentencepiece}, render it both corpus-dependent and non-deterministic. If we adopt these methods, for a piece of text in English, it can be tokenized differently for different (1) language pairs and (2) train sizes. For (1), as an example, the word ‘review’ could be tokenized into [‘rev’, ‘iew’] in the fr-en language pair, but [‘re’, ‘view’] in the es-en language pair. Similarly for (2), ‘review’ can be tokenized differently in 25k and 1M train sizes. We use word-level tokenization to avoid inconsistency in tokenization. With word-level tokenization, a piece of English text is tokenized identically throughout different train sizes and language pairs. This enhances the comparability among different settings.

For {\bf DiaMT}, we prepend a symbol (and a following whitespace), $\varepsilon$ for diacritization and $\tau$ for MT, at the beginning of every source sequence to prime the model which of the two tasks (translation or diacritization) to perform for a specific input sequence. The source side for both sub-tasks is identical, except the prepended symbol. The potential advantage of this design is that the model may be able to gain positive transfer via attaining cross-task knowledge. 

\subsection{Post-processing Predictions}\label{sec:postprocessing}
When processing non-English data, we use whitespace to separate characters and the symbol `|' to denote word boundaries. During post-processing for diacritization output, we consolidate the separated characters back into words and substitute the `|' symbol with whitespace to properly indicate word boundaries. It is after this post-processing step that we compute DER and WER metrics. In contrast, when performing MT, post-processing is not required. This is because the output is always in English, a language we process straightforwardly from the outset, thereby eliminating the need for any post-processing adjustments.

\subsection{Complexity Metrics}\label{sec:proposed_complexity_metric}

\begin{table}[h]
\scriptsize
 \begin{tabular}[t]{cS}
 \toprule
 \textbf{Metric} & \textbf{Definition} \\
\midrule
DCR & Proportion of characters that carry diacritic(s) out of all characters.\\

DWR & Proportion of words with at least a character carrying diacritic(s) out of all words.\\ 

DBR & Average number of variants (including itself) of each base character.\\ 

DWSR & Average number of words with at least a character carrying diacritic(s) per sentence.\\
\midrule

AED & Average entropy of the distributions of each base character's variant(s) and itself.\\

WAED & Weighted AED with weight being the proportion of the number of occurrence of each base character out of that of all base character(s). \\ \hline

\end{tabular}
\caption{Definitions of Proposed Complexity Metrics. }  \label{tab:complexity_metrics_definitions_formula}
\end{table}

\noindent The functional load of diacritics differs from one language to another \cite{jbp:/content/journals/10.1075/wll.12.1.07rob, bird_1999}. As a result, we propose two classes of metrics which may be able to measure some aspects of the functional load of the diacritical system. We refer to these metrics as \textbf{complexity metrics}. They rely only on unlabeled corpora, unlike existing metrics which require a formal lexicon \cite{pauw2007automatic}. Thus, they are well suited for scenarios where lexicons are unavailable. Besides, they are \textbf{language-agnostic} such that they are applicable to any given language. They measure \textbf{(1)} the ratio of diacritics and character/word/sentence, and \textbf{(2)} the entropy of the probability distribution of character-diacritic combinations. A simplified example corpus and the computation of its complexity metrics values are given at Appendix  Table~\ref{tab:aed_waed_ex}.

\begin{figure*}[ht!]
\begin{centering}
\includegraphics[scale=0.315]{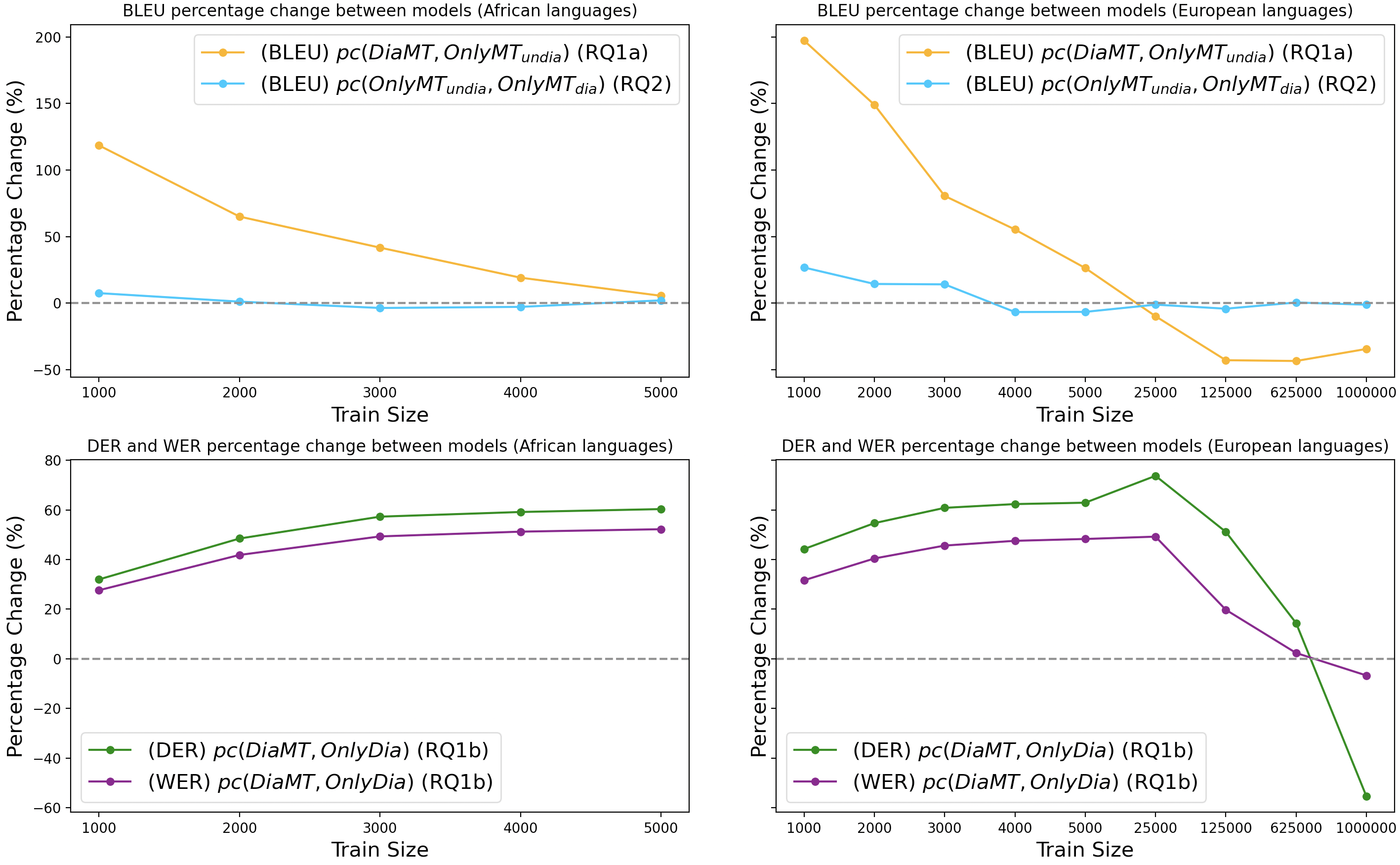}
  \caption{Percentage change of the BLEU/DER/WER averages among languages in each train size. $pc(m1, m2)$ is the percentage change of the metric values produced by model $1$ (m1) over model $2$ (m2) with $pc(m1, m2) = (m1-m2)/m1$. We indicate the research question each line addresses in the legends. Left column: African languages. Right column: European languages. Top row: BLEU scores. Bottom row: DER and WER.}
  \label{fig:percentage_change}
 \end{centering}
\end{figure*}

To determine \textbf{(1)}, we measure Diacritized Character Ratio (\textbf{DCR}), Diacritized Word Ratio (\textbf{DWR}),  Diacritized Base character Ratio (\textbf{DBR}), and Diacritized Word Sentence Ratio (\textbf{DWSR}). To formulate the complexity metrics, for a corpus of any given language, let $c, c_d$ be the number of characters and diacritized characters; let $w, w_d$ be the number of words and words with at least one diacritized character; let $b$ be the number of unique base characters, $b_d$ be the number of unique character-diacritic(s) combinations and $s$ be the number of sentences. Then, $DCR = c_d/c$, $DWR= w_d/w$, $DBR=b_d/b$, and $DWSR = w_d/s$.

For \textbf{(2)}, we measure Average Entropy of Diacritics (\textbf{AED}), and Weighted Average Entropy of Diacritics (\textbf{WAED}). AED serves as an assessment of the challenge faced by a diacritization model in diacritizing a character (including the decision not to diacritize). It is computed by averaging the entropies of the probability distribution of character-diacritic combinations for each base character. The more uniformly distributed they are, the more challenging it becomes for the model to make accurate predictions. WAED is the weighted edition of AED where the weight is the frequency of each base character.

It is important to mention that our proposed complexity metrics are theoretically data-dependent. That is, a single language can have different complexity metric values given different datasets and/or train sizes. However, empirically, as can be seen in Tables~\ref{tab:leveled_dia_complexity_metrics_afri} and~\ref{tab:leveled_dia_complexity_metrics_euro}, the values are similar across different train sizes for each language. This demonstrates that our proposed complexity metrics are robust among different sizes of training data and can capture the complexity of a diacritical system consistently. The proposed metrics are useful because \textbf{(1)} they provide a quantitative view of the diacritical system, \textbf{(2)} it is straightforward to compute them, and \textbf{(3)} they show high correlation with model performance as discussed later in Section~\ref{sec:positive_correlation_complexity_metrics_and_performance_metrics}.

\section{Results and Analyses}\label{sec:results}
\begin{table*}
\centering
\tiny
\setlength{\tabcolsep}{3pt}
\begin{tabular}{lrrrccrrcrrc}
\toprule
\multicolumn{12}{c}{African Languages} \\
\midrule
& \multicolumn{3}{c}{Avg. BLEU} & \multicolumn{2}{c}{pv. BLEU} & \multicolumn{2}{c}{Avg. DER} & pv. DER & \multicolumn{2}{c}{Avg. WER} & pv. WER\\
 \cmidrule(lr){2-4}\cmidrule(lr){5-6}\cmidrule(lr){7-8}\cmidrule(lr){9-9}\cmidrule(lr){10-11}\cmidrule(lr){12-12}
\midrule
Size & $DM$ & $OM_u$ & $OM_d$ & $p(OM_u,OM_d) (ES)$ & $p(DM,OM_u) (ES)$ & $DM$ & $OD$ & $p(DM, OD) (ES)$ & $DM$ & $OD$ & $p(DM, OD) (ES)$ \\
\midrule
1k & 2.306 & 1.055 &  0.981 & >.05 (0.13) & <.01 (1.88) & 0.428 & 0.291 & <.01 (1.57) & 0.478 & 0.346 &   <.01 (1.28) \\
2k & 3.121 & 1.891 &  1.869 & >.05 (0.04) & <.01 (1.61) & 0.455 & 0.235 & <.01 (2.62) & 0.504 & 0.293 &   <.01 (2.05) \\
3k & 3.384 & 2.388 &  2.477 & >.05 (0.21) & <.01 (2.23) & 0.487 & 0.208 & <.01 (3.81) & 0.536 & 0.271 &   <.01 (2.86) \\
4k & 3.495 & 2.934 &  3.017 & >.05 (0.20) & <.01 (1.37) & 0.511 & 0.209 & <.01 (3.89) & 0.559 & 0.272 &   <.01 (2.96) \\
5k & 3.577 & 3.390 &  3.319 & >.05 (0.15) & <.01 (0.43) & 0.512 & 0.203 & <.01 (3.48) & 0.559 & 0.267 &   <.01 (2.75) \\
\midrule
\multicolumn{12}{c}{European Languages} \\
\midrule
1k    &  1.689 &  0.568 &   0.448 & >.05 (0.43) & <.01 (2.87) & 0.468 & 0.261 & <.01 (4.01) & 0.571 & 0.390 &   <.01 (3.10) \\
2k    &  1.994 &  0.801 &   0.700 & >.05 (0.32) & <.01 (2.55) & 0.489 & 0.222 & <.01 (5.06) & 0.591 & 0.352 &   <.01 (4.08) \\
3k    &  2.062 &  1.142 &   1.000 & >.05 (0.39) & <.01 (1.72) & 0.522 & 0.204 & <.01 (6.32) & 0.620 & 0.337 &   <.01 (5.13) \\
4k    &  2.273 &  1.463 &   1.567 & >.05 (0.33) & <.01 (1.65) & 0.555 & 0.209 & <.01 (7.71) & 0.649 & 0.340 &   <.01 (5.56) \\
5k    &  2.337 &  1.849 &   1.978 & >.05 (0.23) & <.01 (0.88) & 0.562 & 0.208 & <.01 (6.48) & 0.655 & 0.339 &   <.01 (5.45) \\
25k   &  4.496 &  4.984 &   5.039 & >.05 (0.06) & <.01 (0.59) & 0.296 & 0.078 & <.01 (5.50) & 0.420 & 0.213 &   <.01 (4.02) \\
125k  &  7.381 & 12.909 &  13.465 & <.05 (0.17) & <.01 (2.21) & 0.091 & 0.045 & <.01 (1.52) & 0.225 & 0.180 &   <.01 (1.05) \\
625k  & 12.085 & 21.357 &  21.246 & >.05 (0.03) & <.01 (2.94) & 0.025 & 0.021 & >.05 (0.34) & 0.163 & 0.159 &   >.05 (0.13) \\
1M    & 15.893 & 24.213 &  24.492 & <.05 (0.08) & <.01 (2.44) & 0.018 & 0.029 & >.05 (0.50) & 0.160 & 0.171 &   >.05 (0.33) \\
\bottomrule
\end{tabular}
\caption{Average (Avg.), p-value and effect size (ES) in terms of Cohen's d of BLEU of 3 different models, $OnlyMT_{undia}(OM_u)$, $OnlyMT_{dia}(OM_d)$ and $DiaMT(DM)$, and DER/WER of 2 different models $OnlyDia(OD)$ and $DiaMT(DM)$, at different train sizes (5 for African, 9 for European languages). $p(m1,m2)$ represents the p-value of two-sided paired t-test between BLEU/DER/WER produced by model $m1$ and model $m2$. Effect sizes are with respect to Cohen's d. }\label{tab:leveled_sig_test_with_cohen_d}
\end{table*}
\normalsize

\subsection{Findings to Research Questions}\label{sec:findings_to_RQ}
We discuss findings to our research questions based on results reported in Table~\ref{tab:leveled_sig_test_with_cohen_d} and the visualization shown in Figure~\ref{fig:percentage_change}. We report a significance test with paired t-test for the performance of each pair of compared models, along with Cohen's \textit{d} ~\cite{cohen1977statistical} to estimate the \textit{effect sizes}, as significance tests alone may not capture the magnitude of the effect~\cite{cumming2013understanding}. To interpret Cohen's \textit{d}, we refer to the standard proposed in~\citet{sawilowsky2009new}: 0.01 (very small), 0.2 (small), 0.5 (medium), 0.8 (large), 1.2 (very large), and 2.0 (huge).

\textbf{RQ1a. Does diacritization benefit MT?} As Figure~\ref{fig:percentage_change} shows, on average, diacritization improves MT performance when train size is  $\le5$k and harms MT performance when train size is $>5$k. For each individual language, the performance gain is in general positive for both African and European languages as can be seen in Appendix Figures~\ref{fig:bar_bleu_afri_diamt_onlymt_undia} and~\ref{fig:bar_bleu_euro_diamt_onlymt_undia}.\footnote{The BLEU scores and exact percentage changes between DiaMT and OnlyMT\textsubscript{undia} are shown in Appendix Tables~\ref{tab:bleu_afri_exact_values} and~\ref{tab:bleu_euro_exact_values} where some of the languages achieve over $300\%$ gain after adding diacritization when train size is $\le5$k.} However, for $>5$k train sizes, adding diacritization in general harms MT performance. As the significance tests in Table~\ref{tab:leveled_sig_test_with_cohen_d} show, $p(DM,OM_u)$, the p-values of paired $t$-test between the BLEU scores of DiaMT and OnlyMT\textsubscript{undia} are lower than $0.01$ throughout all train sizes and language regions. This supports that adding diacritization will significantly affect MT performance, positively when $\le5$k, and negatively when $>5$k. We observe a gradual decrease of effect size from $1$k to $5$k for both African and European languages, and a rapid increase after $25$k for European languages. That is, the benefit of adding diacritization gradually reduces from $1$k to $5$k, and the harm grows rapidly after $25$k from small to huge.

The unexpected negative transfer effect on MT performance following the inclusion of diacritization as an auxiliary task in higher-resource scenarios warrants careful examination. While it might be tempting to attribute this to an inadequately sized model struggling to learn both tasks simultaneously, our analysis, as detailed in \textbf{RQ1b}, reveals a contrary trend. Interestingly, certain languages exhibit enhanced diacritization performance after the incorporation of MT, indicating that the model's capacity is indeed sufficient to accommodate both tasks. Furthermore, the equitable distribution of data between MT and diacritization tasks, each constituting $50\%$, eliminates data imbalance as a contributing factor. Thus, the observed phenomenon likely originates from external variables, underscoring the need for further studies to pinpoint its underlying cause.

\begin{figure}[h!]
\begin{centering}
\includegraphics[scale=0.19]{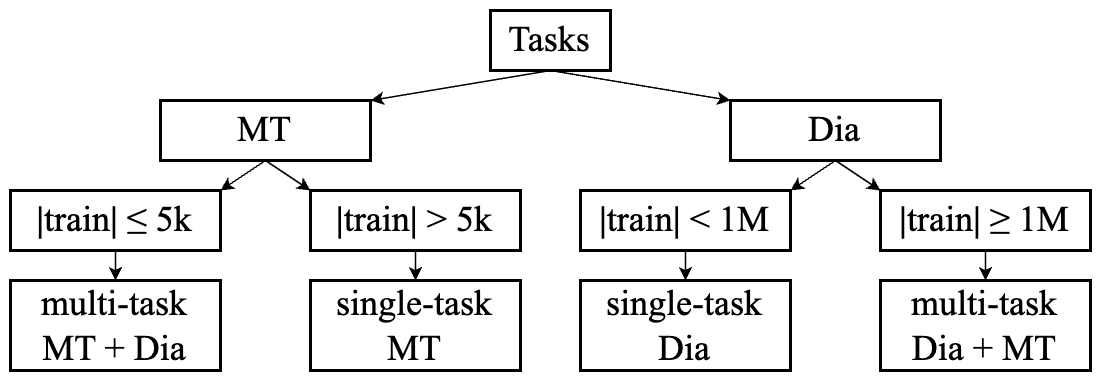}
  \caption{A guideline for training strategies under different data size conditions for diacritization (\textbf{Dia}) and/or machine translation (\textbf{MT}) derived by approaching RQ1a and RQ1b under different train sizes. }
  \label{fig:key_findings}
 \end{centering}
\end{figure}

\noindent\textbf{RQ1b. Does MT benefit diacritization?} We find that adding MT as an auxiliary task on average undermines diacritization performance except when train size is at $1$M as can be seen in Appendix Figure~\ref{fig:percentage_change}.  Appendix Figure~\ref{fig:bar_der_afri_onlydia_diamt} and \ref{fig:bar_wer_afri_onlydia_diamt} show that it is rare to have improvements in diacritization performance after adding MT with two exceptions: Fon (\textit{fon}) at $1$k and Sekpele (\textit{lip}) at $2$k. Appendix Figure~\ref{fig:bar_der_euro_onlydia_diamt} and~\ref{fig:bar_wer_euro_onlydia_diamt} show a similar phenomenon. When train sizes are  $\le125$k, only Slovak (sk) (at $125$k) experiences a small improvement on WER. When train size is $625$k, two languages (Greek and Finnish) out of $10$ languages, experience improvement. When train size is $1$M, four languages, out of nine, experience a gain in DER and WER after adding MT: Greek (\textit{el}), Finnish (\textit{fi}), Italian (\textit{it}), and Portuguese (\textit{pt}) with Greek and Finnish experiencing a great boost. Greek has $79.6\%$ and $28.3\%$ of reduction in DER and WER, respectively. Finnish has $46.2\%$ and $19.4\%$ of reduction in DER and WER, respectively. Despite that the other five European languages do not enjoy the gain, they demonstrate manageable losses in DER and minimal losses in WER. Overall, the paired t-test indicates that adding MT significantly harms diacritization performance when $<625$k and a neutrality when $\ge625$k. We observe huge effect sizes for both DER and WER when train size $<125$k. The effect sizes reduce quickly after $\ge125$k to the values between very small to small. That is, the negative effect of adding MT to diacritization decreases as the train size goes up.

\citet{thompson-alshehri-2022-improving} also find that when the dataset is large, Arabic diacritization can benefit from the addition of MT as an auxiliary task. Hence, we recommend adding MT to diacritization when training with $\ge1$M train size because there potentially can be a performance boost. Even if not, the negative effect is manageable.\footnote{The DER/WER values and percentage change between DiaMT and OnlyDia are shown in Tables~\ref{tab:der_wer_afri_exact_values} and~\ref{tab:der_wer_euro_exact_values}.}

After studying RQ1a and RQ1b, a notable asymmetry emerges in the relationship between MT and diacritization at higher-resource scenarios when introduced as auxiliary tasks. Specifically, while the inclusion of diacritization adversely affects MT performance, the incorporation of MT may yield benefits for diacritization. To summarize, we propose a guideline of either training in single-task or multi-task fashion in Figure~\ref{fig:key_findings}, tailored to varying sizes of the training set.

\noindent\textbf{RQ2. What effect does removing/keeping diacritics have on MT?} As introduced in Section~\ref{sec:intro}, diacritics can carry  semantic meanings. Removing diacritics can lead to the loss of the information. In MT, the lack of diacritics at source side can produce ambiguity and pose challenges to the MT system. Therefore, we hypothesize that removing diacritics (OnlyMT\textsubscript{undia}) would negatively impact the MT performance, compared to diacritics being retained (OnlyMT\textsubscript{dia}). 

Nonetheless, our experimental results show that the MT system perform indifferently regardless of diacritics of source language being kept or removed. The mean difference of BLEU scores between OnlyMT\textsubscript{undia} and OnlyMT\textsubscript{dia} is consistently around zero throughout all train sizes and languages of both regions as can be seen in Figure~\ref{fig:percentage_change}.  As shown in Table~\ref{tab:leveled_sig_test_with_cohen_d}, the $p$-values between the BLEU scores of OnlyMT\textsubscript{undia} and OnlyMT\textsubscript{dia} are consistently larger than $0.05$ when $<125$k for both African and European languages. When $\ge125$k, there is inconsistency in the significance test results where we observe $p$ values being less than $0.05$ at $125$k and $1$M, but larger than $0.05$ at $625$k. At $125$k, $625$k, and $1$M, $95\%$, $50\%$, and $89\%$ of language pairs have better performance when source is diacritized, respectively. It seems that when $\ge125$k, the existence of diacritics may benefit translation performance. However, with a closer look into Table~\ref{tab:bleu_euro_exact_values}, the percentage changes of the two models for each language are in general around zero at $1$M train size. That is, the performance differences between two models are minimal at $1$M. Despite that the paired t-test shows significance at $125$k and $1$M, the Cohen's \textit{d} for $125$k and $1$M are $0.17$ and $0.08$, respectively. Both of them are between very small to small, indicating that the effects are little. 

We speculate two potential reasons of the absent effect when diacritics are removed: (1) the contextual clues provided by adjacent words may enhance machine translation quality as effectively as the inclusion of diacritics. That is, MT systems are capable of inferring the missing information based on the contexts. As suggested in \citet{adelani-etal-2021-effect}, an MT system may be capable of learning to disambiguate and generate correct translation even when diacritics are absent at the source side. (2) The infrequent incidence of ambiguity resulting from the removal of diacritics makes it negligible when assessing the performance difference between retaining and removing diacritics.

\subsection{Function of Diacritics and MT Performance}
Despite that we observe minimal impact on MT performance whether diacritics are removed or retained as discussed in our \textbf{RQ2}, the comparison is between OnlyMT\textsubscript{dia} and OnlyMT\textsubscript{undia} among languages with all types of diacritical functions. To further explore the effect, we investigate whether the way diacritics function in each language influences model performance of MT. This is motivated by linguistic studies which find a reading cost in humans when diacritics that perform lexical functions are mismatched \cite{french_2023}. We split the diacritical functions into \textit{lexical function}, where diacritics influence the lexical semantics of a word and \textit{grammatical function}, where the diacritics can change the grammatical structure of a sentence. Due to limited research on diacritics in African languages, our analysis concentrates on European languages. An overview of diacritical functions in these languages is provided in Appendix Table~\ref{tab:euro_diacritics_function}. To conduct an analysis, we categorize European languages into three groups: \textit{lex only}, \textit{gra only}, \textit{lex+gra}, which represent that diacritics have only lexical function, only grammatical function, and both, respectively. We inspect how different groups of diacritical functions will affect translation quality when diacritics are removed by comparing the average BLEU scores produced by OnlyMT\textsubscript{dia} and OnlyMT\textsubscript{undia}  for each group at different train sizes.

\begin{figure}[h!]
\begin{centering}
\includegraphics[scale=0.5]{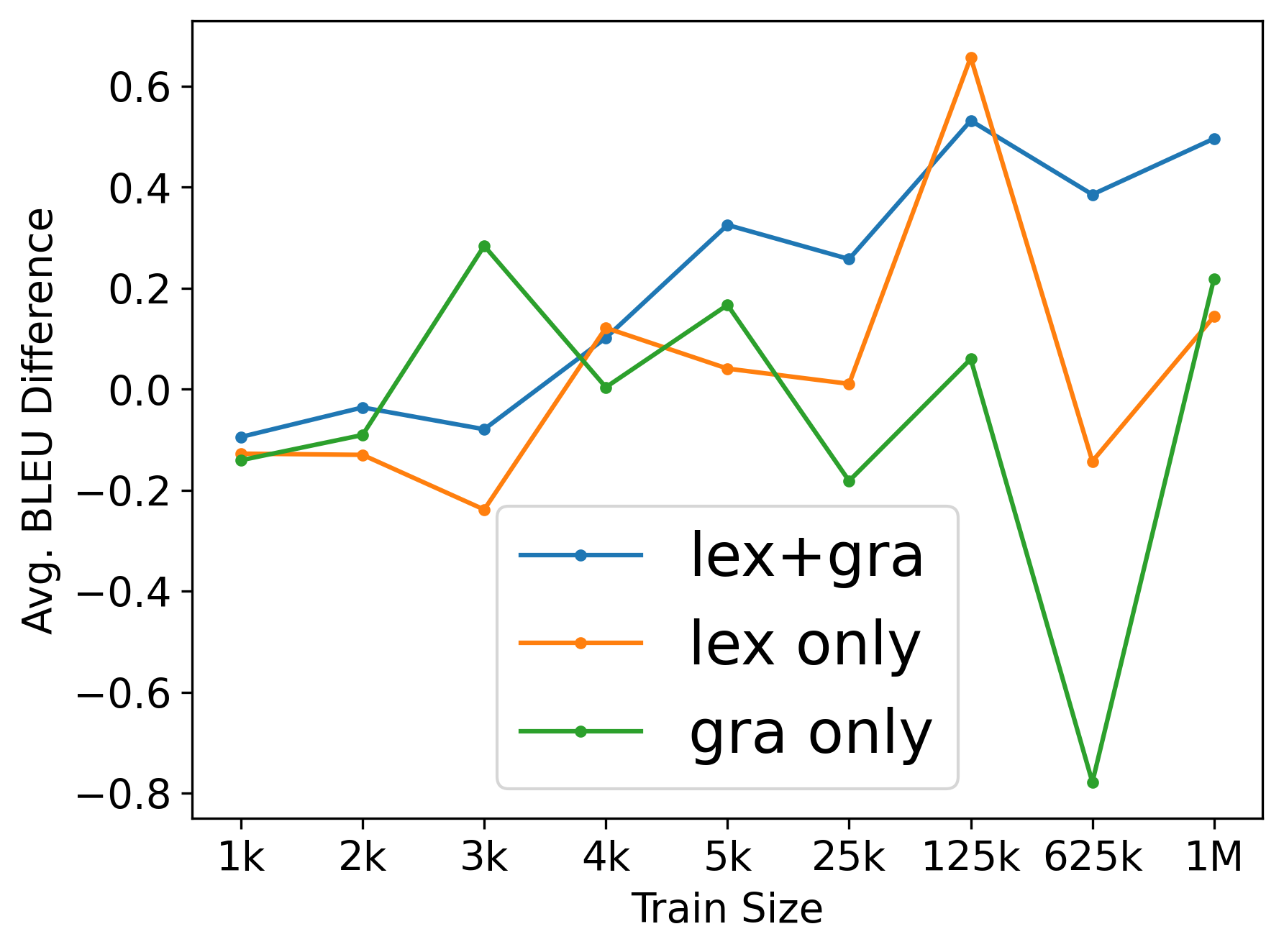}
  \caption{Differences of average BLEU scores between OnlyMT\textsubscript{dia} and OnlyMT\textsubscript{undia} for three different groups of diacritical functions (\textit{lex only}, \textit{gra only} and \textit{lex+gra}) for European languages at different train sizes.}
  \label{fig:onlymt_undia_grouped_by_diacirtical_functions}
 \end{centering}
\end{figure}

We hypothesize that the removal of diacritics would harm languages whose diacritics have lexical function more than those having grammatical function, based on the assumption that grammatical information can be easier to infer from the contexts, compared to lexical information. Hence, we speculate that the differences between mean BLEU scores of OnlyMT\textsubscript{dia} and OnlyMT\textsubscript{undia} would be \textit{lex+gra} $>$ \textit{lex only}  $>$ \textit{gra only} where \textit{lex+gra} having the largest difference because diacritics perform both functions for languages in this group and removing diacritics may lead to heavier loss in information compared to the other two groups. Experimental results, as can be seen in Figure~\ref{fig:onlymt_undia_grouped_by_diacirtical_functions}, show that for train sizes $\le5$k, the differences of average BLEU scores are all around zero among the three different groups without an obvious pattern. However, for $\ge25k$, there is a somewhat consistent order of \textit{lex+gra} $>$ \textit{lex only} $>$ \textit{gra only}, except that the difference for \textit{lex only} is slightly higher than \textit{lex+gra} at $625$k; and \textit{gra only} is slightly higher than \textit{lex only} at $1$M. In part, the experimental results align with our hypothesis.

Although the results show a tendency of performance loss after removing diacritics being \textit{lex+gra} $>$ \textit{lex only}  $>$ \textit{gra only}, it is noteworthy that this finding does not guarantee that languages categorized in these three groups will always follow the order. This is due to the fact that the differences for all three groups are consistently around zero, within the range of 0.66 to -0.78 BLEU score, reflecting the effect of removing diacritics is minimal as discussed in \textbf{RQ2}. Furthermore, this analysis is not conclusive for two reasons: (1) The categorization into groups may overlook subtle but significant linguistic nuances, as languages within the same group might exhibit distinct linguistic characteristics despite their shared classification. (2) A thorough investigation with a representative dataset specifically designed to include ample instances of lexical ambiguity and sentences prone to grammatical ambiguity, after removal of diacritics, is necessary to definitively ascertain the relationship between diacritical functions and MT performance. That is, additional research in this area is needed.

\subsection{Positive Correlation Between Complexity and Performance Metrics}\label{sec:positive_correlation_complexity_metrics_and_performance_metrics}
We propose two classes of complexity metrics as discussed in Section~\ref{sec:proposed_complexity_metric}. The complexity metrics quantify the complexity of the diacritical system of a given language and anticipate that the higher the values of complexity metrics, the more difficult to restore diacritics (i.e. the worse the performance metrics: DER and WER). As for correlation analysis, the proposed complexity metrics exhibit a consistently positive correlation with diacritization performance metrics across both African and European languages at all train sizes.  For instance, the substantial difference in complexity metric DCR between Gidar (\textit{gid}) at $0.001$ and Ndogo (\textit{ndz}) at $0.258$ corresponds to a divergent performance metric DER of $0.097$ for \textit{gid} and $0.330$ for \textit{ndz}.\footnote{OnlyDia model at $5$k as shown in Table~\ref{tab:der_wer_afri_exact_values}.} We use the Train and Dev sets to compute complexity metrics while we measure performance on the Test set alone. We ensure that the data used to measure the complexity metrics and the data used to evaluate model performance are non-overlapping.

To assess the significance of these correlations, three measures, namely Pearson, Kendall, and Spearman correlations, were computed. The resulting p-values, which are predominantly lower than $0.05$ across African and European languages and different train sizes, indicate statistical significance. Examples of profoundly high correlations between complexity and performance metrics include (DCR, DER) with pearson correlation at $0.885$, and (WAED, WER) at $0.788$ at $1$M train size. A high correlation observed with larger training sizes bolsters confidence in the efficacy of the proposed complexity metrics. This finding solidifies the belief that the proposed metrics effectively quantify the complexity of the diacritical system of a language. The correlations between the proposed complexity metrics and DER/WER are detailed in Appendix Table~\ref{tab:leveled_afri_complexity_performance_correlations} for African languages and Table~\ref{tab:leveled_euro_complexity_performance_correlations} for European languages.

There are two exceptions to the strong correlations: DBR and AED. These metrics occasionally exhibit lower correlation with DER and WER. We speculate that (1) DBR in European languages can be biased due to the inclusion of foreign text, as discussed in Section~\ref{sec:data_source}. This may bring about the lower correlation between DBR and performance metrics. (2) The absence of taking character occurrence frequency into consideration may negatively influence the effectiveness of AED. To support this speculation, WAED, the weighted version of AED which takes frequency into consideration shows a high correlation with performance metrics across all train sizes and both language regions.

\section{Conclusion}\label{sec:conclusion}

In this study, we empirically explore the interactions between machine translation (MT), diacritics, and diacritization. We conduct comprehensive experiments involving numerous African and European languages across different dataset sizes. In the multi-task learning setting, we observe that introducing diacritization is advantageous for MT in low-resource scenarios but detrimental otherwise. Additionally, we find that while MT generally has a negative impact on diacritization, it can facilitate substantial performance improvements for specific languages in high-resource settings. In the context of single-task learning, we determine that the removal or retention of diacritics has minimal influence on MT performance. To assess the complexity of diacritical systems, we propose six language-agnostic metrics, establishing a strong positive correlation with our model's performance.

\section*{Limitations}

For our machine translation experiments, we have limited our target language exclusively to English. Consequently, our findings may not be applicable to scenarios where the target language uses diacritics in its orthographic system. Moreover, the datasets used in this study are from religious and political domains, leading us to operate under the assumption that the texts are fully diacritized rather than partially. As such, this introduces a  potential limitation to the generalizability of our results.

\section*{Ethics Statement}
The datasets we employed in this study are derived from two publicly accessible sources: The Bibles and the European Parliament. We consciously chose not to collect or utilize data from any individual subjects to avoid privacy-related ethical issues.

\section*{Acknowledgements}\label{sec:acknow}
We acknowledge support from Canada Research Chairs (CRC), the Natural Sciences and Engineering Research Council of Canada (NSERC; RGPIN-2018-04267), the Social Sciences and Humanities Research Council of Canada (SSHRC; 895-2020-1004; 895-2021-1008), Canadian Foundation for Innovation (CFI; 37771), Digital Research Alliance of Canada,\footnote{\href{https://alliancecan.ca}{https://alliancecan.ca}} and UBC ARC-Sockeye.\footnote{\href{https://arc.ubc.ca/ubc-arc-sockeye}{https://arc.ubc.ca/ubc-arc-sockeye}}

\bibliography{custom}

\appendix

\begin{figure*}[htp!]
\begin{centering}
\includegraphics[scale=0.35]{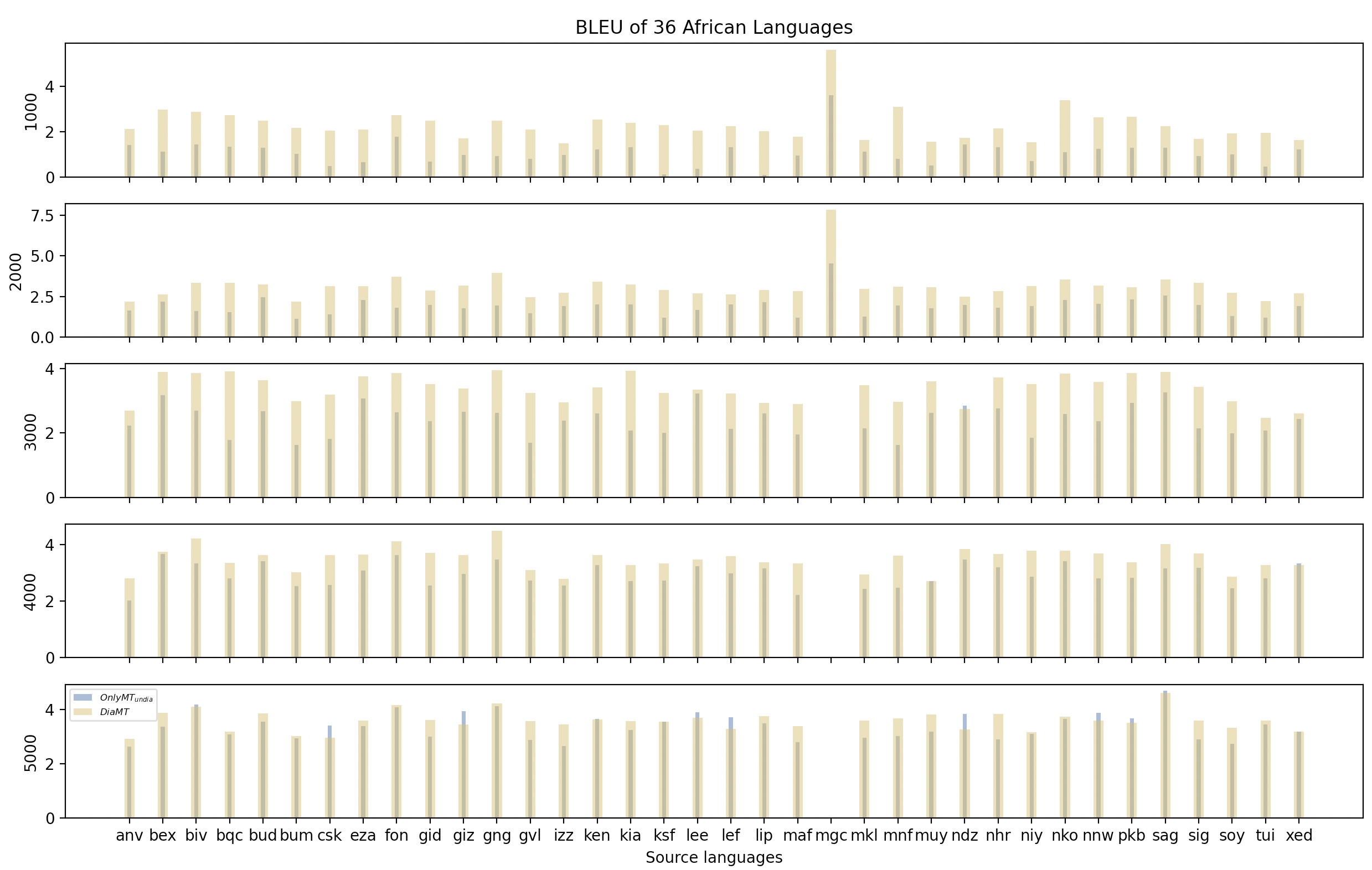}
  \caption{BLEU comparison between DiaMT and OnlyMT\textsubscript{undia} for 36 African languages to English pairs.}
  \label{fig:bar_bleu_afri_diamt_onlymt_undia}
 \end{centering}
\end{figure*}

\begin{figure*}[htp!]
\begin{centering}
\includegraphics[scale=0.35]{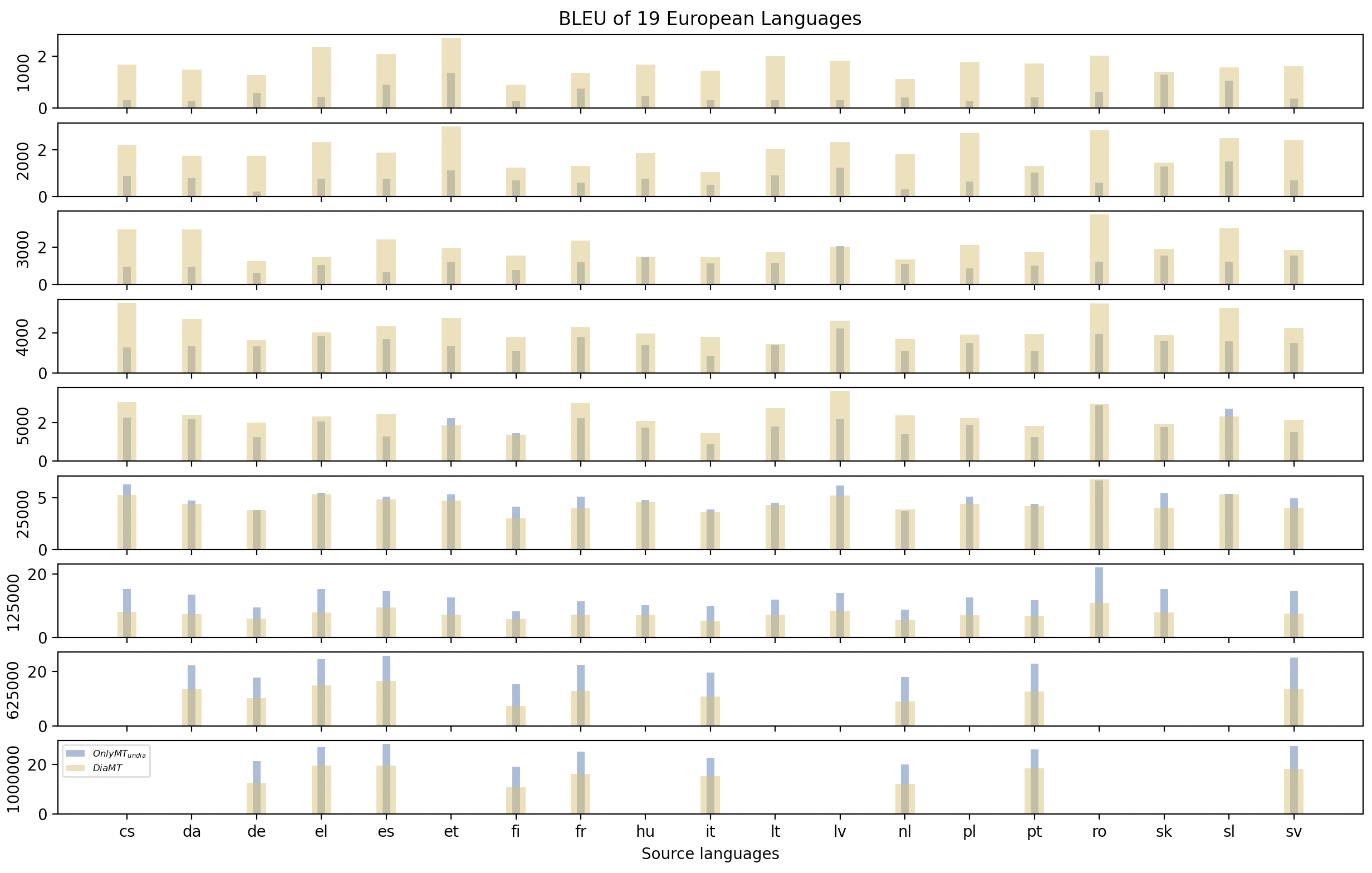}
  \caption{BLEU comparison between DiaMT and OnlyMT\textsubscript{undia} for 19 European languages to English pairs.}
  \label{fig:bar_bleu_euro_diamt_onlymt_undia}
 \end{centering}
\end{figure*}

\begin{figure*}[htp!]
\begin{centering}
\includegraphics[scale=0.35]{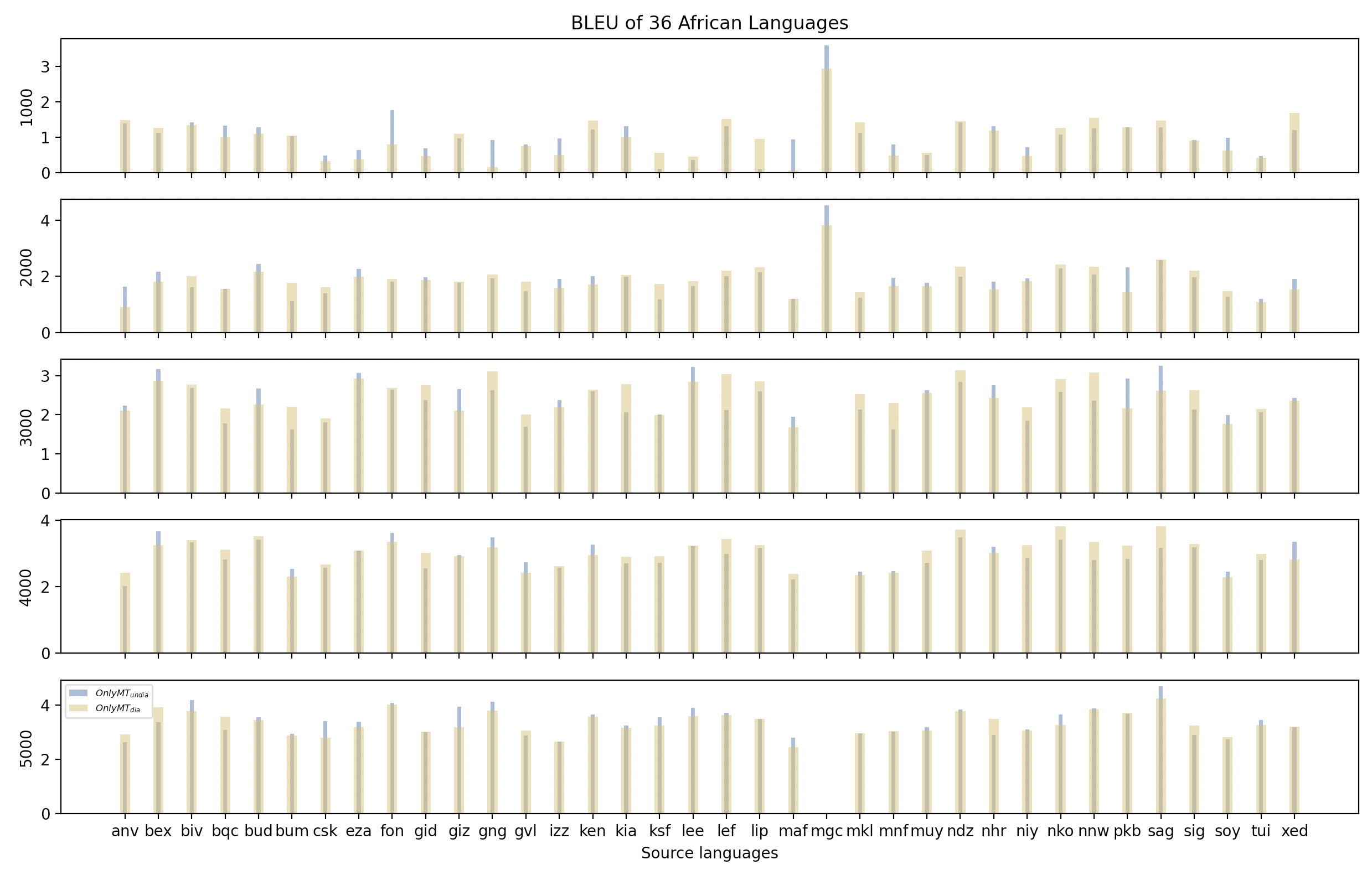}
  \caption{BLEU comparison between OnlyMT\textsubscript{undia} and OnlyMT\textsubscript{dia} for 36 African languages to English pairs.}
  \label{fig:bar_bleu_afri_onlymt_undia_onlymt_dia}
 \end{centering}
\end{figure*}

\begin{figure*}[htp!]
\begin{centering}
\includegraphics[scale=0.35]{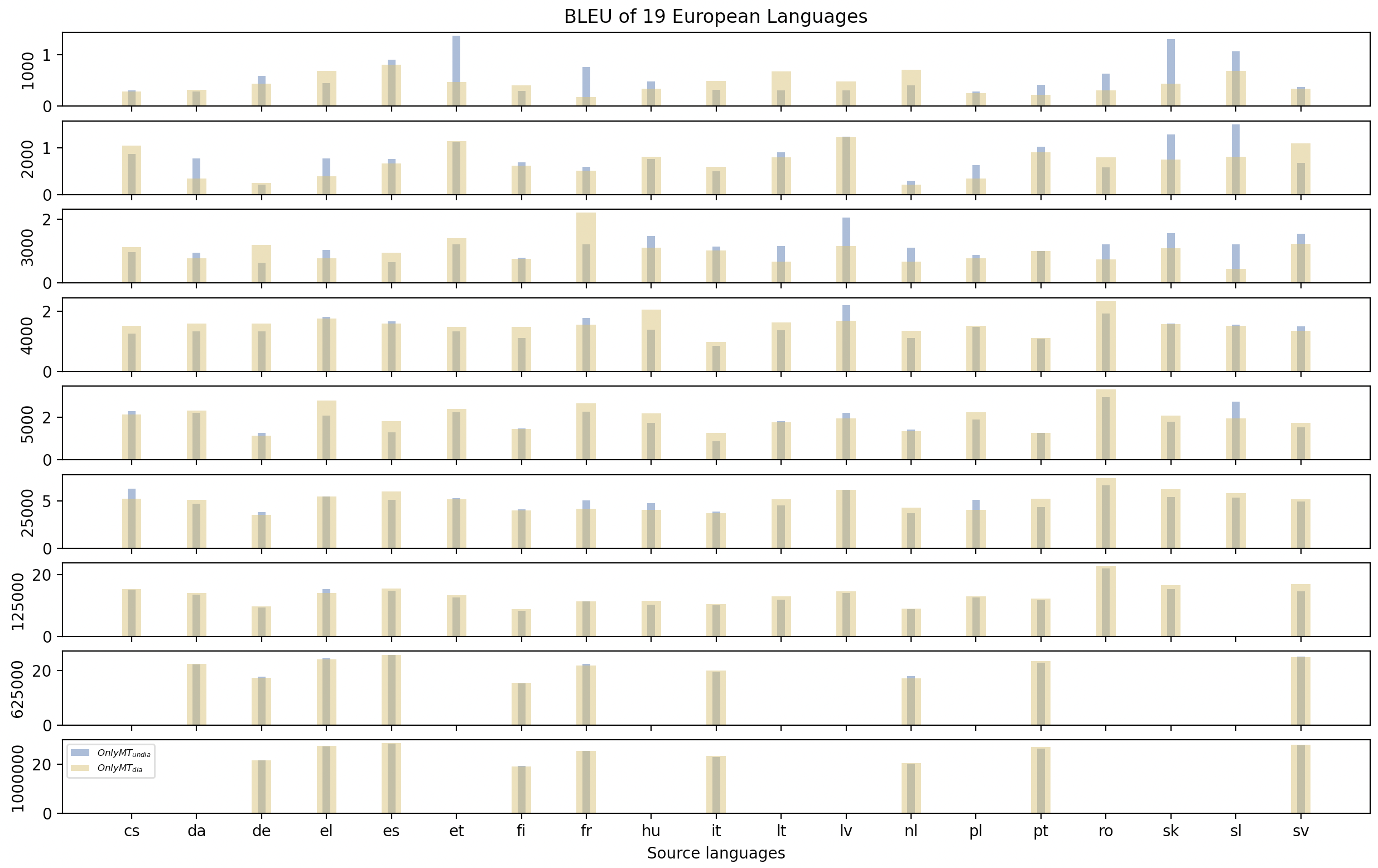}
  \caption{BLEU comparison between OnlyMT\textsubscript{undia} and OnlyMT\textsubscript{dia} for 19 European languages to English pairs.}
  \label{fig:bar_bleu_euro_onlymt_undia_onlymt_dia}
 \end{centering}
\end{figure*}

\begin{figure*}[htp!]
\begin{centering}
\includegraphics[scale=0.35]{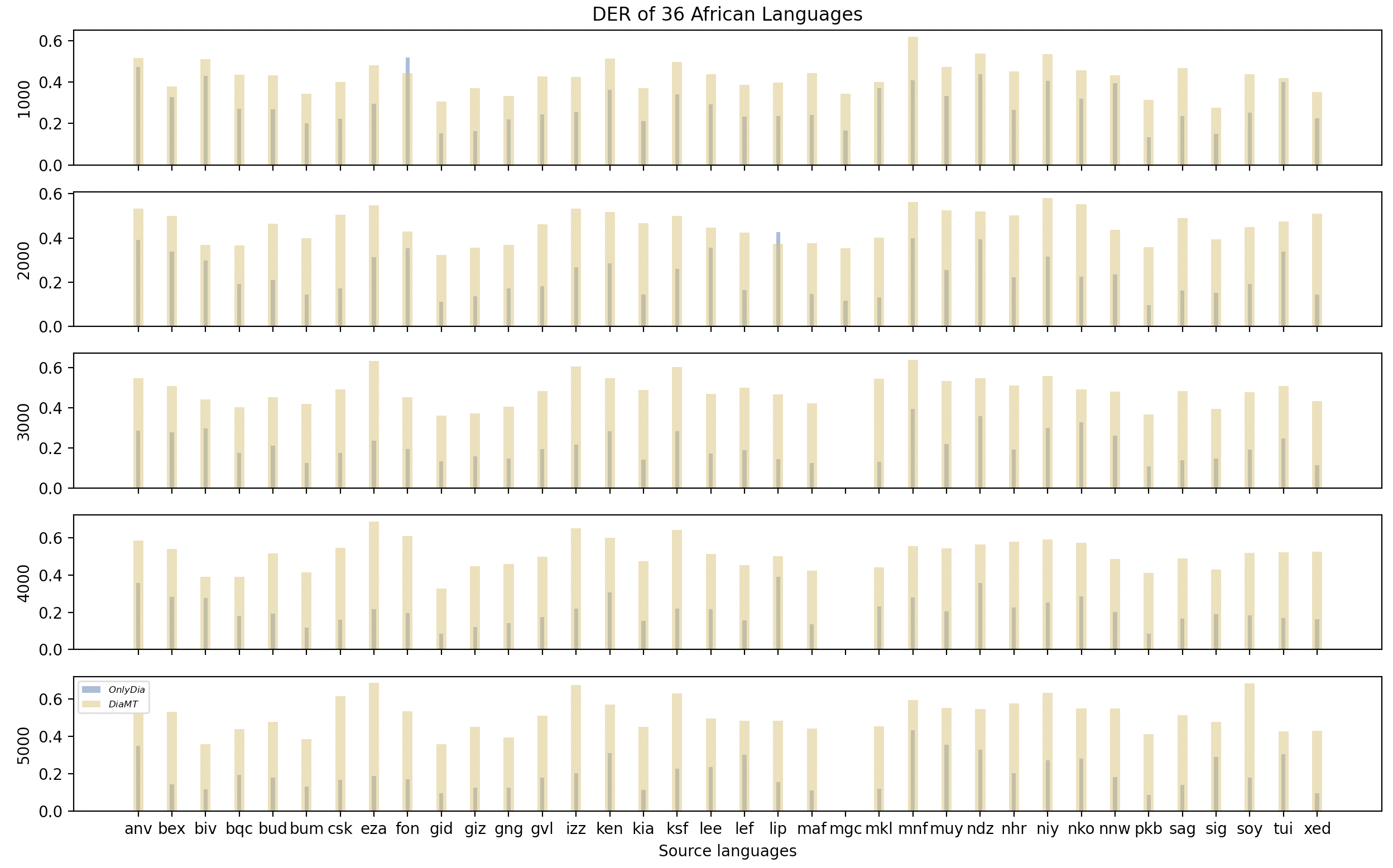}
  \caption{DER comparison between OnlyDia and DiaMT for 36 African languages.}
  \label{fig:bar_der_afri_onlydia_diamt}
 \end{centering}
\end{figure*}

\begin{figure*}[htp!]
\begin{centering}
\includegraphics[scale=0.35]{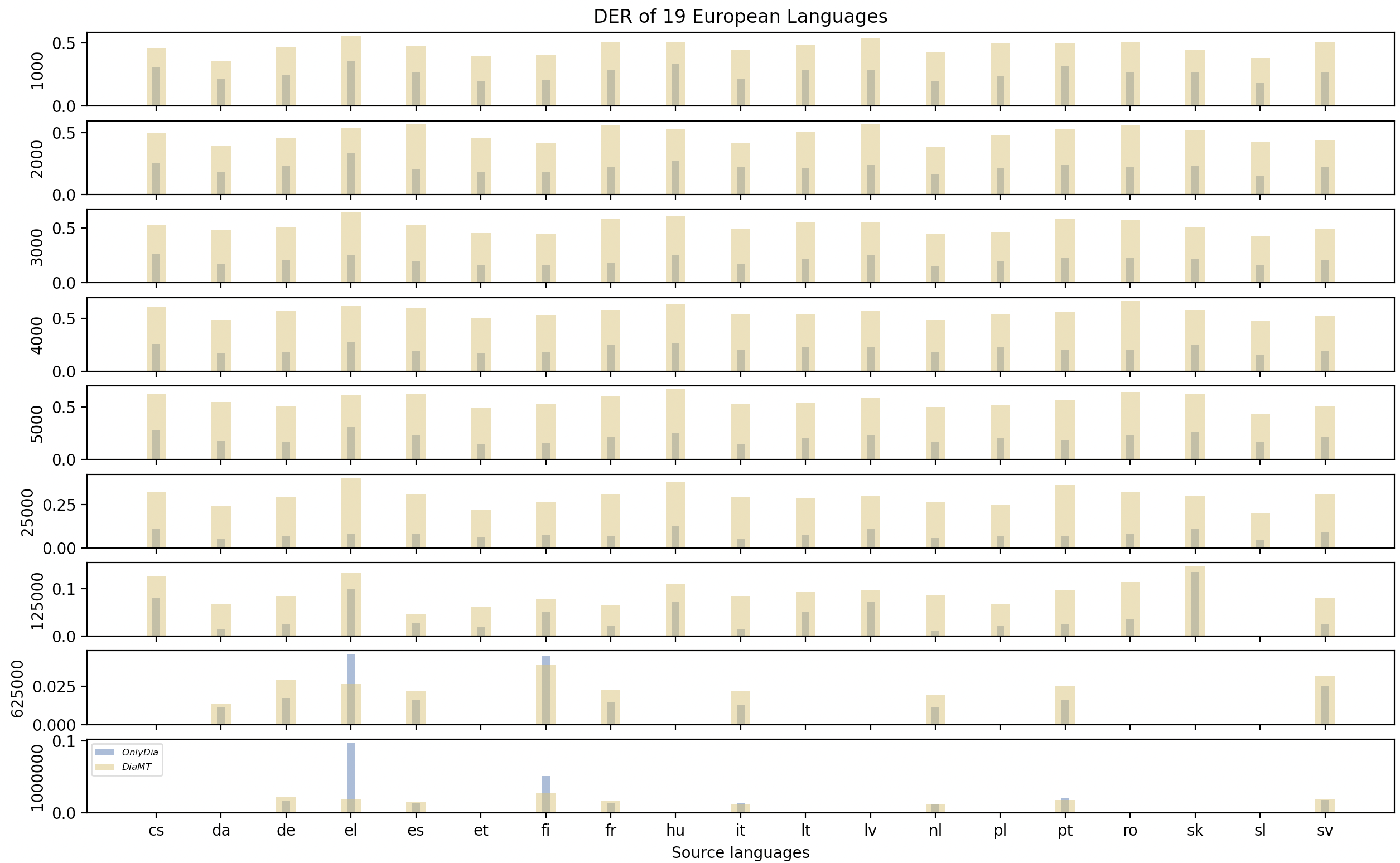}
  \caption{DER comparison between OnlyDia and DiaMT for 19 European languages. Greek (\textit{el}) and Finnish (\textit{fi}) show significant performance gain after adding MT to form a multi-task setting at $1$M train size.}
  \label{fig:bar_der_euro_onlydia_diamt}
 \end{centering}
\end{figure*}

\begin{figure*}[htp!]
\begin{centering}
\includegraphics[scale=0.35]{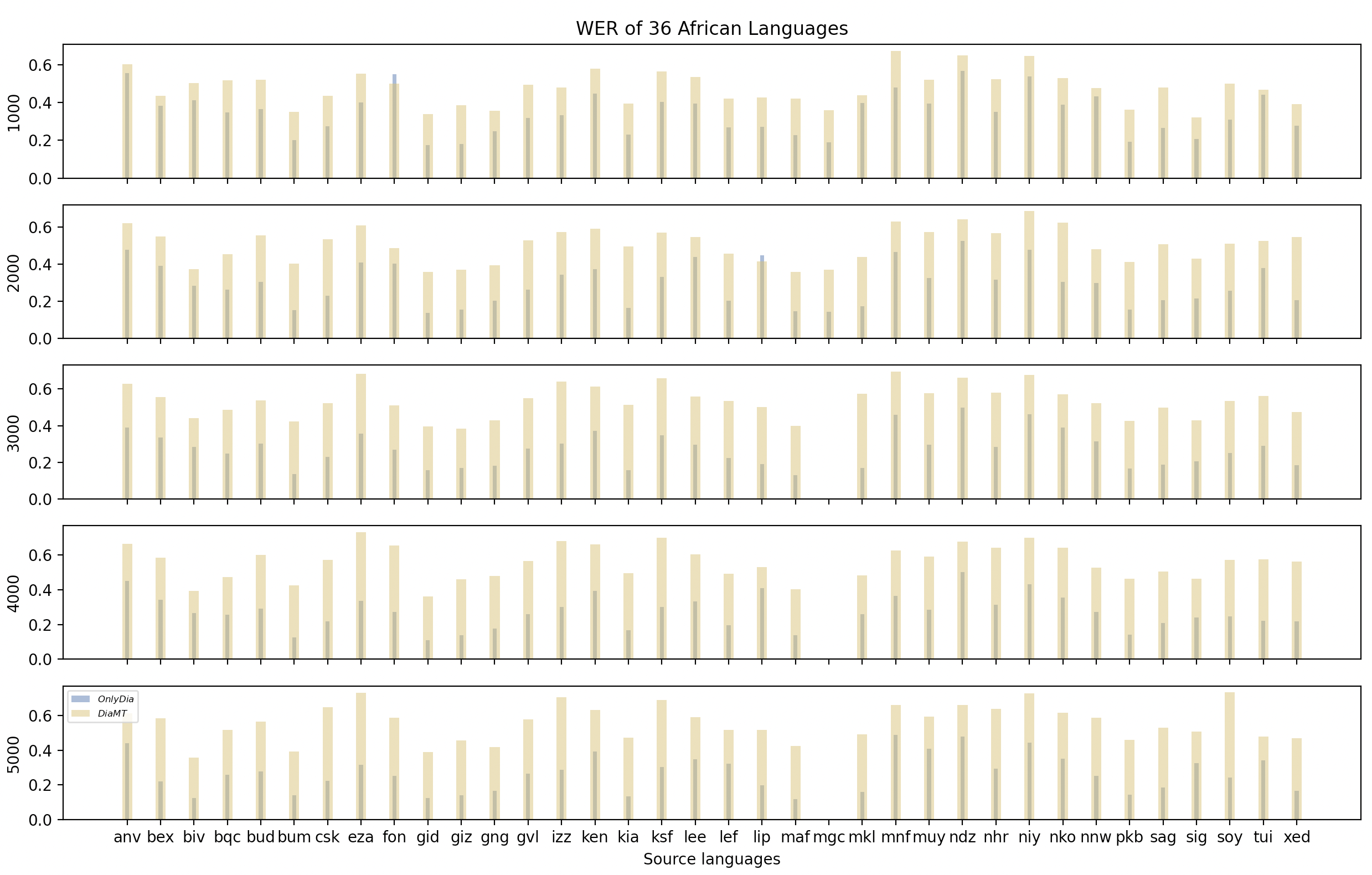}
  \caption{WER comparison between OnlyDia and DiaMT for 36 African languages.}
  \label{fig:bar_wer_afri_onlydia_diamt}
 \end{centering}
\end{figure*}

\begin{figure*}[htp!]
\begin{centering}
\includegraphics[scale=0.35]{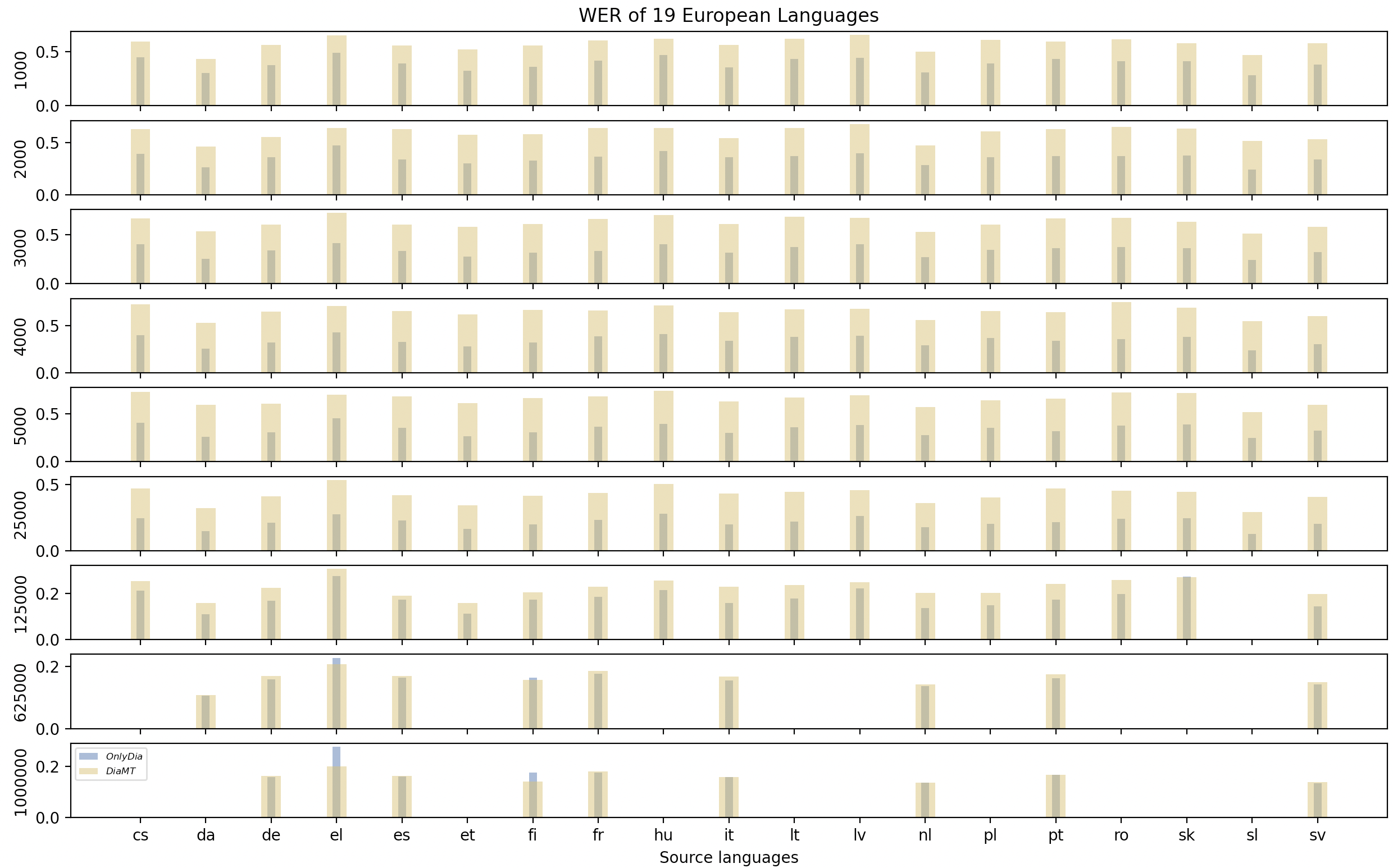}
  \caption{WER comparison between OnlyDia and DiaMT for 19 European languages. Greek (\textit{el}) and Finnish (\textit{fi}) show significant performance gain after adding MT to form a multi-task setting at $1$M train size.}
  \label{fig:bar_wer_euro_onlydia_diamt}
 \end{centering}
\end{figure*}

\clearpage %
\onecolumn
\appendixpage
\addappheadtotoc
\counterwithin{figure}{section}
\counterwithin{table}{section}
There are five sections in the appendix:
\begin{itemize}
    \item Appendix~\ref{sec:miscellaneous} includes    \begin{itemize}
        \item Number of examples for Train/Dev/Test splits for African languages (Table~\ref{tab:dataset_split_datapoints_afri}) and European languages (Table~\ref{tab:dataset_split_datapoints_euro}).
        \item Hyperparameters and software information for the models we train (Table~\ref{tab:model_hyperparameters}).
        \item Set of characters and their diacritical variants for African languages (Table~\ref{tab:chr_variants_afri}) and European languages (Table~\ref{tab:chr_variants_euro}).
        \item Classification of lexical and/or grammatical function for each European language (Table~\ref{tab:euro_diacritics_function}).
                
    \end{itemize}

    \item Appendix~\ref{sec:implementations_DER_WER} includes implementation details of diacritics error rate (DER) and word error rate (WER) metrics for  measuring the performance of diacritization.
    
    \item Appendix~\ref{sec:bar_plots} includes bar plots to demonstrate the comparison between different model settings which are visualization attempts to approach our research questions:
        \begin{itemize}
            \item BLEU scores
            \begin{itemize}
                \item (\textbf{RQ1a.}) DiaMT vs. OnlyMT\textsubscript{undia} for African languages in  Figure~\ref{fig:bar_bleu_afri_diamt_onlymt_undia} and for European languages in Figure~\ref{fig:bar_bleu_euro_diamt_onlymt_undia}
                \item (\textbf{RQ2.}) OnlyMT\textsubscript{dia} vs. OnlyMT\textsubscript{undia} for African languages in Figure~\ref{fig:bar_bleu_afri_onlymt_undia_onlymt_dia} and for European languages in Figure~\ref{fig:bar_bleu_euro_onlymt_undia_onlymt_dia}
            \end{itemize}
            \item DER and WER
            \begin{itemize}
                \item (\textbf{RQ1b.}) DiaMT vs. OnlyDia
                \begin{itemize}
                \item DER for African languages in  Figure~\ref{fig:bar_der_afri_onlydia_diamt} and for European languages in Figure~\ref{fig:bar_der_euro_onlydia_diamt}
                \item WER for African languages in  Figure~\ref{fig:bar_wer_afri_onlydia_diamt} and for European languages in Figure~\ref{fig:bar_wer_euro_onlydia_diamt}
                \end{itemize}
            \end{itemize}
        \end{itemize}

    \item Appendix~\ref{sec:performance_metrics} includes values of metrics (BLEU, DER, WER) to measure the performance of MT and diacritization for all models given different languages at different train sizes; and the percentage change between two different models.
        \begin{itemize}
            \item BLEU scores for every African language (Table~\ref{tab:bleu_afri_exact_values}) and European language (Table~\ref{tab:bleu_euro_exact_values}).
            \item DER and WER for every African language (Table~\ref{tab:der_wer_afri_exact_values}) and European language (Table~\ref{tab:der_wer_euro_exact_values}).
        \end{itemize}
    
   \item Appendix~\ref{sec:complexity_metrics} includes 
   \begin{itemize}
    \item Implementation details of our proposed language-agnostic complexity metrics designed to evaluate the complexity of the diacritical system of any given language.
    \item Correlation analysis of our proposed complexity metrics and diacritization performance metrics (DER, WER) for both African and European languages (Table~\ref{tab:leveled_afri_complexity_performance_correlations} and~\ref{tab:leveled_euro_complexity_performance_correlations}).
    \item The values of complexity metrics for all $55$ included African and European languages at different train sizes (Table~\ref{tab:leveled_dia_complexity_metrics_afri} and~\ref{tab:leveled_dia_complexity_metrics_euro}).
   \end{itemize}
\end{itemize}

\clearpage
\twocolumn
\section{Miscellaneous}\label{sec:miscellaneous}

\begin{table}[h]
\scriptsize
\centering
\begin{tabular}{llrrr}
\toprule
\textbf{Code} &       \textbf{Name} &  \textbf{Train} &  \textbf{Dev} &  \textbf{Test} \\
\midrule
      bex &         JurModo &   4,938 &  617 &   618 \\
      fon &             Fon &   4,948 &  619 &   619 \\
      mkl &          Mokole &   4,930 &  616 &   617 \\
      mnf &         Mundani &   4,921 &  615 &   616 \\
      bud &    Bassar, Ntcham &   4,950 &  619 &   619 \\
      eza &            Ezaa &   4,962 &  620 &   621 \\
      sig &         Paasaal &   4,932 &  616 &   617 \\
      bqc &            Boko &   4,956 &  619 &   620 \\
      kia &             Kim &   4,963 &  620 &   621 \\
      soy &          Miyobe &   4,957 &  620 &   620 \\
      nnw &    Southern Nuni &   4,928 &  616 &   616 \\
      sag &           Sango &   4,964 &  620 &   621 \\
      csk &        JolaKasa &   4,964 &  621 &   621 \\
      izz &            Izii &   4,964 &  621 &   621 \\
      bum &            Bulu &   4,964 &  620 &   621 \\
      gvl &           Gulay &   4,964 &  621 &   621 \\
      ndz &           Ndogo &   4,959 &  620 &   620 \\
      lip &         Sekpele &   4,934 &  617 &   617 \\
      ken &         Kenyang &   4,960 &  620 &   621 \\
      gid &           Gidar &   4,956 &  620 &   620 \\
      gng &         Ngangam &   4,853 &  607 &   607 \\
      muy &          Muyang &   4,952 &  619 &   619 \\
      niy &           Ngiti &   4,964 &  621 &   621 \\
      xed &             Hdi &   4,959 &  620 &   620 \\
      anv &           Denya &   4,958 &  620 &   620 \\
      lee &           Lyele &   4,939 &  617 &   618 \\
      ksf &           Bafia &   4,964 &  620 &   621 \\
      pkb &          Pokomo &   4,936 &  617 &   617 \\
      nko &          Nkonya &   4,930 &  616 &   617 \\
      lef &          Lelemi &   4,938 &  617 &   618 \\
      nhr &            Naro &   4,952 &  619 &   620 \\
      mgc &        Morokodo &   2,124 &  266 &   266 \\
      biv & Southern Birifor &   4,964 &  620 &   621 \\
      maf &            Mafa &   4,964 &  621 &   621 \\
      giz &     South Giziga &   4,964 &  621 &   621 \\
      tui &          Tupuri &   4,961 &  620 &   621 \\
\bottomrule
\end{tabular}

\caption{The number of examples in Train/Dev/Test splits for African languages.}  \label{tab:dataset_split_datapoints_afri}
\end{table}
\normalsize

\begin{table}[htp!]
\scriptsize
\centering

\begin{tabular}{llr}
\toprule
\textbf{Code} &       \textbf{Name} &  \textbf{Train}  \\
\midrule
       cs &      Czech &       125,000  \\
       da &     Danish &       625,000  \\
       de &     German &      1,000,000  \\
       el &      Greek &      1,000,000  \\
       es &    Spanish &      1,000,000  \\
       et &   Estonian &       125,000  \\
       fi &    Finnish &      1,000,000  \\
       fr &     French &      1,000,000  \\
       hu &  Hungarian &       125,000 \\
       it &    Italian &      1,000,000 \\
       lt & Lithuanian &       125,000 \\
       lv &    Latvian &       125,000 \\
       nl &      Dutch &      1,000,000 \\
       pl &     Polish &       125,000 \\
       pt & Portuguese &      1,000,000 \\
       ro &   Romanian &       125,000 \\
       sk &     Slovak &       125,000 \\
       sl &  Slovenian &        25,000 \\
       sv &    Swedish &      1,000,000 \\
\bottomrule
\end{tabular}

\caption{The number of examples in Train split for European languages. Dev and Test have 1,500 datapoints for all languages.}  \label{tab:dataset_split_datapoints_euro}
\end{table}
\normalsize
\begin{table}[h]
\scriptsize
\centering
\begin{tabular}{cc}
\toprule
\textbf{Hyperparamter} & \textbf{Value}  \\
\midrule
Encoder \#layers & 6 \\
Encoder \#heads & 8 \\
Encoder embedding dimensions & 256 \\
Encoder FFN dimension & 1024 \\
\midrule
Decoder \#layers & 6 \\
Decoder \#heads & 8 \\
Decoder embedding dimensions & 256 \\
Decoder FFN dimension & 1024 \\
\midrule
Dropout rate & 0.2 \\
Batch size & 15 \\
Beam size & 6 \\
Optimizer & Adam~\cite{kingma2017adam} \\
\midrule
Software & Fairseq \\
Version & v0.10.2 \\
License & MIT License \\
\bottomrule
\end{tabular}

\caption{Hyperparameters and software information for our transformer models. The estimated GPU hours to complete the experiments (including those taken  during the development stage) is $7500$. The link for Fairseq software is \url{https://github.com/facebookresearch/fairseq}. Our use is consistent with Fairseq's intended use, based on its license.}  \label{tab:model_hyperparameters}
\end{table}
\normalsize

\onecolumn

\begin{landscape}
\vspace*{\fill} %
\begin{center}

\begin{table*}[h!]

    \begin{tabular}{c}
    \begin{minipage}{1\textwidth}
      \includegraphics[scale=0.43]{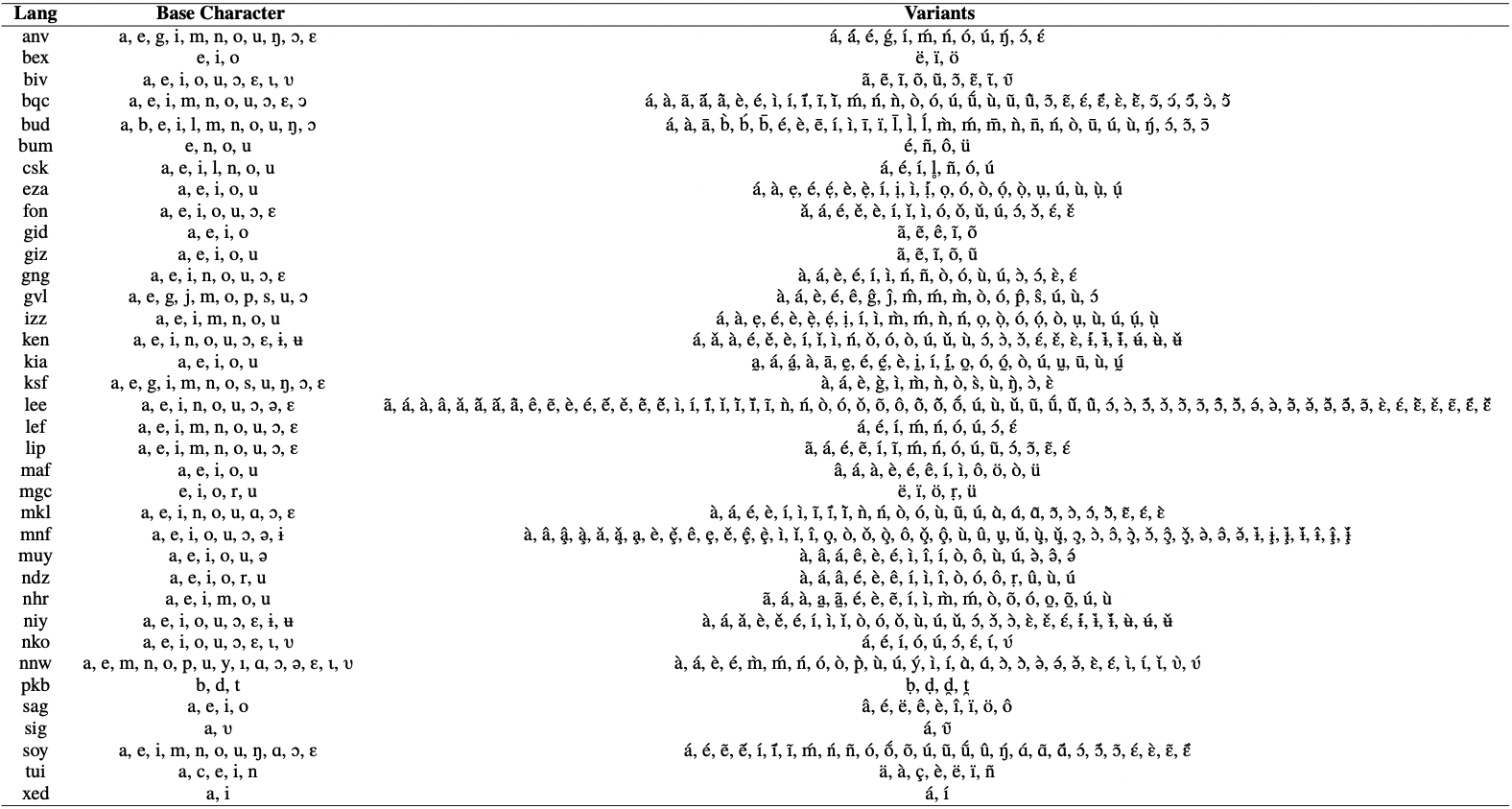}
    \end{minipage}
    \end{tabular}
    \caption{The base and diacritized forms occurred in the dataset of each included African language. The list for each language may not be exhaustive.}
    \label{tab:chr_variants_afri}

\end{table*}

\end{center}
\vspace*{\fill} %
\end{landscape}

\begin{landscape}
\thispagestyle{empty}
\vspace*{\fill} 
\begin{center}

\begin{table}[h!]
    \centering 
    \begin{adjustbox}{center}
        \includegraphics[scale=0.39]{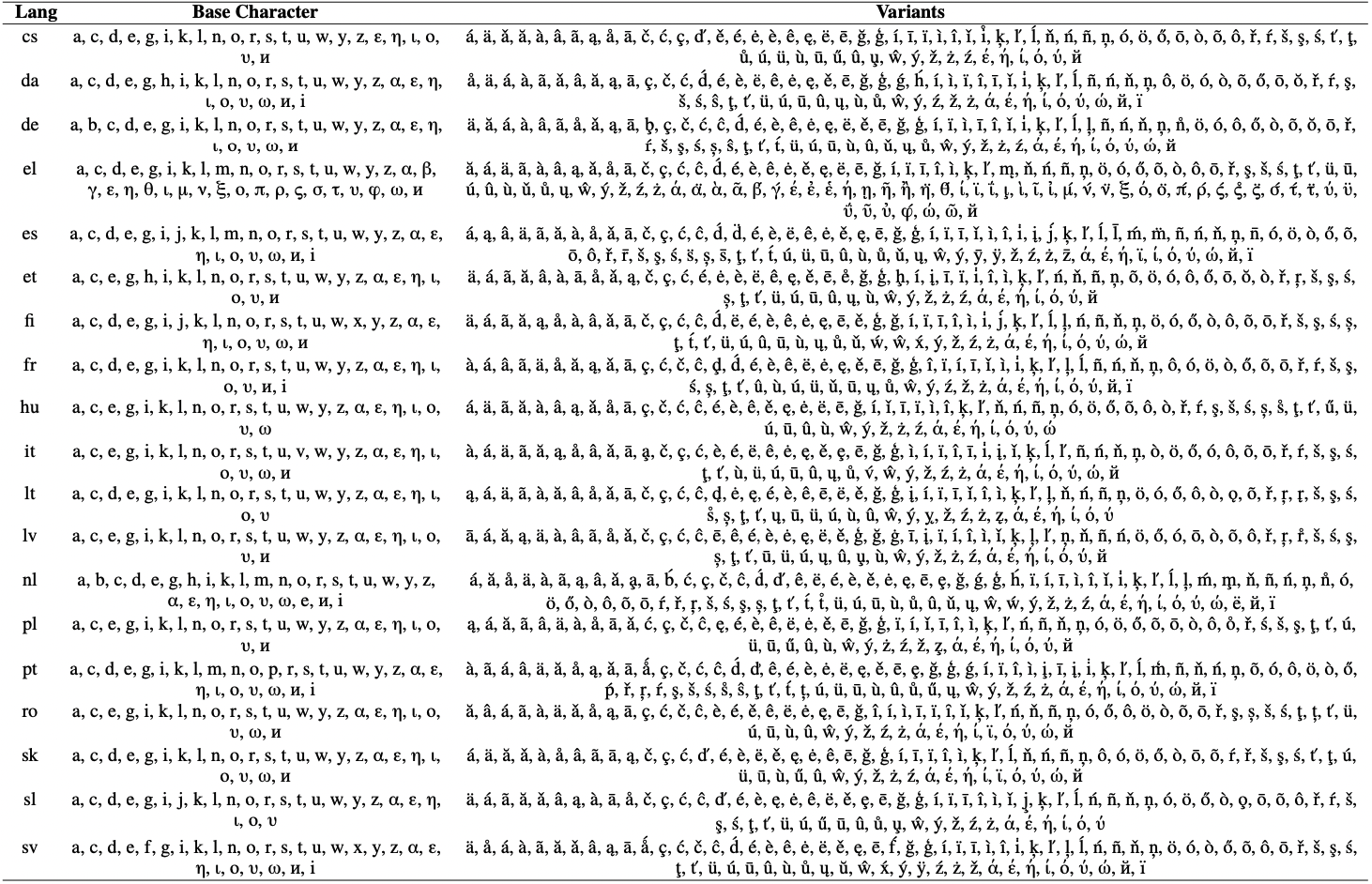}
    \end{adjustbox}
    \caption{The base and diacritized forms occurred in the dataset of each included European language. Note that each of them may contain characters in other language (for example, in Spanish dataset, there can be French word) and therefore the base characters and variants may not be of that single language.}
    \label{tab:chr_variants_euro}
\end{table}

\end{center}
\vspace*{\fill}
\end{landscape}

\begin{table*}[h!]

    \begin{tabular}{c}
    \begin{minipage}{1\textwidth}
      \includegraphics[scale=0.43]{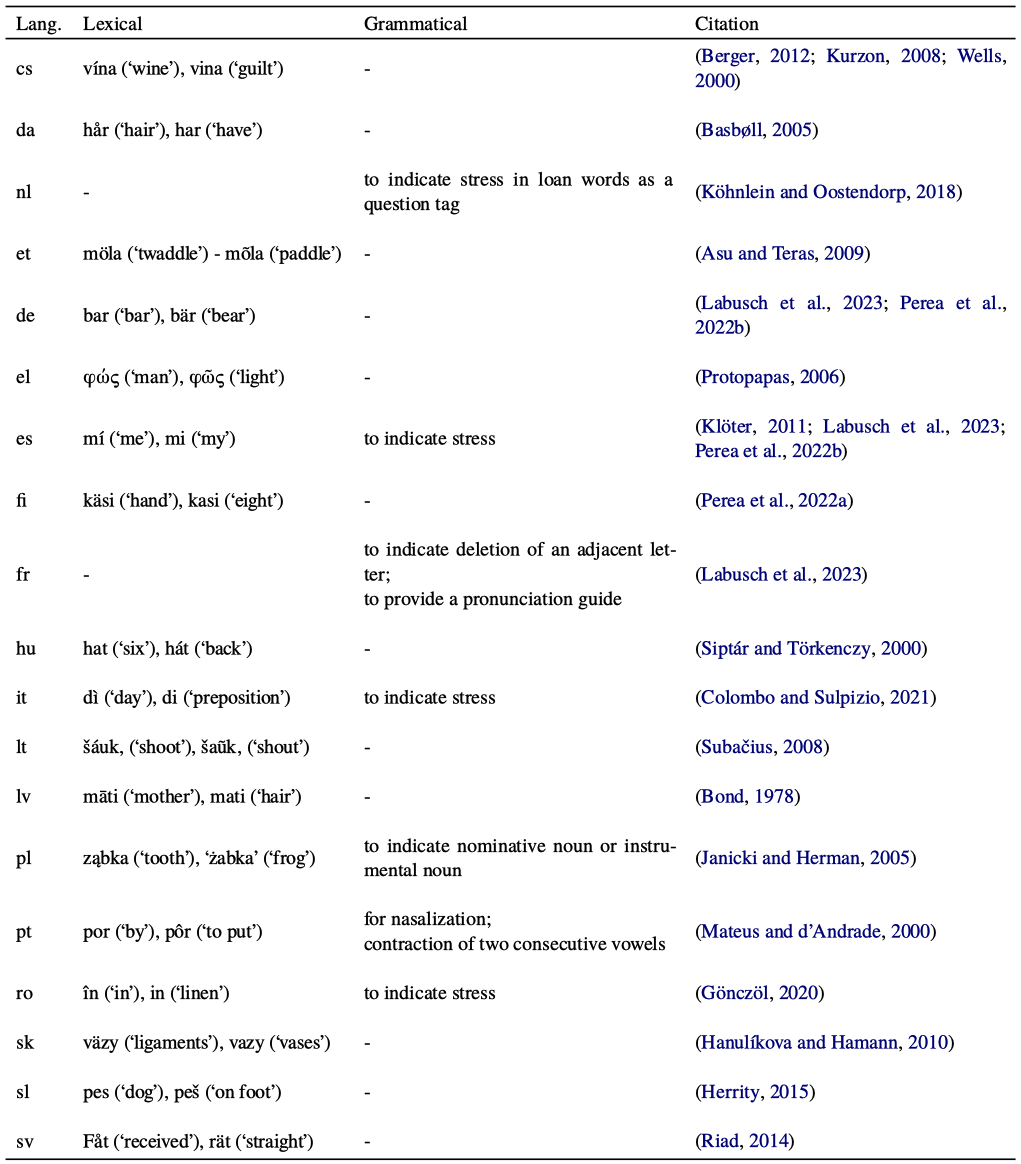}
    \end{minipage}
    \end{tabular}
    \caption{Classification of the function(s) (lexical and/or grammatical) for each European language. For lexical function, we show minimal pairs where an alternation in diacritic changes the meaning to demonstrate that removing diacritic(s) can produce ambiguity. For Lithuania \textit{(lt)},  Polish \textit{(pl), and Swedish (\textit{sv})}, we show near minimal pairs. Both minimal pairs and near minimal pairs show that the undiacritized form poses ambiguity as there are more than one form to diacritize it. For grammatical function, we indicate the grammatical role(s) the diacritical system has in the language.}\label{tab:euro_diacritics_function}

\end{table*}

\clearpage
\onecolumn
\section{Implementations of Diacritization Error Rate (DER) and Word Error Rate (WER)}\label{sec:implementations_DER_WER}

In the field of diacritization system development, two primary methodologies emerge: sequence labeling and sequence-to-sequence modeling~\cite{schlippe-etal-2008-diacritization, hamed2017survey}. In our research, we opt for the latter as our research question 1 (see Section~\ref{sec:intro} for details) requires the model to be able to perform both diacritization and machine translation tasks. However, employing sequence-to-sequence modeling presents challenges, particularly regarding alignment and potentially unequal input-output lengths~\cite{alqahtani-etal-2019-efficient, abandah2020accurate}.

Previous studies employing encoder-decoder architectures for Arabic diacritization have leveraged Arabic linguistic rules to compute these metrics~\cite{fadel2019arabic, qin-etal-2021-improving, thompson-alshehri-2022-improving}. To address the aforementioned issues, \citet{thompson-alshehri-2022-improving} employ Arabic linguistic rules to constrain the decoder and guide the generation of subsequent tokens. However, the proposed decoding constraints cannot be directly applied, given that (1) the included $55$ languages are non-Arabic (2) the potential for multiple diacritics to be attached to a single character in certain languages (see Table~\ref{tab:chr_variants_afri}). 

Despite our comprehensive search, we were unable to locate implementation details for DER and WER in prior works that adopt a sequence-to-sequence approach~\cite{fadel2019arabic, qin-etal-2021-improving, thompson-alshehri-2022-improving, mubarak-etal-2019-highly}. Therefore, we have developed our own DER and WER computation methods, as in Algorithms~\ref{alg:DER} and~\ref{alg:WER}. Our approach adheres to the definitions of DER and WER  established by \citet{Abandah2015}.

In computing DER, we exclude words that exceed the length of the input sequence, while penalizing characters exceeding the length of a certain word, complying with DER's focus on character-level analysis. By restricting the comparison to characters within each word instead of directly comparing a predicted sequence to a gold standard sequence, we ensure a fairer evaluation. This approach maintains evaluation integrity when predictions align reasonably with the input, and prevents over-pessimistic assessments when deviations occur. Regarding WER, we penalize words surpassing the input sequence's length, reflecting WER's word-level focus.

\begin{algorithm*}
\caption{Diacritization Error Rate (DER)}\label{alg:DER}
\footnotesize
\begin{algorithmic}[1]

\Require
    \Statex Golds is a list of n gold standard sequences.
    \Statex Preds is a list of n post-processed predicted sequences (See Section~\ref{sec:postprocessing} for details).

\State $\text{incorrect} \gets 0$
    \State $\text{correct} \gets 0$
    \For{$i$ \textbf{in} $[0, n-1]$} 
        \State $\text{gold\_words} \gets \text{\text{Golds}[i].split(' ')}$
        \State $\text{pred\_words} \gets \text{{Preds}[i].split(' ')}$        
        \For{$j$ \textbf{in} $[0, \text{min(len(pred\_words), len(gold\_words))} - 1]$} 
            \State $\text{gold\_word} \gets \text{gold\_words}[j]$
            \State $\text{pred\_word} \gets \text{pred\_words}[j]$
            \State $\text{incorrect} \gets \text{incorrect} + \text{abs}(\text{len}(\text{pred\_word}) - \text{len}(\text{gold\_word}))$ 
            \For{$k$ \textbf{in} $[0, \min(\text{len}(\text{pred\_word}), \text{len}(\text{gold\_word})) - 1]$}
                \If{$\text{pred\_word}[k] == \text{gold\_word}[k]$}
                    \State $\text{correct} \gets \text{correct} + 1$
                \Else
                    \State $\text{incorrect} \gets \text{incorrect} + 1$
                \EndIf
            \EndFor
        \EndFor
    \EndFor
    \State $\text{DER} \gets \text{incorrect} / (\text{incorrect} + \text{correct})$

\end{algorithmic}
\end{algorithm*}
\begin{algorithm*}
\caption{Word Error Rate (WER)}\label{alg:WER}
\footnotesize
\begin{algorithmic}[1]

\Require
    \Statex Golds is a list of n gold standard sequences.
    \Statex Preds is a list of n post-processed predicted sequences (See Section~\ref{sec:postprocessing} for details).

\State $\text{incorrect} \gets 0$
\State $\text{correct} \gets 0$
\For{$i$ \textbf{in} $[0, n-1]$} 
    \State $\text{gold\_words} \gets \text{\text{Golds}[i].split(' ')}$
    \State $\text{pred\_words} \gets \text{{Preds}[i].split(' ')}$
    \State $\text{incorrect} \gets \text{incorrect} + \text{abs}(\text{len}(\text{gold\_words}) - \text{len}(\text{pred\_words}))$
    \For{$j$ \textbf{in} $[0, \text{min(len(pred\_words), len(gold\_words))} - 1]$}
        \If{$\text{gold\_words}[j] == \text{pred\_words}[j]$}
            \State $\text{correct} \gets \text{correct} + 1$
        \Else
            \State $\text{incorrect} \gets \text{incorrect} + 1$
        \EndIf
    \EndFor
    
\EndFor
\State $ \text{WER} \gets \text{incorrect} / (\text{incorrect} + \text{correct})$

\end{algorithmic}
\end{algorithm*}
\clearpage
\section{Bar Plots}\label{sec:bar_plots}

\clearpage
\section{Performance Metrics}\label{sec:performance_metrics}

\begin{footnotesize}
\begin{longtable}{ccrrrrr}
\caption{BLEU scores of 36 African languages in 5 train sizes produced by 3 models. The highest BLEU score out of the 3 models are boldfaced. DM, OM\textsubscript{u}, OM\textsubscript{d} are shorthands for DiaMT, OnlyMT\textsubscript{undia}, and OnlyMT\textsubscript{d}, respectively. pc(m1, m2) is the percentage change of model m1 over model m2. The higher the percentage change, the better the model m1 is compared to model m2.} \label{tab:bleu_afri_exact_values} \\
\toprule
Size & Lang & DiaMT & OnlyMT\textsubscript{undia} & OnlyMT\textsubscript{dia} & pc(DM, OM\textsubscript{u}) & pc(OM\textsubscript{u}, OM\textsubscript{d}) \\

\midrule
\endfirsthead
\caption[]{BLEU scores of 36 African languages in 5 train sizes produced by 3 models. The highest BLEU score out of the 3 models are boldfaced. DM, OM\textsubscript{u}, OM\textsubscript{d} are shorthands for DiaMT, OnlyMT\textsubscript{undia}, and OnlyMT\textsubscript{d}, respectively. pc(m1, m2) is the percentage change of model m1 over model m2. The higher the percentage change, the better the model m1 is compared to model m2.} \\
\toprule
Size & Lang & DiaMT & OnlyMT\textsubscript{undia} & OnlyMT\textsubscript{dia} & pc(DM, OM\textsubscript{u}) & pc(OM\textsubscript{u}, OM\textsubscript{d}) \\
\midrule
\endhead
\midrule
\multicolumn{7}{r}{Continued on next page} \\
\midrule
\endfoot
\bottomrule
\endlastfoot
\multirow[c]{36}{*}{1k} & bex & \textbf{2.971} & 1.122 & 1.263 & +164.78\% & -11.14\% \\
 & fon & \textbf{2.729} & 1.764 & 0.799 & +54.64\% & +120.74\% \\
 & mkl & \textbf{1.616} & 1.119 & 1.428 & +44.42\% & -21.63\% \\
 & mnf & \textbf{3.086} & 0.792 & 0.474 & +289.48\% & +66.99\% \\
 & bud & \textbf{2.473} & 1.281 & 1.087 & +93.08\% & +17.88\% \\
 & eza & \textbf{2.093} & 0.637 & 0.368 & +228.72\% & +72.91\% \\
 & sig & \textbf{1.678} & 0.913 & 0.898 & +83.68\% & +1.68\% \\
 & bqc & \textbf{2.725} & 1.330 & 0.994 & +104.82\% & +33.86\% \\
 & kia & \textbf{2.376} & 1.313 & 1.002 & +81.02\% & +30.97\% \\
 & soy & \textbf{1.919} & 0.986 & 0.626 & +94.49\% & +57.65\% \\
 & nnw & \textbf{2.618} & 1.242 & 1.545 & +110.73\% & -19.61\% \\
 & sag & \textbf{2.232} & 1.273 & 1.467 & +75.38\% & -13.26\% \\
 & csk & \textbf{2.052} & 0.486 & 0.318 & +322.05\% & +53.06\% \\
 & izz & \textbf{1.492} & 0.963 & 0.498 & +54.91\% & +93.40\% \\
 & bum & \textbf{2.163} & 1.027 & 1.039 & +110.68\% & -1.17\% \\
 & gvl & \textbf{2.097} & 0.791 & 0.738 & +165.00\% & +7.17\% \\
 & ndz & \textbf{1.724} & 1.420 & 1.451 & +21.40\% & -2.13\% \\
 & lip & \textbf{2.027} & 0.081 & 0.944 & +2410.38\% & -91.45\% \\
 & ken & \textbf{2.523} & 1.216 & 1.473 & +107.39\% & -17.40\% \\
 & gid & \textbf{2.488} & 0.678 & 0.463 & +267.12\% & +46.41\% \\
 & gng & \textbf{2.471} & 0.918 & 0.150 & +169.12\% & +513.48\% \\
 & muy & \textbf{1.557} & 0.494 & 0.565 & +215.51\% & -12.72\% \\
 & niy & \textbf{1.522} & 0.708 & 0.458 & +114.99\% & +54.57\% \\
 & xed & 1.619 & 1.202 & \textbf{1.692} & +34.71\% & -28.96\% \\
 & anv & \textbf{2.105} & 1.397 & 1.483 & +50.68\% & -5.82\% \\
 & lee & \textbf{2.045} & 0.351 & 0.445 & +482.62\% & -21.08\% \\
 & ksf & \textbf{2.276} & 0.099 & 0.554 & +2197.03\% & -82.13\% \\
 & pkb & \textbf{2.642} & 1.279 & 1.286 & +106.59\% & -0.59\% \\
 & nko & \textbf{3.389} & 1.083 & 1.261 & +212.90\% & -14.06\% \\
 & lef & \textbf{2.243} & 1.314 & 1.516 & +70.76\% & -13.32\% \\
 & nhr & \textbf{2.142} & 1.305 & 1.191 & +64.11\% & +9.62\% \\
 & mgc & \textbf{5.619} & 3.609 & 2.948 & +55.70\% & +22.42\% \\
 & biv & \textbf{2.861} & 1.424 & 1.338 & +100.83\% & +6.42\% \\
 & maf & \textbf{1.780} & 0.942 & 0.058 & +89.02\% & +1516.17\% \\
 & giz & \textbf{1.694} & 0.960 & 1.086 & +76.35\% & -11.56\% \\
 & tui & \textbf{1.953} & 0.465 & 0.413 & +320.20\% & +12.41\% \\
\hdashline
\multirow[c]{24}{*}{2k} & bex & \textbf{2.639} & 2.173 & 1.820 & +21.44\% & +19.39\% \\
 & fon & \textbf{3.713} & 1.803 & 1.918 & +105.93\% & -5.99\% \\
 & mkl & \textbf{2.966} & 1.248 & 1.436 & +137.68\% & -13.13\% \\
 & mnf & \textbf{3.086} & 1.939 & 1.645 & +59.17\% & +17.88\% \\
 & bud & \textbf{3.226} & 2.449 & 2.156 & +31.71\% & +13.59\% \\
 & eza & \textbf{3.129} & 2.272 & 1.984 & +37.71\% & +14.55\% \\
 & sig & \textbf{3.341} & 1.970 & 2.209 & +69.56\% & -10.83\% \\
 & bqc & \textbf{3.344} & 1.553 & 1.561 & +115.38\% & -0.53\% \\
 & kia & \textbf{3.251} & 1.997 & 2.041 & +62.81\% & -2.15\% \\
 & soy & \textbf{2.733} & 1.288 & 1.472 & +112.23\% & -12.55\% \\
 & nnw & \textbf{3.161} & 2.058 & 2.351 & +53.58\% & -12.46\% \\
 & sag & \textbf{3.548} & 2.568 & 2.592 & +38.13\% & -0.90\% \\
 & csk & \textbf{3.128} & 1.392 & 1.608 & +124.68\% & -13.41\% \\
 & izz & \textbf{2.721} & 1.914 & 1.595 & +42.16\% & +20.01\% \\
 & bum & \textbf{2.191} & 1.123 & 1.781 & +95.19\% & -36.96\% \\
 & gvl & \textbf{2.471} & 1.470 & 1.813 & +68.08\% & -18.91\% \\
 & ndz & \textbf{2.490} & 1.992 & 2.340 & +25.00\% & -14.84\% \\
 & lip & \textbf{2.903} & 2.146 & 2.319 & +35.30\% & -7.47\% \\
 & ken & \textbf{3.408} & 2.009 & 1.707 & +69.64\% & +17.70\% \\
 & gid & \textbf{2.857} & 1.974 & 1.868 & +44.70\% & +5.69\% \\
 & gng & \textbf{3.953} & 1.932 & 2.073 & +104.56\% & -6.81\% \\
 & muy & \textbf{3.074} & 1.779 & 1.657 & +72.78\% & +7.35\% \\
 & niy & \textbf{3.146} & 1.922 & 1.835 & +63.68\% & +4.75\% \\
 & xed & \textbf{2.683} & 1.910 & 1.526 & +40.46\% & +25.17\% \\
\multirow[c]{12}{*}{2k} & anv & \textbf{2.176} & 1.626 & 0.899 & +33.85\% & +80.88\% \\
 & lee & \textbf{2.692} & 1.659 & 1.840 & +62.25\% & -9.83\% \\
 & ksf & \textbf{2.882} & 1.187 & 1.723 & +142.90\% & -31.14\% \\
 & pkb & \textbf{3.079} & 2.321 & 1.443 & +32.64\% & +60.83\% \\
 & nko & \textbf{3.536} & 2.284 & 2.415 & +54.82\% & -5.43\% \\
 & lef & \textbf{2.626} & 1.999 & 2.212 & +31.36\% & -9.62\% \\
 & nhr & \textbf{2.827} & 1.813 & 1.528 & +55.96\% & +18.60\% \\
 & mgc & \textbf{7.822} & 4.517 & 3.818 & +73.15\% & +18.33\% \\
 & biv & \textbf{3.348} & 1.612 & 2.006 & +107.67\% & -19.62\% \\
 & maf & \textbf{2.819} & 1.198 & 1.199 & +135.35\% & -0.14\% \\
 & giz & \textbf{3.153} & 1.771 & 1.811 & +78.07\% & -2.21\% \\
 & tui & \textbf{2.220} & 1.203 & 1.088 & +84.50\% & +10.57\% \\
\hdashline
\multirow[c]{35}{*}{3k} & bex & \textbf{3.889} & 3.168 & 2.867 & +22.75\% & +10.51\% \\
 & fon & \textbf{3.851} & 2.645 & 2.684 & +45.60\% & -1.45\% \\
 & mkl & \textbf{3.478} & 2.138 & 2.529 & +62.67\% & -15.45\% \\
 & mnf & \textbf{2.958} & 1.627 & 2.308 & +81.87\% & -29.54\% \\
 & bud & \textbf{3.638} & 2.669 & 2.261 & +36.32\% & +18.04\% \\
 & eza & \textbf{3.761} & 3.065 & 2.930 & +22.73\% & +4.59\% \\
 & sig & \textbf{3.431} & 2.136 & 2.629 & +60.60\% & -18.76\% \\
 & bqc & \textbf{3.910} & 1.785 & 2.156 & +119.03\% & -17.18\% \\
 & kia & \textbf{3.921} & 2.065 & 2.785 & +89.90\% & -25.86\% \\
 & soy & \textbf{2.982} & 1.995 & 1.759 & +49.46\% & +13.45\% \\
 & nnw & \textbf{3.582} & 2.364 & 3.074 & +51.55\% & -23.11\% \\
 & sag & \textbf{3.900} & 3.252 & 2.614 & +19.91\% & +24.42\% \\
 & csk & \textbf{3.181} & 1.812 & 1.906 & +75.60\% & -4.93\% \\
 & izz & \textbf{2.949} & 2.375 & 2.192 & +24.19\% & +8.33\% \\
 & bum & \textbf{2.985} & 1.631 & 2.204 & +83.06\% & -26.00\% \\
 & gvl & \textbf{3.233} & 1.691 & 2.011 & +91.22\% & -15.94\% \\
 & ndz & 2.746 & 2.846 & \textbf{3.132} & -3.53\% & -9.11\% \\
 & lip & \textbf{2.934} & 2.598 & 2.849 & +12.92\% & -8.82\% \\
 & ken & \textbf{3.410} & 2.605 & 2.641 & +30.90\% & -1.39\% \\
 & gid & \textbf{3.514} & 2.372 & 2.750 & +48.13\% & -13.75\% \\
 & gng & \textbf{3.946} & 2.627 & 3.113 & +50.19\% & -15.61\% \\
 & muy & \textbf{3.594} & 2.624 & 2.563 & +36.96\% & +2.36\% \\
 & niy & \textbf{3.514} & 1.844 & 2.196 & +90.54\% & -16.02\% \\
 & xed & \textbf{2.606} & 2.433 & 2.356 & +7.10\% & +3.29\% \\
 & anv & \textbf{2.699} & 2.233 & 2.108 & +20.84\% & +5.95\% \\
 & lee & \textbf{3.343} & 3.221 & 2.836 & +3.81\% & +13.57\% \\
 & ksf & \textbf{3.247} & 2.011 & 1.997 & +61.50\% & +0.70\% \\
 & pkb & \textbf{3.865} & 2.923 & 2.162 & +32.23\% & +35.23\% \\
 & nko & \textbf{3.834} & 2.592 & 2.915 & +47.90\% & -11.07\% \\
 & lef & \textbf{3.230} & 2.124 & 3.039 & +52.07\% & -30.11\% \\
 & nhr & \textbf{3.728} & 2.752 & 2.436 & +35.44\% & +13.00\% \\
 & biv & \textbf{3.851} & 2.690 & 2.772 & +43.13\% & -2.96\% \\
 & maf & \textbf{2.893} & 1.946 & 1.674 & +48.64\% & +16.27\% \\
 & giz & \textbf{3.372} & 2.657 & 2.103 & +26.91\% & +26.33\% \\
 & tui & \textbf{2.470} & 2.068 & 2.150 & +19.46\% & -3.82\% \\
\hdashline
\multirow[c]{13}{*}{4k} & bex & \textbf{3.735} & 3.668 & 3.243 & +1.83\% & +13.10\% \\
 & fon & \textbf{4.112} & 3.615 & 3.347 & +13.75\% & +8.03\% \\
 & mkl & \textbf{2.948} & 2.441 & 2.341 & +20.73\% & +4.27\% \\
 & mnf & \textbf{3.598} & 2.462 & 2.418 & +46.14\% & +1.80\% \\
 & bud & \textbf{3.626} & 3.408 & 3.510 & +6.39\% & -2.89\% \\
 & eza & \textbf{3.648} & 3.073 & 3.080 & +18.71\% & -0.22\% \\
 & sig & \textbf{3.686} & 3.174 & 3.273 & +16.11\% & -3.03\% \\
 & bqc & \textbf{3.352} & 2.806 & 3.105 & +19.46\% & -9.61\% \\
 & kia & \textbf{3.277} & 2.702 & 2.899 & +21.27\% & -6.79\% \\
 & soy & \textbf{2.867} & 2.444 & 2.282 & +17.32\% & +7.09\% \\
 & nnw & \textbf{3.686} & 2.800 & 3.350 & +31.65\% & -16.43\% \\
 & sag & \textbf{4.014} & 3.164 & 3.817 & +26.86\% & -17.12\% \\
 & csk & \textbf{3.625} & 2.565 & 2.670 & +41.29\% & -3.92\% \\
\multirow[c]{22}{*}{4k} & izz & \textbf{2.784} & 2.559 & 2.611 & +8.79\% & -2.00\% \\
 & bum & \textbf{3.012} & 2.528 & 2.305 & +19.13\% & +9.70\% \\
 & gvl & \textbf{3.105} & 2.727 & 2.422 & +13.86\% & +12.59\% \\
 & ndz & \textbf{3.850} & 3.478 & 3.718 & +10.68\% & -6.44\% \\
 & lip & \textbf{3.377} & 3.165 & 3.243 & +6.68\% & -2.39\% \\
 & ken & \textbf{3.631} & 3.265 & 2.944 & +11.20\% & +10.92\% \\
 & gid & \textbf{3.696} & 2.547 & 3.012 & +45.10\% & -15.43\% \\
 & gng & \textbf{4.494} & 3.472 & 3.172 & +29.46\% & +9.45\% \\
 & muy & 2.711 & 2.705 & \textbf{3.080} & +0.20\% & -12.17\% \\
 & niy & \textbf{3.782} & 2.865 & 3.240 & +31.99\% & -11.56\% \\
 & xed & 3.266 & \textbf{3.338} & 2.818 & -2.14\% & +18.46\% \\
 & anv & \textbf{2.805} & 2.016 & 2.413 & +39.09\% & -16.43\% \\
 & lee & \textbf{3.471} & 3.232 & 3.224 & +7.39\% & +0.25\% \\
 & ksf & \textbf{3.322} & 2.719 & 2.905 & +22.16\% & -6.40\% \\
 & pkb & \textbf{3.380} & 2.832 & 3.221 & +19.34\% & -12.07\% \\
 & nko & 3.786 & 3.414 & \textbf{3.809} & +10.89\% & -10.36\% \\
 & lef & \textbf{3.579} & 2.982 & 3.431 & +20.03\% & -13.09\% \\
 & nhr & \textbf{3.665} & 3.201 & 3.017 & +14.47\% & +6.10\% \\
 & biv & \textbf{4.219} & 3.331 & 3.394 & +26.66\% & -1.86\% \\
 & maf & \textbf{3.332} & 2.222 & 2.375 & +49.93\% & -6.42\% \\
 & giz & \textbf{3.624} & 2.954 & 2.915 & +22.70\% & +1.31\% \\
 & tui & \textbf{3.264} & 2.799 & 2.983 & +16.63\% & -6.18\% \\
\hdashline
\multirow[c]{35}{*}{5k} & bex & 3.883 & 3.368 & \textbf{3.914} & +15.30\% & -13.97\% \\
 & fon & \textbf{4.163} & 4.080 & 4.013 & +2.04\% & +1.67\% \\
 & mkl & \textbf{3.578} & 2.955 & 2.955 & +21.10\% & -0.01\% \\
 & mnf & \textbf{3.670} & 3.012 & 3.035 & +21.82\% & -0.74\% \\
 & bud & \textbf{3.842} & 3.554 & 3.437 & +8.10\% & +3.42\% \\
 & eza & \textbf{3.584} & 3.393 & 3.190 & +5.65\% & +6.37\% \\
 & sig & \textbf{3.589} & 2.898 & 3.247 & +23.85\% & -10.74\% \\
 & bqc & 3.176 & 3.073 & \textbf{3.575} & +3.36\% & -14.05\% \\
 & kia & \textbf{3.560} & 3.235 & 3.158 & +10.07\% & +2.42\% \\
 & soy & \textbf{3.323} & 2.737 & 2.807 & +21.38\% & -2.47\% \\
 & nnw & 3.582 & \textbf{3.864} & 3.842 & -7.30\% & +0.58\% \\
 & sag & 4.615 & \textbf{4.685} & 4.240 & -1.48\% & +10.48\% \\
 & csk & 2.962 & \textbf{3.408} & 2.784 & -13.09\% & +22.43\% \\
 & izz & \textbf{3.443} & 2.645 & 2.647 & +30.20\% & -0.08\% \\
 & bum & \textbf{3.020} & 2.929 & 2.882 & +3.09\% & +1.63\% \\
 & gvl & \textbf{3.573} & 2.877 & 3.048 & +24.21\% & -5.61\% \\
 & ndz & 3.253 & \textbf{3.829} & 3.778 & -15.06\% & +1.35\% \\
 & lip & \textbf{3.756} & 3.477 & 3.490 & +8.02\% & -0.37\% \\
 & ken & 3.628 & \textbf{3.645} & 3.558 & -0.46\% & +2.44\% \\
 & gid & \textbf{3.604} & 3.004 & 3.011 & +19.97\% & -0.24\% \\
 & gng & \textbf{4.214} & 4.126 & 3.801 & +2.12\% & +8.55\% \\
 & muy & \textbf{3.803} & 3.172 & 3.058 & +19.89\% & +3.73\% \\
 & niy & \textbf{3.159} & 3.094 & 3.050 & +2.11\% & +1.45\% \\
 & xed & 3.173 & 3.189 & \textbf{3.191} & -0.48\% & -0.09\% \\
 & anv & \textbf{2.921} & 2.639 & 2.908 & +10.67\% & -9.23\% \\
 & lee & 3.698 & \textbf{3.900} & 3.577 & -5.18\% & +9.02\% \\
 & ksf & \textbf{3.553} & 3.542 & 3.237 & +0.30\% & +9.44\% \\
 & pkb & 3.510 & 3.678 & \textbf{3.712} & -4.57\% & -0.92\% \\
 & nko & \textbf{3.730} & 3.650 & 3.258 & +2.20\% & +12.03\% \\
 & lef & 3.280 & \textbf{3.714} & 3.633 & -11.68\% & +2.21\% \\
 & nhr & \textbf{3.839} & 2.898 & 3.477 & +32.50\% & -16.66\% \\
 & biv & 4.087 & \textbf{4.185} & 3.762 & -2.33\% & +11.25\% \\
 & maf & \textbf{3.379} & 2.798 & 2.448 & +20.78\% & +14.30\% \\
 & giz & 3.447 & \textbf{3.936} & 3.190 & -12.44\% & +23.41\% \\
 & tui & \textbf{3.597} & 3.445 & 3.257 & +4.41\% & +5.75\% \\
\end{longtable}
\end{footnotesize}

\clearpage
\onecolumn
\begin{footnotesize}
\begin{longtable}{ccrrrrr}
\caption{BLEU scores of 19 European languages in 9 train sizes produced by 3 models. The highest BLEU score out of the 3 models are boldfaced. DM, OM\textsubscript{u}, OM\textsubscript{d} are shorthands for DiaMT, OnlyMT\textsubscript{undia}, and OnlyMT\textsubscript{d}, respectively. pc(m1, m2) is the percentage change of model m1 over model m2. The higher the percentage change, the better the model m1 is compared to model m2.} \label{tab:bleu_euro_exact_values} \\
\toprule
Size & Lang & DiaMT & OnlyMT\textsubscript{undia} & OnlyMT\textsubscript{dia} & pc(DM, OM\textsubscript{u}) & pc(OM\textsubscript{u}, OM\textsubscript{d}) \\
\midrule
\endfirsthead
\caption[]{BLEU scores of 19 European languages in 9 train sizes produced by 3 models. The highest BLEU score out of the 3 models are boldfaced. DM, OM\textsubscript{u}, OM\textsubscript{d} are shorthands for DiaMT, OnlyMT\textsubscript{undia}, and OnlyMT\textsubscript{d}, respectively. pc(m1, m2) is the percentage change of model m1 over model m2. The higher the percentage change, the better the model m1 is compared to model m2.} \\
\toprule
Size & Lang & DiaMT & OnlyMT\textsubscript{undia} & OnlyMT\textsubscript{dia} & pc(DM, OM\textsubscript{u}) & pc(OM\textsubscript{u}, OM\textsubscript{d}) \\
\midrule
\endhead
\midrule
\multicolumn{7}{r}{Continued on next page} \\
\midrule
\endfoot
\bottomrule
\endlastfoot
\multirow[c]{19}{*}{1k} & el & \textbf{2.363} & 0.443 & 0.683 & +433.32\% & -35.10\% \\
 & cs & \textbf{1.685} & 0.304 & 0.282 & +453.47\% & +7.85\% \\
 & da & \textbf{1.480} & 0.279 & 0.318 & +429.91\% & -12.26\% \\
 & de & \textbf{1.272} & 0.584 & 0.436 & +117.95\% & +33.77\% \\
 & es & \textbf{2.079} & 0.904 & 0.808 & +129.96\% & +11.91\% \\
 & et & \textbf{2.701} & 1.365 & 0.470 & +97.83\% & +190.56\% \\
 & fi & \textbf{0.908} & 0.294 & 0.405 & +208.70\% & -27.31\% \\
 & fr & \textbf{1.367} & 0.761 & 0.177 & +79.69\% & +330.52\% \\
 & hu & \textbf{1.684} & 0.478 & 0.341 & +252.32\% & +40.03\% \\
 & it & \textbf{1.441} & 0.315 & 0.491 & +357.47\% & -35.82\% \\
 & lt & \textbf{2.007} & 0.302 & 0.675 & +564.34\% & -55.26\% \\
 & lv & \textbf{1.833} & 0.301 & 0.484 & +509.02\% & -37.86\% \\
 & nl & \textbf{1.129} & 0.403 & 0.706 & +179.83\% & -42.85\% \\
 & pl & \textbf{1.791} & 0.288 & 0.256 & +521.32\% & +12.72\% \\
 & pt & \textbf{1.719} & 0.410 & 0.215 & +318.96\% & +90.39\% \\
 & ro & \textbf{2.034} & 0.633 & 0.308 & +221.43\% & +105.44\% \\
 & sk & \textbf{1.390} & 1.302 & 0.439 & +6.81\% & +196.26\% \\
 & sl & \textbf{1.580} & 1.061 & 0.686 & +48.81\% & +54.71\% \\
 & sv & \textbf{1.626} & 0.369 & 0.334 & +340.11\% & +10.63\% \\
\hdashline
\multirow[c]{19}{*}{2k} & el & \textbf{2.336} & 0.772 & 0.385 & +202.46\% & +100.61\% \\
 & cs & \textbf{2.230} & 0.870 & 1.045 & +156.25\% & -16.74\% \\
 & da & \textbf{1.738} & 0.776 & 0.347 & +124.02\% & +123.31\% \\
 & de & \textbf{1.739} & 0.214 & 0.252 & +710.98\% & -14.76\% \\
 & es & \textbf{1.893} & 0.759 & 0.670 & +149.27\% & +13.32\% \\
 & et & \textbf{3.018} & 1.126 & 1.142 & +168.07\% & -1.42\% \\
 & fi & \textbf{1.232} & 0.691 & 0.623 & +78.33\% & +10.91\% \\
 & fr & \textbf{1.312} & 0.598 & 0.510 & +119.45\% & +17.25\% \\
 & hu & \textbf{1.856} & 0.766 & 0.808 & +142.24\% & -5.24\% \\
 & it & \textbf{1.037} & 0.494 & 0.589 & +109.96\% & -16.19\% \\
 & lt & \textbf{2.032} & 0.906 & 0.801 & +124.16\% & +13.20\% \\
 & lv & \textbf{2.339} & 1.241 & 1.223 & +88.42\% & +1.47\% \\
 & nl & \textbf{1.823} & 0.299 & 0.206 & +508.54\% & +45.24\% \\
 & pl & \textbf{2.734} & 0.632 & 0.342 & +332.61\% & +84.81\% \\
 & pt & \textbf{1.314} & 1.021 & 0.909 & +28.76\% & +12.23\% \\
 & ro & \textbf{2.842} & 0.582 & 0.798 & +388.02\% & -27.02\% \\
 & sk & \textbf{1.463} & 1.289 & 0.747 & +13.44\% & +72.51\% \\
 & sl & \textbf{2.514} & 1.502 & 0.805 & +67.35\% & +86.69\% \\
 & sv & \textbf{2.440} & 0.680 & 1.096 & +258.96\% & -38.00\% \\
\hdashline
\multirow[c]{19}{*}{3k} & el & \textbf{1.470} & 1.029 & 0.770 & +42.81\% & +33.63\% \\
 & cs & \textbf{2.949} & 0.962 & 1.113 & +206.45\% & -13.59\% \\
 & da & \textbf{2.925} & 0.946 & 0.766 & +209.32\% & +23.46\% \\
 & de & \textbf{1.237} & 0.633 & 1.195 & +95.38\% & -46.98\% \\
 & es & \textbf{2.395} & 0.647 & 0.943 & +270.33\% & -31.43\% \\
 & et & \textbf{1.974} & 1.203 & 1.407 & +64.08\% & -14.51\% \\
 & fi & \textbf{1.543} & 0.780 & 0.756 & +97.81\% & +3.23\% \\
 & fr & \textbf{2.343} & 1.202 & 2.195 & +94.87\% & -45.22\% \\
 & hu & \textbf{1.484} & 1.464 & 1.102 & +1.36\% & +32.87\% \\
 & it & \textbf{1.463} & 1.129 & 1.009 & +29.58\% & +11.95\% \\
 & lt & \textbf{1.727} & 1.162 & 0.661 & +48.57\% & +75.90\% \\
 & lv & 2.023 & \textbf{2.046} & 1.147 & -1.09\% & +78.41\% \\
 & nl & \textbf{1.351} & 1.099 & 0.674 & +22.91\% & +63.07\% \\
 & pl & \textbf{2.120} & 0.875 & 0.762 & +142.25\% & +14.77\% \\
 & pt & \textbf{1.715} & 1.000 & 1.006 & +71.51\% & -0.59\% \\
 & ro & \textbf{3.727} & 1.208 & 0.743 & +208.47\% & +62.64\% \\
 & sk & \textbf{1.888} & 1.559 & 1.093 & +21.11\% & +42.61\% \\
 & sl & \textbf{2.998} & 1.208 & 0.441 & +148.10\% & +174.28\% \\
 & sv & \textbf{1.837} & 1.542 & 1.224 & +19.08\% & +26.06\% \\
\hdashline
\multirow[c]{3}{*}{4k} & el & \textbf{2.021} & 1.814 & 1.772 & +11.39\% & +2.35\% \\
 & cs & \textbf{3.496} & 1.266 & 1.515 & +176.28\% & -16.45\% \\
 & da & \textbf{2.694} & 1.336 & 1.601 & +101.60\% & -16.55\% \\
\multirow[c]{16}{*}{4k} & de & \textbf{1.631} & 1.338 & 1.602 & +21.97\% & -16.48\% \\
 & es & \textbf{2.314} & 1.677 & 1.600 & +38.00\% & +4.82\% \\
 & et & \textbf{2.753} & 1.340 & 1.488 & +105.36\% & -9.90\% \\
 & fi & \textbf{1.799} & 1.114 & 1.482 & +61.48\% & -24.81\% \\
 & fr & \textbf{2.302} & 1.790 & 1.557 & +28.61\% & +14.98\% \\
 & hu & 1.970 & 1.387 & \textbf{2.065} & +42.04\% & -32.83\% \\
 & it & \textbf{1.811} & 0.856 & 0.979 & +111.52\% & -12.57\% \\
 & lt & 1.447 & 1.379 & \textbf{1.637} & +4.98\% & -15.79\% \\
 & lv & \textbf{2.597} & 2.206 & 1.693 & +17.72\% & +30.31\% \\
 & nl & \textbf{1.697} & 1.115 & 1.356 & +52.23\% & -17.81\% \\
 & pl & \textbf{1.902} & 1.490 & 1.523 & +27.64\% & -2.15\% \\
 & pt & \textbf{1.937} & 1.103 & 1.122 & +75.65\% & -1.73\% \\
 & ro & \textbf{3.470} & 1.935 & 2.346 & +79.36\% & -17.53\% \\
 & sk & \textbf{1.880} & 1.590 & 1.577 & +18.18\% & +0.87\% \\
 & sl & \textbf{3.235} & 1.564 & 1.517 & +106.86\% & +3.08\% \\
 & sv & \textbf{2.240} & 1.503 & 1.348 & +49.03\% & +11.47\% \\
\hdashline
\multirow[c]{19}{*}{5k} & el & 2.321 & 2.076 & \textbf{2.761} & +11.81\% & -24.83\% \\
 & cs & \textbf{3.079} & 2.281 & 2.131 & +35.01\% & +7.01\% \\
 & da & \textbf{2.428} & 2.188 & 2.305 & +10.99\% & -5.09\% \\
 & de & \textbf{2.016} & 1.247 & 1.126 & +61.74\% & +10.68\% \\
 & es & \textbf{2.442} & 1.288 & 1.802 & +89.62\% & -28.53\% \\
 & et & 1.862 & 2.234 & \textbf{2.386} & -16.61\% & -6.39\% \\
 & fi & 1.370 & \textbf{1.473} & 1.452 & -6.98\% & +1.48\% \\
 & fr & \textbf{3.024} & 2.259 & 2.648 & +33.85\% & -14.68\% \\
 & hu & 2.086 & 1.738 & \textbf{2.173} & +20.02\% & -19.99\% \\
 & it & \textbf{1.450} & 0.864 & 1.251 & +67.85\% & -30.90\% \\
 & lt & \textbf{2.762} & 1.801 & 1.764 & +53.39\% & +2.10\% \\
 & lv & \textbf{3.662} & 2.189 & 1.940 & +67.25\% & +12.88\% \\
 & nl & \textbf{2.396} & 1.402 & 1.347 & +70.89\% & +4.04\% \\
 & pl & \textbf{2.259} & 1.889 & 2.228 & +19.60\% & -15.22\% \\
 & pt & \textbf{1.851} & 1.250 & 1.268 & +48.06\% & -1.37\% \\
 & ro & 2.977 & 2.916 & \textbf{3.285} & +2.08\% & -11.23\% \\
 & sk & 1.932 & 1.792 & \textbf{2.068} & +7.85\% & -13.38\% \\
 & sl & 2.332 & \textbf{2.730} & 1.933 & -14.57\% & +41.24\% \\
 & sv & \textbf{2.160} & 1.516 & 1.717 & +42.50\% & -11.71\% \\
\hdashline
\multirow[c]{19}{*}{25k} & el & 5.316 & \textbf{5.485} & 5.451 & -3.07\% & +0.62\% \\
 & cs & 5.226 & \textbf{6.278} & 5.230 & -16.76\% & +20.05\% \\
 & da & 4.395 & 4.707 & \textbf{5.110} & -6.63\% & -7.89\% \\
 & de & 3.799 & \textbf{3.816} & 3.505 & -0.46\% & +8.87\% \\
 & es & 4.839 & 5.105 & \textbf{6.021} & -5.21\% & -15.22\% \\
 & et & 4.720 & \textbf{5.312} & 5.179 & -11.15\% & +2.58\% \\
 & fi & 3.012 & \textbf{4.118} & 3.967 & -26.86\% & +3.82\% \\
 & fr & 3.952 & \textbf{5.073} & 4.160 & -22.10\% & +21.93\% \\
 & hu & 4.555 & \textbf{4.766} & 4.022 & -4.42\% & +18.51\% \\
 & it & 3.609 & \textbf{3.846} & 3.679 & -6.16\% & +4.53\% \\
 & lt & 4.304 & 4.509 & \textbf{5.139} & -4.54\% & -12.26\% \\
 & lv & 5.187 & \textbf{6.161} & 6.153 & -15.81\% & +0.13\% \\
 & nl & 3.865 & 3.705 & \textbf{4.255} & +4.29\% & -12.92\% \\
 & pl & 4.373 & \textbf{5.091} & 4.026 & -14.10\% & +26.44\% \\
 & pt & 4.168 & 4.371 & \textbf{5.218} & -4.66\% & -16.23\% \\
 & ro & 6.768 & 6.662 & \textbf{7.420} & +1.59\% & -10.22\% \\
 & sk & 4.002 & 5.404 & \textbf{6.231} & -25.95\% & -13.26\% \\
 & sl & 5.318 & 5.370 & \textbf{5.800} & -0.97\% & -7.42\% \\
 & sv & 4.020 & 4.910 & \textbf{5.178} & -18.14\% & -5.17\% \\
\hdashline
\multirow[c]{6}{*}{125k} & el & 7.959 & \textbf{15.371} & 14.007 & -48.22\% & +9.74\% \\
 & cs & 8.162 & 15.207 & \textbf{15.404} & -46.33\% & -1.28\% \\
 & da & 7.458 & 13.531 & \textbf{14.038} & -44.88\% & -3.61\% \\
 & de & 6.014 & 9.442 & \textbf{9.753} & -36.30\% & -3.18\% \\
 & es & 9.406 & 14.811 & \textbf{15.585} & -36.50\% & -4.96\% \\
 & et & 7.225 & 12.657 & \textbf{13.286} & -42.92\% & -4.74\% \\
\multirow[c]{12}{*}{125k} & fi & 5.727 & 8.256 & \textbf{8.826} & -30.63\% & -6.46\% \\
 & fr & 7.285 & 11.421 & \textbf{11.451} & -36.21\% & -0.26\% \\
 & hu & 6.950 & 10.277 & \textbf{11.469} & -32.38\% & -10.39\% \\
 & it & 5.200 & 10.023 & \textbf{10.491} & -48.12\% & -4.46\% \\
 & lt & 7.228 & 11.913 & \textbf{13.014} & -39.32\% & -8.47\% \\
 & lv & 8.407 & 14.056 & \textbf{14.562} & -40.19\% & -3.48\% \\
 & nl & 5.547 & 8.862 & \textbf{8.952} & -37.41\% & -1.00\% \\
 & pl & 6.989 & 12.667 & \textbf{13.074} & -44.82\% & -3.12\% \\
 & pt & 6.893 & 11.786 & \textbf{12.211} & -41.51\% & -3.49\% \\
 & ro & 10.953 & 22.082 & \textbf{22.668} & -50.40\% & -2.59\% \\
 & sk & 7.961 & 15.309 & \textbf{16.546} & -48.00\% & -7.48\% \\
 & sv & 7.500 & 14.692 & \textbf{17.036} & -48.96\% & -13.76\% \\
\hdashline
\multirow[c]{10}{*}{625k} & el & 14.839 & \textbf{24.398} & 24.057 & -39.18\% & +1.42\% \\
 & da & 13.473 & 22.328 & \textbf{22.442} & -39.66\% & -0.51\% \\
 & de & 10.130 & \textbf{17.755} & 17.312 & -42.95\% & +2.56\% \\
 & es & 16.397 & \textbf{25.758} & 25.753 & -36.34\% & +0.02\% \\
 & fi & 7.315 & 15.329 & \textbf{15.389} & -52.28\% & -0.39\% \\
 & fr & 12.707 & \textbf{22.489} & 21.858 & -43.50\% & +2.89\% \\
 & it & 10.834 & 19.566 & \textbf{20.046} & -44.63\% & -2.39\% \\
 & nl & 9.007 & \textbf{18.030} & 17.105 & -50.05\% & +5.41\% \\
 & pt & 12.621 & 22.844 & \textbf{23.526} & -44.75\% & -2.90\% \\
 & sv & 13.529 & \textbf{25.070} & 24.967 & -46.03\% & +0.41\% \\
\hdashline
\multirow[c]{9}{*}{1M} & el & 19.648 & 27.089 & \textbf{27.475} & -27.47\% & -1.40\% \\
 & de & 12.562 & 21.479 & \textbf{21.566} & -41.52\% & -0.40\% \\
 & es & 19.741 & 28.380 & \textbf{28.442} & -30.44\% & -0.22\% \\
 & fi & 10.747 & \textbf{19.230} & 18.995 & -44.11\% & +1.24\% \\
 & fr & 16.156 & 25.248 & \textbf{25.289} & -36.01\% & -0.16\% \\
 & it & 15.239 & 22.771 & \textbf{23.383} & -33.08\% & -2.62\% \\
 & nl & 12.163 & 20.052 & \textbf{20.449} & -39.34\% & -1.94\% \\
 & pt & 18.431 & 26.172 & \textbf{26.987} & -29.58\% & -3.02\% \\
 & sv & 18.346 & 27.500 & \textbf{27.839} & -33.29\% & -1.22\% \\
\end{longtable}
\end{footnotesize}
\clearpage
\twocolumn

\clearpage
\onecolumn
\begin{footnotesize}
\begin{longtable}{ccrrrrrr}
\caption{DER and WER of 36 African languages in 5 train sizes produced by 2 models. The lowest DER and WER scores out of the 2 models are boldfaced. DM\textsubscript{D}, OD\textsubscript{D}, DM\textsubscript{W}, OD\textsubscript{W} are shorthands for DiaMT\textsubscript{DER}, OnlyDia\textsubscript{DER}, DiaMT\textsubscript{WER}, and OnlyDia\textsubscript{WER}, respectively. pc(m1, m2) is the percentage change of model m1 over model m2. The lower the percentage change, the better the model m1 is compared to model m2.} \label{tab:der_wer_afri_exact_values} \\
\toprule
Size & Lang & DiaMT\textsubscript{DER} & OnlyDia\textsubscript{DER} & pc(DM\textsubscript{D},OD\textsubscript{D}) & DiaMT\textsubscript{WER} & OnlyDia\textsubscript{WER} & pc(DM\textsubscript{W},OD\textsubscript{W}) \\
\midrule
\endfirsthead
\caption[]{DER and WER of 36 African languages in 5 train sizes produced by 2 models. The lowest DER and WER scores out of the 2 models are boldfaced. DM\textsubscript{D}, OD\textsubscript{D}, DM\textsubscript{W}, OD\textsubscript{W} are shorthands for DiaMT\textsubscript{DER}, OnlyDia\textsubscript{DER}, DiaMT\textsubscript{WER}, and OnlyDia\textsubscript{WER}, respectively. pc(m1, m2) is the percentage change of model m1 over model m2. The lower the percentage change, the better the model m1 is compared to model m2.} \\
\toprule
Size & Lang & DiaMT\textsubscript{DER} & OnlyDia\textsubscript{DER} & pc(DM\textsubscript{D},OD\textsubscript{D}) & DiaMT\textsubscript{WER} & OnlyDia\textsubscript{WER} & pc(DM\textsubscript{W},OD\textsubscript{W}) \\
\midrule
\endhead
\midrule
\multicolumn{8}{r}{Continued on next page} \\
\midrule
\endfoot
\bottomrule
\endlastfoot
\multirow[c]{36}{*}{1k} & bex & 0.379 & \textbf{0.329} & +15.29\% & 0.435 & \textbf{0.384} & +13.50\% \\
 & fon & \textbf{0.443} & 0.520 & -14.68\% & \textbf{0.502} & 0.552 & -9.04\% \\
 & mkl & 0.400 & \textbf{0.372} & +7.62\% & 0.439 & \textbf{0.399} & +9.94\% \\
 & mnf & 0.620 & \textbf{0.408} & +52.01\% & 0.676 & \textbf{0.480} & +40.70\% \\
 & bud & 0.434 & \textbf{0.269} & +61.53\% & 0.521 & \textbf{0.366} & +42.42\% \\
 & eza & 0.482 & \textbf{0.297} & +62.42\% & 0.554 & \textbf{0.400} & +38.31\% \\
 & sig & 0.277 & \textbf{0.151} & +84.13\% & 0.323 & \textbf{0.209} & +54.98\% \\
 & bqc & 0.437 & \textbf{0.272} & +60.75\% & 0.519 & \textbf{0.348} & +48.91\% \\
 & kia & 0.370 & \textbf{0.213} & +73.75\% & 0.397 & \textbf{0.231} & +71.64\% \\
 & soy & 0.440 & \textbf{0.253} & +74.00\% & 0.502 & \textbf{0.312} & +61.13\% \\
 & nnw & 0.434 & \textbf{0.394} & +9.96\% & 0.478 & \textbf{0.435} & +9.88\% \\
 & sag & 0.469 & \textbf{0.236} & +98.85\% & 0.482 & \textbf{0.267} & +80.47\% \\
 & csk & 0.401 & \textbf{0.224} & +79.51\% & 0.437 & \textbf{0.275} & +58.56\% \\
 & izz & 0.425 & \textbf{0.254} & +67.46\% & 0.480 & \textbf{0.335} & +43.32\% \\
 & bum & 0.344 & \textbf{0.200} & +71.93\% & 0.351 & \textbf{0.202} & +73.79\% \\
 & gvl & 0.429 & \textbf{0.244} & +75.58\% & 0.495 & \textbf{0.320} & +54.58\% \\
 & ndz & 0.540 & \textbf{0.439} & +22.95\% & 0.651 & \textbf{0.568} & +14.59\% \\
 & lip & 0.398 & \textbf{0.235} & +69.50\% & 0.429 & \textbf{0.272} & +57.49\% \\
 & ken & 0.514 & \textbf{0.364} & +41.32\% & 0.581 & \textbf{0.447} & +29.97\% \\
 & gid & 0.306 & \textbf{0.151} & +102.02\% & 0.340 & \textbf{0.176} & +92.81\% \\
 & gng & 0.334 & \textbf{0.219} & +52.17\% & 0.357 & \textbf{0.247} & +44.33\% \\
 & muy & 0.475 & \textbf{0.333} & +42.52\% & 0.521 & \textbf{0.395} & +31.86\% \\
 & niy & 0.535 & \textbf{0.407} & +31.58\% & 0.648 & \textbf{0.539} & +20.31\% \\
 & xed & 0.351 & \textbf{0.225} & +55.99\% & 0.392 & \textbf{0.278} & +40.93\% \\
 & anv & 0.517 & \textbf{0.474} & +9.02\% & 0.603 & \textbf{0.556} & +8.38\% \\
 & lee & 0.440 & \textbf{0.293} & +50.13\% & 0.536 & \textbf{0.395} & +35.58\% \\
 & ksf & 0.498 & \textbf{0.341} & +45.93\% & 0.565 & \textbf{0.404} & +40.11\% \\
 & pkb & 0.315 & \textbf{0.134} & +136.14\% & 0.365 & \textbf{0.193} & +88.97\% \\
 & nko & 0.458 & \textbf{0.321} & +43.02\% & 0.532 & \textbf{0.389} & +36.78\% \\
 & lef & 0.387 & \textbf{0.233} & +66.13\% & 0.421 & \textbf{0.269} & +56.92\% \\
 & nhr & 0.451 & \textbf{0.265} & +70.47\% & 0.526 & \textbf{0.351} & +49.80\% \\
 & mgc & 0.344 & \textbf{0.166} & +107.35\% & 0.360 & \textbf{0.189} & +89.88\% \\
 & biv & 0.510 & \textbf{0.431} & +18.33\% & 0.504 & \textbf{0.413} & +21.98\% \\
 & maf & 0.444 & \textbf{0.240} & +84.47\% & 0.422 & \textbf{0.229} & +84.04\% \\
 & giz & 0.371 & \textbf{0.162} & +128.91\% & 0.386 & \textbf{0.180} & +114.35\% \\
 & tui & 0.419 & \textbf{0.400} & +4.61\% & 0.468 & \textbf{0.443} & +5.64\% \\
\hdashline
\multirow[c]{24}{*}{2k} & bex & 0.500 & \textbf{0.337} & +48.33\% & 0.548 & \textbf{0.389} & +40.86\% \\
 & fon & 0.429 & \textbf{0.353} & +21.47\% & 0.487 & \textbf{0.403} & +20.64\% \\
 & mkl & 0.401 & \textbf{0.133} & +202.72\% & 0.439 & \textbf{0.172} & +155.29\% \\
 & mnf & 0.563 & \textbf{0.399} & +41.21\% & 0.628 & \textbf{0.465} & +35.10\% \\
 & bud & 0.465 & \textbf{0.210} & +121.69\% & 0.554 & \textbf{0.305} & +81.52\% \\
 & eza & 0.548 & \textbf{0.313} & +75.48\% & 0.608 & \textbf{0.410} & +48.38\% \\
 & sig & 0.394 & \textbf{0.152} & +158.50\% & 0.430 & \textbf{0.213} & +101.45\% \\
 & bqc & 0.367 & \textbf{0.191} & +91.97\% & 0.452 & \textbf{0.263} & +71.79\% \\
 & kia & 0.468 & \textbf{0.143} & +227.17\% & 0.493 & \textbf{0.164} & +200.68\% \\
 & soy & 0.449 & \textbf{0.193} & +132.57\% & 0.510 & \textbf{0.256} & +99.45\% \\
 & nnw & 0.437 & \textbf{0.234} & +86.59\% & 0.479 & \textbf{0.297} & +61.43\% \\
 & sag & 0.491 & \textbf{0.162} & +202.25\% & 0.507 & \textbf{0.206} & +145.70\% \\
 & csk & 0.506 & \textbf{0.173} & +193.14\% & 0.533 & \textbf{0.230} & +131.88\% \\
 & izz & 0.533 & \textbf{0.268} & +99.16\% & 0.571 & \textbf{0.341} & +67.15\% \\
 & bum & 0.400 & \textbf{0.144} & +178.16\% & 0.402 & \textbf{0.152} & +165.09\% \\
 & gvl & 0.462 & \textbf{0.182} & +154.05\% & 0.528 & \textbf{0.262} & +101.08\% \\
 & ndz & 0.521 & \textbf{0.393} & +32.40\% & 0.639 & \textbf{0.526} & +21.63\% \\
 & lip & \textbf{0.375} & 0.427 & -12.20\% & \textbf{0.414} & 0.446 & -7.33\% \\
 & ken & 0.518 & \textbf{0.286} & +81.07\% & 0.590 & \textbf{0.373} & +58.15\% \\
 & gid & 0.324 & \textbf{0.111} & +190.70\% & 0.357 & \textbf{0.136} & +162.53\% \\
 & gng & 0.370 & \textbf{0.173} & +113.68\% & 0.393 & \textbf{0.204} & +92.19\% \\
 & muy & 0.525 & \textbf{0.256} & +105.50\% & 0.572 & \textbf{0.326} & +75.63\% \\
 & niy & 0.580 & \textbf{0.317} & +83.29\% & 0.684 & \textbf{0.476} & +43.90\% \\
 & xed & 0.511 & \textbf{0.144} & +255.18\% & 0.546 & \textbf{0.207} & +163.58\% \\
\multirow[c]{12}{*}{2k} & anv & 0.534 & \textbf{0.392} & +36.11\% & 0.619 & \textbf{0.478} & +29.62\% \\
 & lee & 0.446 & \textbf{0.356} & +25.50\% & 0.546 & \textbf{0.438} & +24.54\% \\
 & ksf & 0.500 & \textbf{0.260} & +92.45\% & 0.568 & \textbf{0.332} & +71.27\% \\
 & pkb & 0.359 & \textbf{0.097} & +270.73\% & 0.411 & \textbf{0.155} & +165.69\% \\
 & nko & 0.554 & \textbf{0.224} & +146.96\% & 0.622 & \textbf{0.303} & +105.05\% \\
 & lef & 0.424 & \textbf{0.164} & +157.92\% & 0.457 & \textbf{0.204} & +124.29\% \\
 & nhr & 0.502 & \textbf{0.222} & +125.87\% & 0.566 & \textbf{0.317} & +78.39\% \\
 & mgc & 0.355 & \textbf{0.116} & +207.01\% & 0.371 & \textbf{0.143} & +159.39\% \\
 & biv & 0.369 & \textbf{0.298} & +23.91\% & 0.373 & \textbf{0.284} & +31.40\% \\
 & maf & 0.377 & \textbf{0.146} & +158.81\% & 0.358 & \textbf{0.147} & +143.98\% \\
 & giz & 0.355 & \textbf{0.137} & +158.46\% & 0.371 & \textbf{0.155} & +139.60\% \\
 & tui & 0.475 & \textbf{0.337} & +40.79\% & 0.525 & \textbf{0.377} & +39.12\% \\
\hdashline
\multirow[c]{35}{*}{3k} & bex & 0.509 & \textbf{0.277} & +83.48\% & 0.556 & \textbf{0.335} & +66.17\% \\
 & fon & 0.453 & \textbf{0.193} & +134.66\% & 0.511 & \textbf{0.269} & +89.80\% \\
 & mkl & 0.543 & \textbf{0.129} & +320.15\% & 0.574 & \textbf{0.169} & +239.95\% \\
 & mnf & 0.639 & \textbf{0.394} & +62.40\% & 0.695 & \textbf{0.459} & +51.63\% \\
 & bud & 0.451 & \textbf{0.210} & +115.08\% & 0.538 & \textbf{0.303} & +77.42\% \\
 & eza & 0.633 & \textbf{0.237} & +167.05\% & 0.682 & \textbf{0.356} & +91.66\% \\
 & sig & 0.395 & \textbf{0.146} & +170.11\% & 0.428 & \textbf{0.205} & +109.43\% \\
 & bqc & 0.404 & \textbf{0.173} & +132.87\% & 0.486 & \textbf{0.246} & +97.71\% \\
 & kia & 0.489 & \textbf{0.140} & +249.77\% & 0.514 & \textbf{0.155} & +230.41\% \\
 & soy & 0.477 & \textbf{0.190} & +151.06\% & 0.535 & \textbf{0.250} & +114.28\% \\
 & nnw & 0.480 & \textbf{0.259} & +84.93\% & 0.522 & \textbf{0.315} & +65.83\% \\
 & sag & 0.484 & \textbf{0.139} & +247.13\% & 0.497 & \textbf{0.186} & +166.73\% \\
 & csk & 0.490 & \textbf{0.174} & +181.82\% & 0.522 & \textbf{0.228} & +128.87\% \\
 & izz & 0.604 & \textbf{0.217} & +177.87\% & 0.640 & \textbf{0.301} & +112.81\% \\
 & bum & 0.419 & \textbf{0.126} & +233.42\% & 0.424 & \textbf{0.134} & +216.14\% \\
 & gvl & 0.482 & \textbf{0.194} & +149.05\% & 0.549 & \textbf{0.276} & +98.96\% \\
 & ndz & 0.548 & \textbf{0.359} & +52.79\% & 0.662 & \textbf{0.499} & +32.61\% \\
 & lip & 0.466 & \textbf{0.144} & +222.73\% & 0.501 & \textbf{0.191} & +162.68\% \\
 & ken & 0.547 & \textbf{0.283} & +93.58\% & 0.613 & \textbf{0.372} & +64.88\% \\
 & gid & 0.360 & \textbf{0.133} & +170.86\% & 0.395 & \textbf{0.157} & +151.61\% \\
 & gng & 0.406 & \textbf{0.148} & +174.73\% & 0.430 & \textbf{0.182} & +136.78\% \\
 & muy & 0.533 & \textbf{0.219} & +143.12\% & 0.577 & \textbf{0.296} & +94.89\% \\
 & niy & 0.557 & \textbf{0.299} & +86.51\% & 0.676 & \textbf{0.463} & +46.05\% \\
 & xed & 0.434 & \textbf{0.115} & +278.74\% & 0.474 & \textbf{0.184} & +157.66\% \\
 & anv & 0.547 & \textbf{0.286} & +91.41\% & 0.630 & \textbf{0.389} & +61.77\% \\
 & lee & 0.469 & \textbf{0.172} & +173.21\% & 0.560 & \textbf{0.295} & +89.73\% \\
 & ksf & 0.602 & \textbf{0.283} & +112.98\% & 0.660 & \textbf{0.348} & +89.72\% \\
 & pkb & 0.367 & \textbf{0.108} & +240.02\% & 0.425 & \textbf{0.166} & +156.50\% \\
 & nko & 0.492 & \textbf{0.328} & +50.11\% & 0.570 & \textbf{0.391} & +45.87\% \\
 & lef & 0.501 & \textbf{0.187} & +167.64\% & 0.534 & \textbf{0.223} & +138.86\% \\
 & nhr & 0.512 & \textbf{0.190} & +169.21\% & 0.581 & \textbf{0.284} & +104.41\% \\
 & biv & 0.443 & \textbf{0.296} & +49.73\% & 0.441 & \textbf{0.283} & +55.92\% \\
 & maf & 0.422 & \textbf{0.125} & +238.63\% & 0.399 & \textbf{0.130} & +206.54\% \\
 & giz & 0.373 & \textbf{0.159} & +134.82\% & 0.383 & \textbf{0.168} & +128.71\% \\
 & tui & 0.509 & \textbf{0.246} & +107.13\% & 0.562 & \textbf{0.291} & +93.25\% \\
\hdashline
\multirow[c]{14}{*}{4k} & bex & 0.541 & \textbf{0.283} & +90.91\% & 0.586 & \textbf{0.341} & +72.00\% \\
 & fon & 0.608 & \textbf{0.195} & +212.10\% & 0.653 & \textbf{0.271} & +141.44\% \\
 & mkl & 0.440 & \textbf{0.231} & +90.63\% & 0.481 & \textbf{0.260} & +84.86\% \\
 & mnf & 0.556 & \textbf{0.278} & +99.72\% & 0.627 & \textbf{0.365} & +71.78\% \\
 & bud & 0.516 & \textbf{0.193} & +167.13\% & 0.599 & \textbf{0.291} & +105.50\% \\
 & eza & 0.688 & \textbf{0.217} & +217.17\% & 0.732 & \textbf{0.337} & +117.22\% \\
 & sig & 0.428 & \textbf{0.189} & +126.06\% & 0.462 & \textbf{0.241} & +91.87\% \\
 & bqc & 0.391 & \textbf{0.182} & +114.58\% & 0.472 & \textbf{0.255} & +85.16\% \\
 & kia & 0.474 & \textbf{0.152} & +211.11\% & 0.496 & \textbf{0.168} & +195.58\% \\
 & soy & 0.518 & \textbf{0.184} & +181.13\% & 0.571 & \textbf{0.246} & +132.02\% \\
 & nnw & 0.486 & \textbf{0.203} & +139.40\% & 0.527 & \textbf{0.273} & +93.32\% \\
 & sag & 0.488 & \textbf{0.167} & +192.92\% & 0.505 & \textbf{0.207} & +143.79\% \\
 & csk & 0.545 & \textbf{0.160} & +241.34\% & 0.571 & \textbf{0.217} & +163.93\% \\
 & izz & 0.650 & \textbf{0.220} & +195.77\% & 0.681 & \textbf{0.302} & +125.71\% \\
\multirow[c]{21}{*}{4k} & bum & 0.415 & \textbf{0.118} & +251.30\% & 0.424 & \textbf{0.126} & +236.71\% \\
 & gvl & 0.497 & \textbf{0.174} & +184.97\% & 0.566 & \textbf{0.258} & +119.11\% \\
 & ndz & 0.565 & \textbf{0.357} & +58.34\% & 0.675 & \textbf{0.501} & +34.73\% \\
 & lip & 0.501 & \textbf{0.390} & +28.45\% & 0.531 & \textbf{0.408} & +30.02\% \\
 & ken & 0.599 & \textbf{0.307} & +94.97\% & 0.660 & \textbf{0.394} & +67.32\% \\
 & gid & 0.328 & \textbf{0.086} & +283.23\% & 0.361 & \textbf{0.110} & +228.29\% \\
 & gng & 0.458 & \textbf{0.142} & +223.44\% & 0.479 & \textbf{0.177} & +170.99\% \\
 & muy & 0.544 & \textbf{0.204} & +166.17\% & 0.591 & \textbf{0.286} & +106.82\% \\
 & niy & 0.592 & \textbf{0.253} & +134.08\% & 0.700 & \textbf{0.431} & +62.38\% \\
 & xed & 0.525 & \textbf{0.162} & +224.47\% & 0.562 & \textbf{0.218} & +157.70\% \\
 & anv & 0.583 & \textbf{0.357} & +63.52\% & 0.663 & \textbf{0.449} & +47.52\% \\
 & lee & 0.512 & \textbf{0.217} & +135.90\% & 0.604 & \textbf{0.331} & +82.51\% \\
 & ksf & 0.641 & \textbf{0.221} & +190.13\% & 0.699 & \textbf{0.300} & +133.44\% \\
 & pkb & 0.412 & \textbf{0.086} & +377.14\% & 0.463 & \textbf{0.142} & +224.93\% \\
 & nko & 0.573 & \textbf{0.287} & +99.85\% & 0.641 & \textbf{0.354} & +80.83\% \\
 & lef & 0.454 & \textbf{0.158} & +188.10\% & 0.493 & \textbf{0.196} & +150.89\% \\
 & nhr & 0.579 & \textbf{0.227} & +155.26\% & 0.642 & \textbf{0.312} & +105.58\% \\
 & biv & 0.390 & \textbf{0.278} & +40.28\% & 0.394 & \textbf{0.265} & +48.80\% \\
 & maf & 0.422 & \textbf{0.137} & +209.16\% & 0.404 & \textbf{0.138} & +192.74\% \\
 & giz & 0.449 & \textbf{0.120} & +274.16\% & 0.459 & \textbf{0.139} & +231.04\% \\
 & tui & 0.521 & \textbf{0.169} & +207.68\% & 0.574 & \textbf{0.222} & +158.45\% \\
\hdashline
\multirow[c]{35}{*}{5k} & bex & 0.533 & \textbf{0.144} & +270.79\% & 0.583 & \textbf{0.221} & +163.19\% \\
 & fon & 0.536 & \textbf{0.171} & +214.23\% & 0.588 & \textbf{0.253} & +132.42\% \\
 & mkl & 0.454 & \textbf{0.120} & +279.24\% & 0.491 & \textbf{0.159} & +209.50\% \\
 & mnf & 0.596 & \textbf{0.434} & +37.38\% & 0.661 & \textbf{0.489} & +35.34\% \\
 & bud & 0.479 & \textbf{0.179} & +168.25\% & 0.566 & \textbf{0.277} & +104.66\% \\
 & eza & 0.687 & \textbf{0.188} & +265.50\% & 0.731 & \textbf{0.316} & +131.07\% \\
 & sig & 0.479 & \textbf{0.290} & +65.10\% & 0.509 & \textbf{0.325} & +56.52\% \\
 & bqc & 0.441 & \textbf{0.193} & +127.91\% & 0.518 & \textbf{0.258} & +100.45\% \\
 & kia & 0.450 & \textbf{0.113} & +298.43\% & 0.473 & \textbf{0.133} & +255.41\% \\
 & soy & 0.684 & \textbf{0.180} & +279.85\% & 0.734 & \textbf{0.242} & +202.99\% \\
 & nnw & 0.549 & \textbf{0.181} & +202.55\% & 0.587 & \textbf{0.252} & +132.34\% \\
 & sag & 0.515 & \textbf{0.140} & +267.78\% & 0.530 & \textbf{0.185} & +187.28\% \\
 & csk & 0.615 & \textbf{0.169} & +264.27\% & 0.647 & \textbf{0.225} & +187.93\% \\
 & izz & 0.675 & \textbf{0.203} & +232.99\% & 0.706 & \textbf{0.288} & +144.78\% \\
 & bum & 0.387 & \textbf{0.132} & +194.21\% & 0.392 & \textbf{0.139} & +181.72\% \\
 & gvl & 0.510 & \textbf{0.181} & +182.14\% & 0.579 & \textbf{0.264} & +119.31\% \\
 & ndz & 0.546 & \textbf{0.330} & +65.61\% & 0.662 & \textbf{0.480} & +37.92\% \\
 & lip & 0.484 & \textbf{0.155} & +213.02\% & 0.516 & \textbf{0.199} & +159.32\% \\
 & ken & 0.570 & \textbf{0.311} & +83.48\% & 0.632 & \textbf{0.394} & +60.49\% \\
 & gid & 0.360 & \textbf{0.097} & +272.09\% & 0.391 & \textbf{0.123} & +217.05\% \\
 & gng & 0.396 & \textbf{0.127} & +211.51\% & 0.419 & \textbf{0.166} & +151.77\% \\
 & muy & 0.553 & \textbf{0.356} & +55.43\% & 0.594 & \textbf{0.409} & +45.42\% \\
 & niy & 0.635 & \textbf{0.273} & +132.38\% & 0.729 & \textbf{0.444} & +64.35\% \\
 & xed & 0.430 & \textbf{0.095} & +352.39\% & 0.469 & \textbf{0.165} & +184.70\% \\
 & anv & 0.523 & \textbf{0.349} & +49.86\% & 0.614 & \textbf{0.441} & +39.12\% \\
 & lee & 0.497 & \textbf{0.237} & +109.77\% & 0.590 & \textbf{0.348} & +69.40\% \\
 & ksf & 0.630 & \textbf{0.226} & +178.96\% & 0.688 & \textbf{0.303} & +127.04\% \\
 & pkb & 0.414 & \textbf{0.088} & +369.56\% & 0.460 & \textbf{0.145} & +217.36\% \\
 & nko & 0.550 & \textbf{0.281} & +95.98\% & 0.617 & \textbf{0.351} & +76.10\% \\
 & lef & 0.483 & \textbf{0.301} & +60.75\% & 0.518 & \textbf{0.324} & +59.96\% \\
 & nhr & 0.577 & \textbf{0.205} & +182.17\% & 0.637 & \textbf{0.295} & +115.85\% \\
 & biv & 0.360 & \textbf{0.118} & +206.06\% & 0.358 & \textbf{0.125} & +187.14\% \\
 & maf & 0.442 & \textbf{0.111} & +299.20\% & 0.425 & \textbf{0.117} & +263.49\% \\
 & giz & 0.452 & \textbf{0.125} & +260.40\% & 0.458 & \textbf{0.142} & +222.80\% \\
 & tui & 0.427 & \textbf{0.304} & +40.55\% & 0.480 & \textbf{0.343} & +39.92\% \\
\end{longtable}
\end{footnotesize}
\clearpage
\twocolumn
\clearpage
\onecolumn
\begin{footnotesize}
\begin{longtable}{ccrrrrrr}
\caption{DER and WER of 19 European languages in 9 train sizes produced by 2 models. The lowest DER and WER scores out of the 2 models are boldfaced. DM\textsubscript{D}, OD\textsubscript{D}, DM\textsubscript{W}, OD\textsubscript{W} are shorthands for DiaMT\textsubscript{DER}, OnlyDia\textsubscript{DER}, DiaMT\textsubscript{WER}, and OnlyDia\textsubscript{WER}, respectively. pc(m1, m2) is the percentage change of model m1 over model m2. The lower the percentage change, the better the model m1 is compared to model m2.} \label{tab:der_wer_euro_exact_values} \\
\toprule
Size & Lang & DiaMT\textsubscript{DER} & OnlyDia\textsubscript{DER} & pc(DM\textsubscript{D},OD\textsubscript{D}) & DiaMT\textsubscript{WER} & OnlyDia\textsubscript{WER} & pc(DM\textsubscript{W},OD\textsubscript{W}) \\
\midrule
\endfirsthead
\caption[]{DER and WER of 19 European languages in 9 train sizes produced by 2 models. The lowest DER and WER scores out of the 2 models are boldfaced. DM\textsubscript{D}, OD\textsubscript{D}, DM\textsubscript{W}, OD\textsubscript{W} are shorthands for DiaMT\textsubscript{DER}, OnlyDia\textsubscript{DER}, DiaMT\textsubscript{WER}, and OnlyDia\textsubscript{WER}, respectively. pc(m1, m2) is the percentage change of model m1 over model m2. The lower the percentage change, the better the model m1 is compared to model m2.} \\
\toprule
Size & Lang & DiaMT\textsubscript{DER} & OnlyDia\textsubscript{DER} & pc(DM\textsubscript{D},OD\textsubscript{D}) & DiaMT\textsubscript{WER} & OnlyDia\textsubscript{WER} & pc(DM\textsubscript{W},OD\textsubscript{W}) \\
\midrule
\endhead
\midrule
\multicolumn{8}{r}{Continued on next page} \\
\midrule
\endfoot
\bottomrule
\endlastfoot
\multirow[c]{19}{*}{1k} & el & 0.557 & \textbf{0.356} & +56.76\% & 0.648 & \textbf{0.491} & +31.87\% \\
 & cs & 0.464 & \textbf{0.308} & +50.90\% & 0.593 & \textbf{0.445} & +33.17\% \\
 & da & 0.360 & \textbf{0.215} & +67.17\% & 0.430 & \textbf{0.300} & +43.41\% \\
 & de & 0.465 & \textbf{0.248} & +87.62\% & 0.560 & \textbf{0.376} & +49.04\% \\
 & es & 0.476 & \textbf{0.271} & +75.55\% & 0.557 & \textbf{0.389} & +43.11\% \\
 & et & 0.401 & \textbf{0.200} & +100.50\% & 0.519 & \textbf{0.321} & +61.78\% \\
 & fi & 0.403 & \textbf{0.207} & +94.85\% & 0.555 & \textbf{0.361} & +53.95\% \\
 & fr & 0.512 & \textbf{0.289} & +77.22\% & 0.603 & \textbf{0.418} & +44.40\% \\
 & hu & 0.512 & \textbf{0.335} & +52.95\% & 0.617 & \textbf{0.467} & +32.25\% \\
 & it & 0.446 & \textbf{0.212} & +110.38\% & 0.560 & \textbf{0.355} & +57.78\% \\
 & lt & 0.487 & \textbf{0.283} & +72.33\% & 0.617 & \textbf{0.434} & +42.25\% \\
 & lv & 0.542 & \textbf{0.285} & +89.86\% & 0.653 & \textbf{0.441} & +47.97\% \\
 & nl & 0.425 & \textbf{0.197} & +115.26\% & 0.501 & \textbf{0.308} & +62.79\% \\
 & pl & 0.497 & \textbf{0.240} & +107.08\% & 0.610 & \textbf{0.391} & +55.87\% \\
 & pt & 0.498 & \textbf{0.313} & +58.70\% & 0.593 & \textbf{0.434} & +36.61\% \\
 & ro & 0.508 & \textbf{0.270} & +88.41\% & 0.613 & \textbf{0.411} & +48.92\% \\
 & sk & 0.446 & \textbf{0.270} & +64.99\% & 0.575 & \textbf{0.412} & +39.42\% \\
 & sl & 0.383 & \textbf{0.185} & +107.25\% & 0.466 & \textbf{0.281} & +66.06\% \\
 & sv & 0.505 & \textbf{0.271} & +86.46\% & 0.578 & \textbf{0.380} & +51.94\% \\
\hdashline
\multirow[c]{19}{*}{2k} & el & 0.544 & \textbf{0.340} & +59.89\% & 0.640 & \textbf{0.474} & +34.98\% \\
 & cs & 0.497 & \textbf{0.251} & +97.77\% & 0.628 & \textbf{0.389} & +61.18\% \\
 & da & 0.396 & \textbf{0.182} & +117.72\% & 0.463 & \textbf{0.265} & +75.06\% \\
 & de & 0.456 & \textbf{0.236} & +93.19\% & 0.551 & \textbf{0.361} & +52.78\% \\
 & es & 0.568 & \textbf{0.208} & +173.83\% & 0.630 & \textbf{0.337} & +87.26\% \\
 & et & 0.459 & \textbf{0.186} & +146.18\% & 0.575 & \textbf{0.299} & +92.18\% \\
 & fi & 0.419 & \textbf{0.181} & +131.59\% & 0.578 & \textbf{0.328} & +76.25\% \\
 & fr & 0.565 & \textbf{0.222} & +154.53\% & 0.637 & \textbf{0.366} & +74.24\% \\
 & hu & 0.535 & \textbf{0.276} & +94.25\% & 0.637 & \textbf{0.419} & +52.18\% \\
 & it & 0.420 & \textbf{0.226} & +85.49\% & 0.540 & \textbf{0.357} & +51.58\% \\
 & lt & 0.510 & \textbf{0.216} & +136.15\% & 0.637 & \textbf{0.369} & +72.48\% \\
 & lv & 0.571 & \textbf{0.237} & +140.41\% & 0.677 & \textbf{0.395} & +71.26\% \\
 & nl & 0.384 & \textbf{0.168} & +128.80\% & 0.473 & \textbf{0.283} & +66.99\% \\
 & pl & 0.483 & \textbf{0.214} & +125.77\% & 0.605 & \textbf{0.357} & +69.64\% \\
 & pt & 0.534 & \textbf{0.239} & +123.01\% & 0.629 & \textbf{0.372} & +69.36\% \\
 & ro & 0.564 & \textbf{0.219} & +156.92\% & 0.651 & \textbf{0.369} & +76.32\% \\
 & sk & 0.518 & \textbf{0.234} & +121.75\% & 0.631 & \textbf{0.373} & +69.39\% \\
 & sl & 0.431 & \textbf{0.153} & +181.64\% & 0.513 & \textbf{0.239} & +114.51\% \\
 & sv & 0.445 & \textbf{0.227} & +95.89\% & 0.532 & \textbf{0.336} & +58.63\% \\
\hdashline
\multirow[c]{19}{*}{3k} & el & 0.644 & \textbf{0.255} & +152.68\% & 0.718 & \textbf{0.413} & +73.81\% \\
 & cs & 0.533 & \textbf{0.266} & +100.49\% & 0.663 & \textbf{0.402} & +65.06\% \\
 & da & 0.484 & \textbf{0.168} & +188.14\% & 0.533 & \textbf{0.254} & +110.12\% \\
 & de & 0.506 & \textbf{0.210} & +140.89\% & 0.598 & \textbf{0.336} & +78.08\% \\
 & es & 0.529 & \textbf{0.200} & +164.20\% & 0.601 & \textbf{0.330} & +82.04\% \\
 & et & 0.454 & \textbf{0.159} & +186.09\% & 0.577 & \textbf{0.273} & +110.97\% \\
 & fi & 0.452 & \textbf{0.166} & +172.87\% & 0.604 & \textbf{0.314} & +92.78\% \\
 & fr & 0.585 & \textbf{0.178} & +228.82\% & 0.657 & \textbf{0.334} & +96.61\% \\
 & hu & 0.610 & \textbf{0.253} & +140.64\% & 0.695 & \textbf{0.398} & +74.55\% \\
 & it & 0.499 & \textbf{0.168} & +197.40\% & 0.608 & \textbf{0.312} & +94.55\% \\
 & lt & 0.559 & \textbf{0.216} & +159.37\% & 0.678 & \textbf{0.371} & +82.78\% \\
 & lv & 0.556 & \textbf{0.252} & +120.35\% & 0.667 & \textbf{0.402} & +66.00\% \\
 & nl & 0.447 & \textbf{0.155} & +188.47\% & 0.526 & \textbf{0.271} & +93.84\% \\
 & pl & 0.461 & \textbf{0.195} & +136.07\% & 0.598 & \textbf{0.341} & +75.14\% \\
 & pt & 0.583 & \textbf{0.227} & +157.27\% & 0.666 & \textbf{0.360} & +84.88\% \\
 & ro & 0.577 & \textbf{0.228} & +152.74\% & 0.669 & \textbf{0.372} & +79.83\% \\
 & sk & 0.507 & \textbf{0.216} & +134.41\% & 0.630 & \textbf{0.358} & +75.81\% \\
 & sl & 0.427 & \textbf{0.161} & +164.45\% & 0.512 & \textbf{0.241} & +112.09\% \\
 & sv & 0.496 & \textbf{0.207} & +138.92\% & 0.577 & \textbf{0.321} & +79.45\% \\
\hdashline
\multirow[c]{3}{*}{4k} & el & 0.618 & \textbf{0.270} & +128.83\% & 0.705 & \textbf{0.427} & +65.19\% \\
 & cs & 0.604 & \textbf{0.258} & +133.97\% & 0.721 & \textbf{0.394} & +83.10\% \\
 & da & 0.481 & \textbf{0.170} & +182.69\% & 0.528 & \textbf{0.257} & +105.85\% \\
 \multirow[c]{16}{*}{4k}& de & 0.566 & \textbf{0.183} & +208.93\% & 0.648 & \textbf{0.317} & +104.64\% \\
 & es & 0.593 & \textbf{0.191} & +210.33\% & 0.655 & \textbf{0.324} & +102.32\% \\
 & et & 0.500 & \textbf{0.165} & +202.62\% & 0.617 & \textbf{0.280} & +120.53\% \\
 & fi & 0.529 & \textbf{0.176} & +200.90\% & 0.666 & \textbf{0.318} & +109.26\% \\
 & fr & 0.576 & \textbf{0.245} & +134.79\% & 0.657 & \textbf{0.382} & +71.82\% \\
 & hu & 0.632 & \textbf{0.263} & +139.90\% & 0.710 & \textbf{0.405} & +75.20\% \\
 & it & 0.541 & \textbf{0.199} & +171.22\% & 0.641 & \textbf{0.334} & +91.81\% \\
 & lt & 0.537 & \textbf{0.231} & +132.08\% & 0.667 & \textbf{0.376} & +77.63\% \\
 & lv & 0.564 & \textbf{0.232} & +143.00\% & 0.678 & \textbf{0.389} & +74.51\% \\
 & nl & 0.481 & \textbf{0.183} & +162.85\% & 0.557 & \textbf{0.292} & +90.47\% \\
 & pl & 0.536 & \textbf{0.224} & +139.66\% & 0.652 & \textbf{0.365} & +78.60\% \\
 & pt & 0.554 & \textbf{0.196} & +182.91\% & 0.643 & \textbf{0.336} & +91.18\% \\
 & ro & 0.661 & \textbf{0.203} & +226.52\% & 0.748 & \textbf{0.355} & +110.89\% \\
 & sk & 0.578 & \textbf{0.244} & +136.87\% & 0.689 & \textbf{0.376} & +83.33\% \\
 & sl & 0.473 & \textbf{0.152} & +210.13\% & 0.546 & \textbf{0.236} & +131.21\% \\
 & sv & 0.523 & \textbf{0.185} & +182.02\% & 0.600 & \textbf{0.303} & +98.17\% \\
\hdashline
\multirow[c]{19}{*}{5k} & el & 0.610 & \textbf{0.310} & +96.67\% & 0.699 & \textbf{0.453} & +54.31\% \\
 & cs & 0.625 & \textbf{0.278} & +125.01\% & 0.729 & \textbf{0.405} & +79.80\% \\
 & da & 0.550 & \textbf{0.176} & +212.84\% & 0.597 & \textbf{0.261} & +128.86\% \\
 & de & 0.511 & \textbf{0.168} & +203.34\% & 0.604 & \textbf{0.305} & +98.31\% \\
 & es & 0.625 & \textbf{0.234} & +167.17\% & 0.681 & \textbf{0.351} & +93.84\% \\
 & et & 0.497 & \textbf{0.145} & +241.71\% & 0.612 & \textbf{0.264} & +132.09\% \\
 & fi & 0.527 & \textbf{0.162} & +224.40\% & 0.668 & \textbf{0.309} & +115.86\% \\
 & fr & 0.605 & \textbf{0.220} & +175.28\% & 0.681 & \textbf{0.364} & +87.19\% \\
 & hu & 0.668 & \textbf{0.249} & +168.02\% & 0.740 & \textbf{0.393} & +88.11\% \\
 & it & 0.526 & \textbf{0.151} & +247.62\% & 0.632 & \textbf{0.299} & +111.17\% \\
 & lt & 0.545 & \textbf{0.202} & +169.47\% & 0.674 & \textbf{0.360} & +87.06\% \\
 & lv & 0.585 & \textbf{0.230} & +154.33\% & 0.696 & \textbf{0.383} & +81.78\% \\
 & nl & 0.502 & \textbf{0.163} & +207.59\% & 0.570 & \textbf{0.276} & +106.50\% \\
 & pl & 0.518 & \textbf{0.209} & +148.13\% & 0.641 & \textbf{0.353} & +81.60\% \\
 & pt & 0.569 & \textbf{0.182} & +213.21\% & 0.658 & \textbf{0.321} & +104.78\% \\
 & ro & 0.642 & \textbf{0.236} & +172.49\% & 0.726 & \textbf{0.379} & +91.80\% \\
 & sk & 0.629 & \textbf{0.263} & +138.78\% & 0.721 & \textbf{0.388} & +85.78\% \\
 & sl & 0.434 & \textbf{0.169} & +156.48\% & 0.521 & \textbf{0.246} & +111.92\% \\
 & sv & 0.512 & \textbf{0.214} & +139.54\% & 0.593 & \textbf{0.324} & +82.86\% \\
\hdashline
\multirow[c]{19}{*}{25k} & el & 0.405 & \textbf{0.084} & +382.09\% & 0.533 & \textbf{0.273} & +95.21\% \\
 & cs & 0.323 & \textbf{0.110} & +195.10\% & 0.469 & \textbf{0.245} & +91.38\% \\
 & da & 0.240 & \textbf{0.050} & +376.81\% & 0.322 & \textbf{0.145} & +122.50\% \\
 & de & 0.291 & \textbf{0.071} & +312.39\% & 0.408 & \textbf{0.210} & +94.08\% \\
 & es & 0.306 & \textbf{0.083} & +267.10\% & 0.417 & \textbf{0.226} & +84.85\% \\
 & et & 0.221 & \textbf{0.064} & +243.91\% & 0.340 & \textbf{0.162} & +110.21\% \\
 & fi & 0.262 & \textbf{0.071} & +269.04\% & 0.411 & \textbf{0.199} & +107.00\% \\
 & fr & 0.308 & \textbf{0.067} & +363.22\% & 0.436 & \textbf{0.231} & +89.02\% \\
 & hu & 0.378 & \textbf{0.127} & +198.30\% & 0.501 & \textbf{0.276} & +81.23\% \\
 & it & 0.295 & \textbf{0.050} & +486.40\% & 0.429 & \textbf{0.197} & +117.66\% \\
 & lt & 0.289 & \textbf{0.077} & +273.96\% & 0.442 & \textbf{0.217} & +103.03\% \\
 & lv & 0.301 & \textbf{0.108} & +178.02\% & 0.454 & \textbf{0.260} & +74.84\% \\
 & nl & 0.264 & \textbf{0.055} & +377.52\% & 0.360 & \textbf{0.178} & +102.51\% \\
 & pl & 0.250 & \textbf{0.066} & +277.89\% & 0.399 & \textbf{0.203} & +96.85\% \\
 & pt & 0.363 & \textbf{0.068} & +434.92\% & 0.469 & \textbf{0.216} & +116.84\% \\
 & ro & 0.320 & \textbf{0.083} & +285.20\% & 0.452 & \textbf{0.242} & +87.16\% \\
 & sk & 0.303 & \textbf{0.111} & +171.61\% & 0.444 & \textbf{0.244} & +82.14\% \\
 & sl & 0.200 & \textbf{0.044} & +350.37\% & 0.289 & \textbf{0.124} & +133.36\% \\
 & sv & 0.307 & \textbf{0.088} & +249.19\% & 0.406 & \textbf{0.204} & +99.51\% \\
\hdashline
\multirow[c]{6}{*}{125k} & el & 0.134 & \textbf{0.098} & +35.70\% & 0.305 & \textbf{0.275} & +11.00\% \\
 & cs & 0.125 & \textbf{0.081} & +54.39\% & 0.252 & \textbf{0.211} & +19.89\% \\
 & da & 0.067 & \textbf{0.015} & +360.99\% & 0.157 & \textbf{0.110} & +42.86\% \\
 & de & 0.085 & \textbf{0.025} & +243.59\% & 0.223 & \textbf{0.167} & +33.73\% \\
 & es & 0.047 & \textbf{0.028} & +68.79\% & 0.190 & \textbf{0.173} & +9.93\% \\
 & et & 0.062 & \textbf{0.020} & +202.85\% & 0.158 & \textbf{0.113} & +39.70\% \\
 \multirow[c]{12}{*}{125k}& fi & 0.078 & \textbf{0.051} & +52.76\% & 0.203 & \textbf{0.172} & +17.74\% \\
 & fr & 0.065 & \textbf{0.021} & +211.43\% & 0.228 & \textbf{0.185} & +22.76\% \\
 & hu & 0.111 & \textbf{0.071} & +55.06\% & 0.254 & \textbf{0.215} & +18.44\% \\
 & it & 0.084 & \textbf{0.015} & +443.37\% & 0.228 & \textbf{0.159} & +43.38\% \\
 & lt & 0.094 & \textbf{0.050} & +87.14\% & 0.236 & \textbf{0.178} & +32.82\% \\
 & lv & 0.098 & \textbf{0.072} & +35.96\% & 0.248 & \textbf{0.222} & +11.85\% \\
 & nl & 0.085 & \textbf{0.011} & +658.69\% & 0.201 & \textbf{0.135} & +48.47\% \\
 & pl & 0.067 & \textbf{0.022} & +213.24\% & 0.202 & \textbf{0.148} & +36.75\% \\
 & pt & 0.097 & \textbf{0.024} & +295.81\% & 0.239 & \textbf{0.172} & +39.29\% \\
 & ro & 0.114 & \textbf{0.037} & +210.51\% & 0.257 & \textbf{0.196} & +30.81\% \\
 & sk & 0.148 & \textbf{0.135} & +9.50\% & \textbf{0.268} & 0.271 & -0.93\% \\
 & sv & 0.081 & \textbf{0.026} & +215.25\% & 0.196 & \textbf{0.144} & +35.89\% \\
\hdashline
\multirow[c]{10}{*}{625k} & el & \textbf{0.026} & 0.046 & -42.14\% & \textbf{0.207} & 0.227 & -9.01\% \\
 & da & 0.014 & \textbf{0.011} & +20.91\% & 0.108 & \textbf{0.106} & +2.03\% \\
 & de & 0.029 & \textbf{0.017} & +67.84\% & 0.169 & \textbf{0.158} & +6.39\% \\
 & es & 0.021 & \textbf{0.016} & +32.11\% & 0.168 & \textbf{0.163} & +3.35\% \\
 & fi & \textbf{0.039} & 0.044 & -11.75\% & \textbf{0.156} & 0.164 & -5.10\% \\
 & fr & 0.023 & \textbf{0.015} & +54.78\% & 0.186 & \textbf{0.177} & +4.89\% \\
 & it & 0.022 & \textbf{0.013} & +65.54\% & 0.168 & \textbf{0.155} & +8.31\% \\
 & nl & 0.019 & \textbf{0.011} & +67.21\% & 0.143 & \textbf{0.135} & +5.25\% \\
 & pt & 0.025 & \textbf{0.016} & +52.92\% & 0.174 & \textbf{0.162} & +7.29\% \\
 & sv & 0.032 & \textbf{0.025} & +27.32\% & 0.149 & \textbf{0.143} & +4.51\% \\
\hdashline
\multirow[c]{9}{*}{1M} & el & \textbf{0.020} & 0.098 & -79.82\% & \textbf{0.198} & 0.276 & -28.02\% \\
 & de & 0.022 & \textbf{0.017} & +31.38\% & 0.163 & \textbf{0.158} & +3.09\% \\
 & es & 0.016 & \textbf{0.013} & +19.55\% & 0.163 & \textbf{0.160} & +1.93\% \\
 & fi & \textbf{0.028} & 0.052 & -46.13\% & \textbf{0.141} & 0.175 & -19.54\% \\
 & fr & 0.017 & \textbf{0.014} & +16.21\% & 0.180 & \textbf{0.176} & +2.61\% \\
 & it & \textbf{0.013} & 0.014 & -7.60\% & \textbf{0.157} & \textbf{0.157} & -0.13\% \\
 & nl & 0.013 & \textbf{0.012} & +9.09\% & 0.137 & \textbf{0.135} & +1.05\% \\
 & pt & \textbf{0.018} & 0.020 & -11.33\% & \textbf{0.166} & 0.167 & -0.37\% \\
 & sv & 0.019 & \textbf{0.018} & +6.08\% & 0.137 & \textbf{0.135} & +1.84\% \\
\end{longtable}
\end{footnotesize}
\clearpage
\section{Complexity Metrics}\label{sec:complexity_metrics}
We propose two classes of complexity metrics to assess the complexity of the diacritical system of a given language. The first class is based on the ratio of diacritics and character/word/sentence. The second class is based on the entropy of combinations of diacritic(s) and characters, measuring from the perspective of probability distribution. For the first class, we propose diacritized character ratio (\textbf{DCR}), diacritized word ratio (\textbf{DWR}), diacritized base character ratio (\textbf{DBR}), and diacritized word sentence ratio (\textbf{DWSR}). For the second class, we propose average entropy of diacritics (\textbf{AED}), and weighted average entropy of diacritics (\textbf{WAED}). Their definition can be seen in Table~\ref{tab:complexity_metrics_definitions_formula_appendix}. An example corpus and the computation of values of complexity metrics is given in Table~\ref{tab:aed_waed_ex}.

\begin{table*}[h]
\centering

 \begin{tabular}[t]{cp{13.2cm}}
 \toprule
 \textbf{Metric} & \textbf{Definition} \\
\midrule
DCR & Proportion of characters that carry diacritic(s) out of all characters.\\

DWR & Proportion of words with at least a character carrying diacritic(s) out of all words.\\ 

DBR & Average number of variants (including itself) of each base character.\\ 

DWSR & Average number of words with at least a character carrying diacritic(s) per sentence.\\
\midrule

AED & Average entropy of the distributions of each base character's variant(s) and itself.\\

WAED & Weighted AED with weight being the proportion of the number of occurrence of each base character out of that of all base character(s). \\ \hline

\end{tabular}
\caption{Definitions of Proposed Complexity Metrics. }  \label{tab:complexity_metrics_definitions_formula_appendix}
\end{table*}

\begin{table*}[h]
\centering
 \begin{tabular}[t]{cS}
 \toprule
Corpus & Sh\underline{ë} w\underline{a}nts \underline{â}n \underline{â}ppl\underline{e}. \newline
I drink coconut w\underline{ä}t\underline{ë}r for fun.\\[0.06cm]
\midrule
DCR & $\frac{5}{39} = 0.128$\\[0.1cm]
DWR & $\frac{4}{10} = 0.4$\\[0.1cm]
DBR & $\frac{5}{2} = 2.5$\\[0.1cm]
DWSR & $\frac{4}{2} = 2$\\[0.1cm]
\midrule
P(X) & P(a):\{a:0.25, â:0.5, ä:0.25\}\newline P(e):\{e:0.33, ë:0.67\}\\ [0.1cm]

H(P(X)) & H(P(a)) = 1.05; H(P(e)) = 0.63 \\[0.1cm]

AED & $ \frac{1}{2} \times H(P(a)) + \frac{1}{2} \times H(P(e)) = 0.845$ \\[0.1cm]

WAED & $\frac{4}{7} \times H(P(a)) + \frac{3}{7} \times H(P(e))$ = 0.875\\
\bottomrule
\end{tabular}
\caption{An example of computing complexity metrics with a mock corpus where base characters are underlined. $P(\cdot)$ represents probability distribution. $H(\cdot)$ represents entropy.}  \label{tab:aed_waed_ex}
\end{table*}

In Table~\ref{tab:aed_waed_ex}, WAED is larger than AED because the total number of occurrences of the base character `a' is larger than `e' and therefore the weight ($\frac{4}{7}$) for its entropy is higher than that for `e' ($\frac{3}{7}$) which draws the weighted average closer toward the entropy of `a'. In contrast, AED gives even weight to each base character which is $\frac{1}{2}$ in this example and does not take frequency of each base character into consideration. WAED takes distribution of the language data into consideration when measuring the complexity of a diacritical system.

\begin{table*}
\centering
\begin{tabular}{lccccc}
\toprule
Stat/Train Size &          1k &          2k &          3k &          4k &          5k \\
\midrule
p(DCR,DER)  &  0.613 / <.05 &  0.581 / <.05 &  0.612 / <.05 &  0.468 / <.05 &  0.487 / <.05 \\
s(DCR,DER)  &  0.681 / <.05 &  0.610 / <.05 &  0.641 / <.05 &  0.567 / <.05 &  0.564 / <.05 \\
k(DCR,DER)  &  0.485 / <.05 &  0.444 / <.05 &  0.446 / <.05 &  0.396 / <.05 &  0.417 / <.05 \\
p(DWR,DER)  &  0.608 / <.05 &  0.581 / <.05 &  0.621 / <.05 &  0.476 / <.05 &  0.500 / <.05 \\
s(DWR,DER)  &  0.690 / <.05 &  0.620 / <.05 &  0.645 / <.05 &  0.573 / <.05 &  0.567 / <.05 \\
k(DWR,DER)  &  0.491 / <.05 &  0.444 / <.05 &  0.446 / <.05 &  0.396 / <.05 &  0.424 / <.05 \\
p(DBR,DER)  &  0.301 / >.05 &  0.343 / <.05 &  0.177 / >.05 &  0.172 / >.05 &  0.263 / >.05 \\
s(DBR,DER)  &  0.367 / <.05 &  0.345 / <.05 &  0.169 / >.05 &  0.235 / >.05 &  0.262 / >.05 \\
k(DBR,DER)  &  0.276 / <.05 &  0.246 / <.05 &  0.120 / >.05 &  0.200 / >.05 &  0.202 / >.05 \\
p(DWSR,DER) &  0.616 / <.05 &  0.620 / <.05 &  0.648 / <.05 &  0.505 / <.05 &  0.514 / <.05 \\
s(DWSR,DER) &  0.726 / <.05 &  0.677 / <.05 &  0.694 / <.05 &  0.617 / <.05 &  0.613 / <.05 \\
k(DWSR,DER) &  0.539 / <.05 &  0.520 / <.05 &  0.503 / <.05 &  0.460 / <.05 &  0.474 / <.05 \\
p(AED,DER)  &  0.566 / <.05 &  0.555 / <.05 &  0.528 / <.05 &  0.386 / <.05 &  0.406 / <.05 \\
s(AED,DER)  &  0.626 / <.05 &  0.564 / <.05 &  0.521 / <.05 &  0.481 / <.05 &  0.420 / <.05 \\
k(AED,DER)  &  0.453 / <.05 &  0.425 / <.05 &  0.359 / <.05 &  0.332 / <.05 &  0.306 / <.05 \\
p(WAED,DER) &  0.522 / <.05 &  0.498 / <.05 &  0.517 / <.05 &  0.371 / <.05 &  0.391 / <.05 \\
s(WAED,DER) &  0.548 / <.05 &  0.479 / <.05 &  0.513 / <.05 &  0.453 / <.05 &  0.410 / <.05 \\
k(WAED,DER) &  0.389 / <.05 &  0.348 / <.05 &  0.342 / <.05 &  0.309 / <.05 &  0.303 / <.05 \\
p(DCR,WER)  &  0.737 / <.05 &  0.696 / <.05 &  0.750 / <.05 &  0.673 / <.05 &  0.658 / <.05 \\
s(DCR,WER)  &  0.701 / <.05 &  0.642 / <.05 &  0.724 / <.05 &  0.676 / <.05 &  0.620 / <.05 \\
k(DCR,WER)  &  0.513 / <.05 &  0.458 / <.05 &  0.536 / <.05 &  0.482 / <.05 &  0.442 / <.05 \\
p(DWR,WER)  &  0.738 / <.05 &  0.702 / <.05 &  0.762 / <.05 &  0.684 / <.05 &  0.673 / <.05 \\
s(DWR,WER)  &  0.710 / <.05 &  0.654 / <.05 &  0.729 / <.05 &  0.683 / <.05 &  0.624 / <.05 \\
k(DWR,WER)  &  0.519 / <.05 &  0.464 / <.05 &  0.536 / <.05 &  0.482 / <.05 &  0.449 / <.05 \\
p(DBR,WER)  &  0.419 / <.05 &  0.428 / <.05 &  0.333 / >.05 &  0.331 / >.05 &  0.366 / <.05 \\
s(DBR,WER)  &  0.405 / <.05 &  0.418 / <.05 &  0.299 / >.05 &  0.356 / <.05 &  0.331 / >.05 \\
k(DBR,WER)  &  0.299 / <.05 &  0.292 / <.05 &  0.204 / >.05 &  0.284 / <.05 &  0.256 / <.05 \\
p(DWSR,WER) &  0.763 / <.05 &  0.758 / <.05 &  0.811 / <.05 &  0.736 / <.05 &  0.713 / <.05 \\
s(DWSR,WER) &  0.763 / <.05 &  0.727 / <.05 &  0.794 / <.05 &  0.745 / <.05 &  0.685 / <.05 \\
k(DWSR,WER) &  0.580 / <.05 &  0.550 / <.05 &  0.607 / <.05 &  0.560 / <.05 &  0.519 / <.05 \\
p(AED,WER)  &  0.693 / <.05 &  0.663 / <.05 &  0.668 / <.05 &  0.588 / <.05 &  0.574 / <.05 \\
s(AED,WER)  &  0.667 / <.05 &  0.616 / <.05 &  0.622 / <.05 &  0.593 / <.05 &  0.512 / <.05 \\
k(AED,WER)  &  0.494 / <.05 &  0.452 / <.05 &  0.459 / <.05 &  0.432 / <.05 &  0.351 / <.05 \\
p(WAED,WER) &  0.660 / <.05 &  0.623 / <.05 &  0.673 / <.05 &  0.591 / <.05 &  0.575 / <.05 \\
s(WAED,WER) &  0.590 / <.05 &  0.541 / <.05 &  0.625 / <.05 &  0.592 / <.05 &  0.516 / <.05 \\
k(WAED,WER) &  0.431 / <.05 &  0.394 / <.05 &  0.435 / <.05 &  0.422 / <.05 &  0.355 / <.05 \\
\bottomrule
\end{tabular}
\caption{The Pearson (p), Spearman (s), and Kendall (k) correlation statistics and p value between complexity metrics (DCR, DWR, DBR, AED, WAED) and performance metrics (DER, WER) produced by OnlyDia model for African languages.}\label{tab:leveled_afri_complexity_performance_correlations}
\end{table*}

\begin{landscape}
\begin{table}
\small
\centering
\begin{tabular}{lccccccccc}
\toprule
Stat/Train Size &        1k    &        2k    &        3k    &        4k    &        5k    &        25k   &        125k  &        625k  &        1M \\
\midrule
p(DCR,DER)  &   0.694 / <.05 &   0.636 / <.05 &   0.827 / <.05 &   0.778 / <.05 &   0.800 / <.05 &   0.857 / <.05 &   0.859 / <.05 &   0.867 / <.05 &   0.885 / <.05 \\
s(DCR,DER)  &   0.618 / <.05 &   0.596 / <.05 &   0.786 / <.05 &   0.719 / <.05 &   0.737 / <.05 &   0.814 / <.05 &   0.892 / <.05 &   0.884 / <.05 &   0.879 / <.05 \\
k(DCR,DER)  &   0.465 / <.05 &   0.427 / <.05 &   0.618 / <.05 &   0.516 / <.05 &   0.544 / <.05 &   0.649 / <.05 &   0.734 / <.05 &   0.750 / <.05 &   0.761 / <.05 \\
p(DWR,DER)  &   0.688 / <.05 &   0.633 / <.05 &   0.823 / <.05 &   0.776 / <.05 &   0.793 / <.05 &   0.858 / <.05 &   0.859 / <.05 &   0.879 / <.05 &   0.892 / <.05 \\
s(DWR,DER)  &   0.601 / <.05 &   0.579 / <.05 &   0.768 / <.05 &   0.670 / <.05 &   0.698 / <.05 &   0.807 / <.05 &   0.890 / <.05 &   0.884 / <.05 &   0.879 / <.05 \\
k(DWR,DER)  &   0.465 / <.05 &   0.415 / <.05 &   0.582 / <.05 &   0.504 / <.05 &   0.509 / <.05 &   0.637 / <.05 &   0.721 / <.05 &   0.750 / <.05 &   0.761 / <.05 \\
p(DBR,DER)  &   0.022 / >.05 &  -0.181 / >.05 &  -0.245 / >.05 &  -0.220 / >.05 &  -0.448 / >.05 &  -0.083 / >.05 &  -0.124 / >.05 &  -0.544 / >.05 &  -0.677 / <.05 \\
s(DBR,DER)  &   0.080 / >.05 &   0.069 / >.05 &  -0.193 / >.05 &  -0.121 / >.05 &  -0.306 / >.05 &  -0.162 / >.05 &  -0.306 / >.05 &  -0.413 / >.05 &  -0.445 / >.05 \\
k(DBR,DER)  &   0.083 / >.05 &   0.071 / >.05 &  -0.142 / >.05 &  -0.078 / >.05 &  -0.241 / >.05 &  -0.179 / >.05 &  -0.224 / >.05 &  -0.368 / >.05 &  -0.343 / >.05 \\
p(DWSR,DER) &   0.732 / <.05 &   0.700 / <.05 &   0.838 / <.05 &   0.805 / <.05 &   0.832 / <.05 &   0.854 / <.05 &   0.875 / <.05 &   0.859 / <.05 &   0.911 / <.05 \\
s(DWSR,DER) &   0.659 / <.05 &   0.621 / <.05 &   0.797 / <.05 &   0.736 / <.05 &   0.765 / <.05 &   0.808 / <.05 &   0.904 / <.05 &   0.884 / <.05 &   0.879 / <.05 \\
k(DWSR,DER) &   0.500 / <.05 &   0.474 / <.05 &   0.641 / <.05 &   0.563 / <.05 &   0.591 / <.05 &   0.649 / <.05 &   0.748 / <.05 &   0.750 / <.05 &   0.761 / <.05 \\
p(AED,DER)  &   0.577 / <.05 &   0.499 / <.05 &   0.703 / <.05 &   0.654 / <.05 &   0.610 / <.05 &   0.783 / <.05 &   0.735 / <.05 &   0.359 / >.05 &   0.112 / >.05 \\
s(AED,DER)  &   0.515 / <.05 &   0.530 / <.05 &   0.687 / <.05 &   0.596 / <.05 &   0.591 / <.05 &   0.711 / <.05 &   0.793 / <.05 &   0.366 / >.05 &   0.201 / >.05 \\
k(AED,DER)  &   0.335 / <.05 &   0.333 / <.05 &   0.512 / <.05 &   0.446 / <.05 &   0.450 / <.05 &   0.543 / <.05 &   0.616 / <.05 &   0.250 / >.05 &   0.085 / >.05 \\
p(WAED,DER) &   0.787 / <.05 &   0.733 / <.05 &   0.826 / <.05 &   0.777 / <.05 &   0.807 / <.05 &   0.863 / <.05 &   0.835 / <.05 &   0.760 / <.05 &   0.783 / <.05 \\
s(WAED,DER) &   0.699 / <.05 &   0.688 / <.05 &   0.806 / <.05 &   0.777 / <.05 &   0.765 / <.05 &   0.833 / <.05 &   0.869 / <.05 &   0.817 / <.05 &   0.845 / <.05 \\
k(WAED,DER) &   0.559 / <.05 &   0.544 / <.05 &   0.629 / <.05 &   0.610 / <.05 &   0.591 / <.05 &   0.661 / <.05 &   0.721 / <.05 &   0.659 / <.05 &   0.704 / <.05 \\
p(DCR,WER)  &   0.754 / <.05 &   0.691 / <.05 &   0.797 / <.05 &   0.749 / <.05 &   0.807 / <.05 &   0.728 / <.05 &   0.787 / <.05 &   0.789 / <.05 &   0.828 / <.05 \\
s(DCR,WER)  &   0.789 / <.05 &   0.771 / <.05 &   0.821 / <.05 &   0.781 / <.05 &   0.849 / <.05 &   0.781 / <.05 &   0.793 / <.05 &   0.697 / <.05 &   0.636 / >.05 \\
k(DCR,WER)  &   0.610 / <.05 &   0.571 / <.05 &   0.610 / <.05 &   0.587 / <.05 &   0.661 / <.05 &   0.567 / <.05 &   0.577 / <.05 &   0.556 / <.05 &   0.592 / <.05 \\
p(DWR,WER)  &   0.751 / <.05 &   0.689 / <.05 &   0.794 / <.05 &   0.748 / <.05 &   0.802 / <.05 &   0.726 / <.05 &   0.784 / <.05 &   0.786 / <.05 &   0.827 / <.05 \\
s(DWR,WER)  &   0.778 / <.05 &   0.746 / <.05 &   0.804 / <.05 &   0.739 / <.05 &   0.814 / <.05 &   0.746 / <.05 &   0.772 / <.05 &   0.697 / <.05 &   0.636 / >.05 \\
k(DWR,WER)  &   0.610 / <.05 &   0.559 / <.05 &   0.598 / <.05 &   0.575 / <.05 &   0.649 / <.05 &   0.556 / <.05 &   0.564 / <.05 &   0.556 / <.05 &   0.592 / <.05 \\
p(DBR,WER)  &   0.049 / >.05 &  -0.086 / >.05 &  -0.176 / >.05 &  -0.149 / >.05 &  -0.314 / >.05 &  -0.286 / >.05 &  -0.213 / >.05 &  -0.130 / >.05 &  -0.434 / >.05 \\
s(DBR,WER)  &  -0.005 / >.05 &  -0.069 / >.05 &  -0.163 / >.05 &  -0.150 / >.05 &  -0.241 / >.05 &  -0.249 / >.05 &  -0.202 / >.05 &  -0.085 / >.05 &  -0.176 / >.05 \\
k(DBR,WER)  &   0.006 / >.05 &  -0.048 / >.05 &  -0.112 / >.05 &  -0.102 / >.05 &  -0.194 / >.05 &  -0.243 / >.05 &  -0.172 / >.05 &   0.000 / >.05 &  -0.086 / >.05 \\
p(DWSR,WER) &   0.785 / <.05 &   0.740 / <.05 &   0.818 / <.05 &   0.780 / <.05 &   0.841 / <.05 &   0.763 / <.05 &   0.825 / <.05 &   0.818 / <.05 &   0.872 / <.05 \\
s(DWSR,WER) &   0.821 / <.05 &   0.785 / <.05 &   0.839 / <.05 &   0.792 / <.05 &   0.874 / <.05 &   0.791 / <.05 &   0.812 / <.05 &   0.697 / <.05 &   0.636 / >.05 \\
k(DWSR,WER) &   0.645 / <.05 &   0.606 / <.05 &   0.657 / <.05 &   0.610 / <.05 &   0.731 / <.05 &   0.591 / <.05 &   0.643 / <.05 &   0.556 / <.05 &   0.592 / <.05 \\
p(AED,WER)  &   0.622 / <.05 &   0.540 / <.05 &   0.631 / <.05 &   0.593 / <.05 &   0.604 / <.05 &   0.585 / <.05 &   0.578 / <.05 &  -0.143 / >.05 &  -0.152 / >.05 \\
s(AED,WER)  &   0.695 / <.05 &   0.673 / <.05 &   0.717 / <.05 &   0.626 / <.05 &   0.658 / <.05 &   0.642 / <.05 &   0.599 / <.05 &  -0.297 / >.05 &  -0.293 / >.05 \\
k(AED,WER)  &   0.504 / <.05 &   0.453 / <.05 &   0.528 / <.05 &   0.446 / <.05 &   0.497 / <.05 &   0.450 / <.05 &   0.407 / <.05 &  -0.156 / >.05 &  -0.197 / >.05 \\
p(WAED,WER) &   0.819 / <.05 &   0.755 / <.05 &   0.802 / <.05 &   0.758 / <.05 &   0.834 / <.05 &   0.784 / <.05 &   0.775 / <.05 &   0.800 / <.05 &   0.788 / <.05 \\
s(WAED,WER) &   0.821 / <.05 &   0.783 / <.05 &   0.817 / <.05 &   0.798 / <.05 &   0.860 / <.05 &   0.823 / <.05 &   0.804 / <.05 &   0.685 / <.05 &   0.603 / >.05 \\
k(WAED,WER) &   0.657 / <.05 &   0.618 / <.05 &   0.622 / <.05 &   0.633 / <.05 &   0.708 / <.05 &   0.649 / <.05 &   0.616 / <.05 &   0.556 / <.05 &   0.535 / <.05 \\
\bottomrule
\end{tabular}
\caption{The Pearson (p), Spearman (s), and Kendall (k) correlation statistics and p value between complexity metrics (DCR, DWR, DBR, AED, WAED) and performance metrics (DER, WER) produced by OnlyDia model for European languages.}\label{tab:leveled_euro_complexity_performance_correlations}
\end{table}
\end{landscape}

\clearpage
\onecolumn
\begin{footnotesize}
\begin{longtable}{ccrrrrrr}
\caption{Complexity metrics for diacritical system of each African language at 5 train sizes. For a given language, a metric may occasionally have identical values throughout different train sizes because they are rounded to 3 digits.} \label{tab:leveled_dia_complexity_metrics_afri} \\
\toprule
 &  & DCR & DWR & DBR & DWSR & AED & WAED \\
Lang & Size &  &  &  &  &  &  \\
\midrule
\endfirsthead
\caption[]{Complexity metrics for diacritical system of each African language at 5 train sizes. For a given language, a metric may occasionally have identical values throughout different train sizes because they are rounded to 3 digits.} \\
\toprule
 &  & DCR & DWR & DBR & DWSR & AED & WAED \\
Lang & Size &  &  &  &  &  &  \\
\midrule
\endhead
\midrule
\multicolumn{8}{r}{Continued on next page} \\
\midrule
\endfoot
\bottomrule
\endlastfoot
\multirow[c]{5}{*}{bex} & 1k & 0.090 & 0.067 & 2.000 & 11.426 & 0.563 & 0.562 \\
 & 2k & 0.091 & 0.068 & 2.000 & 11.515 & 0.564 & 0.564 \\
 & 3k & 0.091 & 0.068 & 2.000 & 11.511 & 0.565 & 0.565 \\
 & 4k & 0.090 & 0.068 & 2.000 & 11.452 & 0.564 & 0.564 \\
 & 5k & 0.090 & 0.067 & 2.000 & 11.348 & 0.563 & 0.563 \\
\hdashline
\multirow[c]{5}{*}{fon} & 1k & 0.193 & 0.141 & 3.286 & 22.280 & 0.794 & 0.795 \\
 & 2k & 0.193 & 0.141 & 3.286 & 22.522 & 0.793 & 0.794 \\
 & 3k & 0.194 & 0.142 & 3.286 & 22.645 & 0.794 & 0.795 \\
 & 4k & 0.194 & 0.142 & 3.286 & 22.541 & 0.794 & 0.795 \\
 & 5k & 0.194 & 0.141 & 3.286 & 22.474 & 0.794 & 0.795 \\
\hdashline
\multirow[c]{5}{*}{mkl} & 1k & 0.072 & 0.052 & 3.556 & 6.665 & 0.334 & 0.398 \\
 & 2k & 0.072 & 0.052 & 3.556 & 6.637 & 0.332 & 0.397 \\
 & 3k & 0.072 & 0.053 & 3.556 & 6.646 & 0.332 & 0.398 \\
 & 4k & 0.072 & 0.053 & 3.556 & 6.642 & 0.332 & 0.398 \\
 & 5k & 0.072 & 0.052 & 3.556 & 6.629 & 0.332 & 0.397 \\
\hdashline
\multirow[c]{5}{*}{mnf} & 1k & 0.198 & 0.151 & 4.750 & 23.874 & 0.862 & 0.871 \\
 & 2k & 0.199 & 0.151 & 4.750 & 23.899 & 0.862 & 0.870 \\
 & 3k & 0.199 & 0.151 & 4.750 & 23.960 & 0.862 & 0.870 \\
 & 4k & 0.199 & 0.151 & 4.750 & 23.805 & 0.862 & 0.871 \\
 & 5k & 0.199 & 0.150 & 4.750 & 23.883 & 0.862 & 0.870 \\
\hdashline
\multirow[c]{5}{*}{bud} & 1k & 0.140 & 0.109 & 3.800 & 15.894 & 0.495 & 0.615 \\
 & 2k & 0.140 & 0.109 & 3.800 & 15.985 & 0.496 & 0.615 \\
 & 3k & 0.140 & 0.108 & 3.636 & 15.939 & 0.448 & 0.601 \\
 & 4k & 0.140 & 0.109 & 3.636 & 15.927 & 0.450 & 0.602 \\
 & 5k & 0.140 & 0.108 & 3.636 & 15.906 & 0.450 & 0.602 \\
\hdashline
\multirow[c]{5}{*}{eza} & 1k & 0.101 & 0.077 & 3.800 & 14.469 & 0.422 & 0.463 \\
 & 2k & 0.101 & 0.077 & 3.800 & 14.710 & 0.423 & 0.463 \\
 & 3k & 0.101 & 0.076 & 3.800 & 14.808 & 0.420 & 0.461 \\
 & 4k & 0.101 & 0.077 & 3.800 & 14.794 & 0.422 & 0.462 \\
 & 5k & 0.101 & 0.076 & 3.800 & 14.772 & 0.422 & 0.462 \\
\hdashline
\multirow[c]{5}{*}{sig} & 1k & 0.004 & 0.003 & 2.000 & 0.440 & 0.099 & 0.099 \\
 & 2k & 0.004 & 0.003 & 2.000 & 0.476 & 0.052 & 0.084 \\
 & 3k & 0.004 & 0.003 & 2.000 & 0.479 & 0.052 & 0.085 \\
 & 4k & 0.004 & 0.003 & 2.000 & 0.485 & 0.053 & 0.086 \\
 & 5k & 0.004 & 0.003 & 2.000 & 0.488 & 0.053 & 0.086 \\
\hdashline
\multirow[c]{5}{*}{bqc} & 1k & 0.195 & 0.147 & 3.300 & 13.789 & 0.661 & 0.812 \\
 & 2k & 0.194 & 0.146 & 3.300 & 13.683 & 0.659 & 0.811 \\
 & 3k & 0.194 & 0.146 & 3.300 & 13.670 & 0.657 & 0.809 \\
 & 4k & 0.193 & 0.145 & 3.300 & 13.600 & 0.656 & 0.809 \\
 & 5k & 0.194 & 0.144 & 3.300 & 13.650 & 0.656 & 0.809 \\
\hdashline
\multirow[c]{5}{*}{kia} & 1k & 0.022 & 0.015 & 3.400 & 1.911 & 0.184 & 0.212 \\
 & 2k & 0.022 & 0.016 & 3.600 & 1.944 & 0.189 & 0.214 \\
 & 3k & 0.022 & 0.015 & 3.800 & 1.917 & 0.189 & 0.213 \\
 & 4k & 0.022 & 0.016 & 4.200 & 1.939 & 0.190 & 0.215 \\
 & 5k & 0.022 & 0.015 & 4.200 & 1.919 & 0.189 & 0.214 \\
\hdashline
\multirow[c]{5}{*}{soy} & 1k & 0.123 & 0.096 & 2.909 & 13.394 & 0.457 & 0.488 \\
 & 2k & 0.122 & 0.095 & 2.909 & 13.400 & 0.456 & 0.488 \\
 & 3k & 0.122 & 0.095 & 2.909 & 13.469 & 0.455 & 0.487 \\
 & 4k & 0.122 & 0.095 & 2.909 & 13.455 & 0.454 & 0.487 \\
 & 5k & 0.122 & 0.095 & 2.909 & 13.471 & 0.455 & 0.487 \\
\hdashline
\multirow[c]{5}{*}{nnw} & 1k & 0.118 & 0.082 & 2.857 & 13.720 & 0.457 & 0.507 \\
 & 2k & 0.118 & 0.082 & 2.857 & 13.759 & 0.460 & 0.508 \\
 & 3k & 0.117 & 0.082 & 2.857 & 13.789 & 0.457 & 0.508 \\
 & 4k & 0.118 & 0.082 & 2.929 & 13.774 & 0.459 & 0.509 \\
 & 5k & 0.118 & 0.081 & 2.929 & 13.791 & 0.456 & 0.509 \\
\hdashline
\multirow[c]{5}{*}{sag} & 1k & 0.014 & 0.010 & 3.000 & 1.586 & 0.127 & 0.128 \\
 & 2k & 0.014 & 0.010 & 3.250 & 1.592 & 0.127 & 0.128 \\
 & 3k & 0.014 & 0.010 & 3.250 & 1.617 & 0.129 & 0.130 \\
 & 4k & 0.014 & 0.010 & 3.250 & 1.621 & 0.130 & 0.130 \\
 & 5k & 0.014 & 0.010 & 3.250 & 1.629 & 0.131 & 0.131 \\
\hdashline
csk & 1k & 0.036 & 0.030 & 2.000 & 4.723 & 0.207 & 0.205 \\
\multirow[c]{4}{*}{csk} & 2k & 0.036 & 0.030 & 2.000 & 4.700 & 0.207 & 0.205 \\
 & 3k & 0.036 & 0.029 & 2.000 & 4.685 & 0.206 & 0.205 \\
 & 4k & 0.036 & 0.030 & 2.000 & 4.712 & 0.207 & 0.205 \\
 & 5k & 0.037 & 0.030 & 2.000 & 4.718 & 0.208 & 0.206 \\
\hdashline
\multirow[c]{5}{*}{izz} & 1k & 0.103 & 0.078 & 3.429 & 13.685 & 0.305 & 0.411 \\
 & 2k & 0.103 & 0.079 & 3.429 & 13.705 & 0.303 & 0.409 \\
 & 3k & 0.104 & 0.079 & 3.571 & 13.738 & 0.304 & 0.410 \\
 & 4k & 0.103 & 0.078 & 3.571 & 13.667 & 0.303 & 0.410 \\
 & 5k & 0.103 & 0.078 & 3.571 & 13.611 & 0.304 & 0.409 \\
\hdashline
\multirow[c]{5}{*}{bum} & 1k & 0.084 & 0.062 & 2.000 & 7.378 & 0.363 & 0.445 \\
 & 2k & 0.084 & 0.062 & 2.000 & 7.445 & 0.364 & 0.445 \\
 & 3k & 0.084 & 0.062 & 2.000 & 7.501 & 0.364 & 0.445 \\
 & 4k & 0.084 & 0.062 & 2.000 & 7.477 & 0.366 & 0.446 \\
 & 5k & 0.084 & 0.061 & 2.000 & 7.458 & 0.366 & 0.446 \\
\hdashline
\multirow[c]{5}{*}{gvl} & 1k & 0.075 & 0.055 & 3.000 & 9.155 & 0.259 & 0.504 \\
 & 2k & 0.076 & 0.056 & 2.875 & 9.216 & 0.229 & 0.502 \\
 & 3k & 0.076 & 0.055 & 2.700 & 9.219 & 0.183 & 0.452 \\
 & 4k & 0.076 & 0.056 & 2.700 & 9.248 & 0.183 & 0.452 \\
 & 5k & 0.076 & 0.055 & 2.700 & 9.209 & 0.182 & 0.452 \\
\hdashline
\multirow[c]{5}{*}{ndz} & 1k & 0.258 & 0.192 & 3.667 & 42.915 & 0.965 & 1.024 \\
 & 2k & 0.258 & 0.192 & 3.667 & 42.549 & 0.965 & 1.024 \\
 & 3k & 0.258 & 0.192 & 3.667 & 42.994 & 0.966 & 1.024 \\
 & 4k & 0.258 & 0.192 & 3.667 & 42.987 & 0.966 & 1.024 \\
 & 5k & 0.258 & 0.191 & 3.667 & 42.835 & 0.966 & 1.024 \\
\hdashline
\multirow[c]{5}{*}{lip} & 1k & 0.021 & 0.016 & 2.500 & 2.416 & 0.167 & 0.175 \\
 & 2k & 0.021 & 0.016 & 2.444 & 2.418 & 0.150 & 0.164 \\
 & 3k & 0.021 & 0.016 & 2.667 & 2.415 & 0.150 & 0.165 \\
 & 4k & 0.021 & 0.016 & 2.667 & 2.422 & 0.151 & 0.165 \\
 & 5k & 0.021 & 0.016 & 2.667 & 2.408 & 0.150 & 0.164 \\
\hdashline
\multirow[c]{5}{*}{ken} & 1k & 0.119 & 0.093 & 3.800 & 14.357 & 0.630 & 0.588 \\
 & 2k & 0.119 & 0.094 & 3.800 & 14.292 & 0.629 & 0.589 \\
 & 3k & 0.119 & 0.094 & 3.800 & 14.356 & 0.630 & 0.590 \\
 & 4k & 0.119 & 0.094 & 3.800 & 14.337 & 0.631 & 0.590 \\
 & 5k & 0.119 & 0.093 & 3.800 & 14.291 & 0.631 & 0.590 \\
\hdashline
\multirow[c]{5}{*}{gid} & 1k & 0.001 & 0.001 & 2.250 & 0.070 & 0.018 & 0.016 \\
 & 2k & 0.001 & 0.001 & 2.250 & 0.076 & 0.019 & 0.016 \\
 & 3k & 0.001 & 0.001 & 2.250 & 0.075 & 0.018 & 0.016 \\
 & 4k & 0.001 & 0.001 & 2.250 & 0.074 & 0.018 & 0.016 \\
 & 5k & 0.001 & 0.001 & 2.250 & 0.075 & 0.018 & 0.016 \\
\hdashline
\multirow[c]{5}{*}{gng} & 1k & 0.047 & 0.034 & 3.000 & 4.666 & 0.283 & 0.299 \\
 & 2k & 0.047 & 0.033 & 3.000 & 4.612 & 0.281 & 0.297 \\
 & 3k & 0.047 & 0.034 & 3.000 & 4.622 & 0.280 & 0.298 \\
 & 4k & 0.047 & 0.033 & 3.000 & 4.566 & 0.279 & 0.297 \\
 & 5k & 0.047 & 0.033 & 3.000 & 4.541 & 0.278 & 0.296 \\
\hdashline
\multirow[c]{5}{*}{muy} & 1k & 0.034 & 0.026 & 3.333 & 4.746 & 0.234 & 0.268 \\
 & 2k & 0.034 & 0.026 & 3.667 & 4.765 & 0.235 & 0.268 \\
 & 3k & 0.034 & 0.026 & 3.667 & 4.819 & 0.235 & 0.268 \\
 & 4k & 0.034 & 0.026 & 3.667 & 4.817 & 0.235 & 0.268 \\
 & 5k & 0.034 & 0.026 & 3.667 & 4.824 & 0.234 & 0.268 \\
\hdashline
\multirow[c]{5}{*}{niy} & 1k & 0.254 & 0.201 & 4.000 & 42.593 & 1.045 & 1.056 \\
 & 2k & 0.253 & 0.200 & 4.000 & 42.478 & 1.043 & 1.055 \\
 & 3k & 0.253 & 0.200 & 4.000 & 42.504 & 1.043 & 1.055 \\
 & 4k & 0.253 & 0.200 & 4.000 & 42.470 & 1.043 & 1.055 \\
 & 5k & 0.253 & 0.199 & 4.000 & 42.235 & 1.042 & 1.055 \\
\hdashline
\multirow[c]{5}{*}{xed} & 1k & 0.011 & 0.008 & 2.000 & 1.280 & 0.086 & 0.137 \\
 & 2k & 0.011 & 0.008 & 2.000 & 1.292 & 0.087 & 0.139 \\
 & 3k & 0.011 & 0.008 & 2.000 & 1.304 & 0.088 & 0.139 \\
 & 4k & 0.011 & 0.008 & 2.000 & 1.294 & 0.089 & 0.139 \\
 & 5k & 0.011 & 0.008 & 2.000 & 1.298 & 0.089 & 0.139 \\
\hdashline
\multirow[c]{3}{*}{anv} & 1k & 0.148 & 0.117 & 2.000 & 18.907 & 0.472 & 0.496 \\
 & 2k & 0.147 & 0.116 & 2.000 & 18.594 & 0.376 & 0.435 \\
 & 3k & 0.147 & 0.116 & 2.000 & 18.642 & 0.342 & 0.433 \\
 \multirow[c]{2}{*}{anv}& 4k & 0.147 & 0.116 & 2.000 & 18.647 & 0.342 & 0.433 \\
 & 5k & 0.147 & 0.115 & 2.000 & 18.724 & 0.342 & 0.434 \\
\hdashline
\multirow[c]{5}{*}{lee} & 1k & 0.262 & 0.195 & 5.222 & 31.564 & 1.100 & 1.080 \\
 & 2k & 0.262 & 0.195 & 5.222 & 31.690 & 1.100 & 1.079 \\
 & 3k & 0.262 & 0.194 & 5.222 & 31.770 & 1.099 & 1.079 \\
 & 4k & 0.262 & 0.194 & 5.222 & 31.683 & 1.100 & 1.080 \\
 & 5k & 0.262 & 0.193 & 5.222 & 31.509 & 1.100 & 1.080 \\
\hdashline
\multirow[c]{5}{*}{ksf} & 1k & 0.154 & 0.119 & 2.091 & 18.205 & 0.388 & 0.499 \\
 & 2k & 0.154 & 0.119 & 2.091 & 18.283 & 0.390 & 0.500 \\
 & 3k & 0.154 & 0.119 & 2.083 & 18.353 & 0.357 & 0.478 \\
 & 4k & 0.154 & 0.119 & 2.083 & 18.316 & 0.357 & 0.478 \\
 & 5k & 0.154 & 0.119 & 2.083 & 18.301 & 0.357 & 0.478 \\
\hdashline
\multirow[c]{5}{*}{pkb} & 1k & 0.022 & 0.018 & 2.333 & 2.689 & 0.587 & 0.639 \\
 & 2k & 0.022 & 0.018 & 2.333 & 2.704 & 0.590 & 0.641 \\
 & 3k & 0.022 & 0.018 & 2.333 & 2.743 & 0.591 & 0.644 \\
 & 4k & 0.022 & 0.018 & 2.333 & 2.732 & 0.590 & 0.643 \\
 & 5k & 0.022 & 0.018 & 2.333 & 2.723 & 0.589 & 0.642 \\
\hdashline
\multirow[c]{5}{*}{nko} & 1k & 0.152 & 0.119 & 2.000 & 15.933 & 0.539 & 0.562 \\
 & 2k & 0.152 & 0.119 & 2.000 & 15.987 & 0.539 & 0.562 \\
 & 3k & 0.152 & 0.119 & 2.000 & 16.038 & 0.538 & 0.562 \\
 & 4k & 0.151 & 0.119 & 2.000 & 15.984 & 0.538 & 0.562 \\
 & 5k & 0.151 & 0.117 & 2.000 & 15.865 & 0.537 & 0.561 \\
\hdashline
\multirow[c]{5}{*}{lef} & 1k & 0.027 & 0.021 & 2.000 & 3.093 & 0.146 & 0.150 \\
 & 2k & 0.027 & 0.021 & 2.000 & 3.070 & 0.146 & 0.150 \\
 & 3k & 0.026 & 0.021 & 2.000 & 3.051 & 0.145 & 0.150 \\
 & 4k & 0.026 & 0.021 & 2.000 & 3.053 & 0.145 & 0.150 \\
 & 5k & 0.026 & 0.020 & 2.000 & 3.035 & 0.144 & 0.150 \\
\hdashline
\multirow[c]{5}{*}{nhr} & 1k & 0.159 & 0.120 & 3.833 & 20.830 & 0.729 & 0.793 \\
 & 2k & 0.159 & 0.120 & 3.833 & 20.924 & 0.732 & 0.794 \\
 & 3k & 0.159 & 0.120 & 3.833 & 20.815 & 0.730 & 0.793 \\
 & 4k & 0.159 & 0.120 & 3.833 & 20.784 & 0.731 & 0.792 \\
 & 5k & 0.158 & 0.119 & 3.833 & 20.770 & 0.731 & 0.792 \\
\hdashline
\multirow[c]{2}{*}{mgc} & 1k & 0.110 & 0.081 & 2.000 & 10.836 & 0.355 & 0.518 \\
 & 2k & 0.110 & 0.081 & 2.000 & 10.869 & 0.355 & 0.519 \\
\hdashline
\multirow[c]{5}{*}{biv} & 1k & 0.049 & 0.034 & 2.000 & 4.115 & 0.284 & 0.288 \\
 & 2k & 0.049 & 0.034 & 2.000 & 4.130 & 0.285 & 0.287 \\
 & 3k & 0.050 & 0.035 & 2.000 & 4.203 & 0.288 & 0.290 \\
 & 4k & 0.049 & 0.035 & 2.000 & 4.162 & 0.287 & 0.290 \\
 & 5k & 0.050 & 0.034 & 2.000 & 4.159 & 0.287 & 0.290 \\
\hdashline
\multirow[c]{5}{*}{maf} & 1k & 0.056 & 0.040 & 3.400 & 4.939 & 0.197 & 0.238 \\
 & 2k & 0.056 & 0.040 & 3.400 & 4.966 & 0.198 & 0.237 \\
 & 3k & 0.056 & 0.040 & 3.400 & 4.978 & 0.198 & 0.238 \\
 & 4k & 0.056 & 0.040 & 3.400 & 4.953 & 0.198 & 0.238 \\
 & 5k & 0.056 & 0.040 & 3.400 & 4.946 & 0.199 & 0.239 \\
\hdashline
\multirow[c]{5}{*}{giz} & 1k & 0.003 & 0.002 & 2.000 & 0.257 & 0.037 & 0.042 \\
 & 2k & 0.003 & 0.002 & 2.000 & 0.259 & 0.036 & 0.042 \\
 & 3k & 0.003 & 0.002 & 2.000 & 0.253 & 0.035 & 0.041 \\
 & 4k & 0.003 & 0.002 & 2.000 & 0.254 & 0.029 & 0.035 \\
 & 5k & 0.003 & 0.002 & 2.000 & 0.256 & 0.029 & 0.035 \\
\hdashline
\multirow[c]{5}{*}{tui} & 1k & 0.083 & 0.062 & 2.400 & 9.815 & 0.413 & 0.420 \\
 & 2k & 0.083 & 0.062 & 2.400 & 9.773 & 0.412 & 0.419 \\
 & 3k & 0.083 & 0.061 & 2.400 & 9.705 & 0.412 & 0.417 \\
 & 4k & 0.083 & 0.061 & 2.400 & 9.629 & 0.410 & 0.417 \\
 & 5k & 0.083 & 0.061 & 2.400 & 9.625 & 0.410 & 0.417 \\
\end{longtable}
\end{footnotesize}
\clearpage
\twocolumn
\clearpage
\onecolumn
\begin{footnotesize}
\begin{longtable}{ccrrrrrr}
\caption{Complexity metrics for diacritical system of each European language at 9 train sizes. For a given language, a metric may occasionally have identical values throughout different train sizes because they are rounded to 3 digits.} \label{tab:leveled_dia_complexity_metrics_euro} \\
\toprule
 &  & DCR & DWR & DBR & DWSR & AED & WAED \\
Lang & Size &  &  &  &  &  &  \\
\midrule
\endfirsthead
\caption[]{Complexity metrics for diacritical system of each European language at 9 train sizes. For a given language, a metric may occasionally have identical values throughout different train sizes because they are rounded to 3 digits.} \\
\toprule
 &  & DCR & DWR & DBR & DWSR & AED & WAED \\
Lang & Size &  &  &  &  &  &  \\
\midrule
\endhead
\midrule
\multicolumn{8}{r}{Continued on next page} \\
\midrule
\endfoot
\bottomrule
\endlastfoot
\multirow[c]{9}{*}{el} & 1k & 0.102 & 0.086 & 2.286 & 16.310 & 0.282 & 0.475 \\
 & 2k & 0.102 & 0.086 & 2.286 & 16.190 & 0.281 & 0.475 \\
 & 3k & 0.102 & 0.086 & 2.412 & 16.155 & 0.235 & 0.404 \\
 & 4k & 0.102 & 0.086 & 2.444 & 16.145 & 0.224 & 0.380 \\
 & 5k & 0.102 & 0.086 & 2.500 & 16.114 & 0.202 & 0.380 \\
 & 25k & 0.102 & 0.087 & 2.760 & 16.005 & 0.162 & 0.354 \\
 & 125k & 0.102 & 0.087 & 3.394 & 15.951 & 0.126 & 0.298 \\
 & 625k & 0.102 & 0.087 & 3.649 & 15.945 & 0.125 & 0.294 \\
 & 1M & 0.102 & 0.087 & 3.632 & 15.947 & 0.121 & 0.294 \\
\hdashline
\multirow[c]{7}{*}{cs} & 1k & 0.125 & 0.106 & 2.643 & 16.582 & 0.354 & 0.409 \\
 & 2k & 0.124 & 0.106 & 2.786 & 16.555 & 0.354 & 0.408 \\
 & 3k & 0.124 & 0.106 & 2.786 & 16.448 & 0.353 & 0.408 \\
 & 4k & 0.124 & 0.106 & 3.000 & 16.497 & 0.354 & 0.408 \\
 & 5k & 0.124 & 0.106 & 3.143 & 16.450 & 0.354 & 0.408 \\
 & 25k & 0.125 & 0.106 & 3.643 & 16.348 & 0.354 & 0.409 \\
 & 125k & 0.125 & 0.106 & 4.375 & 16.311 & 0.310 & 0.393 \\
\hdashline
\multirow[c]{8}{*}{da} & 1k & 0.011 & 0.009 & 2.857 & 1.314 & 0.065 & 0.077 \\
 & 2k & 0.011 & 0.009 & 3.143 & 1.328 & 0.067 & 0.078 \\
 & 3k & 0.011 & 0.009 & 3.571 & 1.333 & 0.067 & 0.078 \\
 & 4k & 0.011 & 0.009 & 3.714 & 1.335 & 0.067 & 0.078 \\
 & 5k & 0.011 & 0.009 & 3.625 & 1.327 & 0.059 & 0.066 \\
 & 25k & 0.011 & 0.009 & 4.333 & 1.317 & 0.051 & 0.058 \\
 & 125k & 0.011 & 0.009 & 4.071 & 1.304 & 0.034 & 0.043 \\
 & 625k & 0.011 & 0.009 & 3.909 & 1.308 & 0.131 & 0.039 \\
\hdashline
\multirow[c]{9}{*}{de} & 1k & 0.017 & 0.014 & 3.250 & 2.416 & 0.132 & 0.091 \\
 & 2k & 0.017 & 0.014 & 3.375 & 2.401 & 0.131 & 0.090 \\
 & 3k & 0.017 & 0.014 & 3.625 & 2.400 & 0.131 & 0.090 \\
 & 4k & 0.017 & 0.014 & 3.556 & 2.400 & 0.116 & 0.086 \\
 & 5k & 0.017 & 0.014 & 3.667 & 2.412 & 0.116 & 0.086 \\
 & 25k & 0.017 & 0.014 & 4.000 & 2.401 & 0.095 & 0.075 \\
 & 125k & 0.017 & 0.014 & 3.938 & 2.414 & 0.097 & 0.063 \\
 & 625k & 0.017 & 0.014 & 3.917 & 2.408 & 0.194 & 0.061 \\
 & 1M & 0.017 & 0.014 & 4.083 & 2.407 & 0.192 & 0.061 \\
\hdashline
\multirow[c]{9}{*}{es} & 1k & 0.022 & 0.018 & 2.750 & 3.061 & 0.123 & 0.132 \\
 & 2k & 0.022 & 0.018 & 3.250 & 3.055 & 0.123 & 0.132 \\
 & 3k & 0.022 & 0.018 & 3.500 & 3.009 & 0.122 & 0.131 \\
 & 4k & 0.022 & 0.018 & 3.500 & 3.009 & 0.122 & 0.131 \\
 & 5k & 0.022 & 0.018 & 3.625 & 3.030 & 0.123 & 0.131 \\
 & 25k & 0.022 & 0.018 & 3.727 & 3.013 & 0.090 & 0.128 \\
 & 125k & 0.022 & 0.018 & 3.938 & 2.999 & 0.090 & 0.105 \\
 & 625k & 0.022 & 0.018 & 4.389 & 3.005 & 0.072 & 0.098 \\
 & 1M & 0.022 & 0.018 & 4.227 & 3.004 & 0.105 & 0.095 \\
\hdashline
\multirow[c]{7}{*}{et} & 1k & 0.035 & 0.030 & 3.500 & 4.546 & 0.239 & 0.193 \\
 & 2k & 0.034 & 0.030 & 3.750 & 4.523 & 0.243 & 0.192 \\
 & 3k & 0.035 & 0.030 & 3.889 & 4.522 & 0.217 & 0.179 \\
 & 4k & 0.035 & 0.030 & 4.000 & 4.528 & 0.216 & 0.179 \\
 & 5k & 0.034 & 0.030 & 4.222 & 4.497 & 0.214 & 0.178 \\
 & 25k & 0.034 & 0.030 & 4.067 & 4.487 & 0.131 & 0.130 \\
 & 125k & 0.034 & 0.030 & 4.500 & 4.465 & 0.124 & 0.128 \\
\hdashline
\multirow[c]{9}{*}{fi} & 1k & 0.052 & 0.046 & 2.625 & 7.081 & 0.140 & 0.225 \\
 & 2k & 0.052 & 0.046 & 3.000 & 7.070 & 0.135 & 0.225 \\
 & 3k & 0.052 & 0.045 & 3.000 & 7.023 & 0.104 & 0.191 \\
 & 4k & 0.052 & 0.045 & 3.200 & 7.049 & 0.105 & 0.191 \\
 & 5k & 0.052 & 0.045 & 3.300 & 7.105 & 0.106 & 0.191 \\
 & 25k & 0.052 & 0.045 & 3.917 & 7.107 & 0.095 & 0.186 \\
 & 125k & 0.052 & 0.045 & 4.200 & 7.093 & 0.122 & 0.153 \\
 & 625k & 0.052 & 0.045 & 3.750 & 7.086 & 0.177 & 0.143 \\
 & 1M & 0.052 & 0.045 & 3.833 & 7.086 & 0.167 & 0.143 \\
\hdashline
\multirow[c]{2}{*}{fr} & 1k & 0.035 & 0.029 & 3.556 & 4.892 & 0.102 & 0.193 \\
 & 2k & 0.035 & 0.029 & 3.556 & 4.924 & 0.101 & 0.192 \\
 \multirow[c]{7}{*}{fr}& 3k & 0.035 & 0.029 & 3.778 & 4.948 & 0.100 & 0.192 \\
 & 4k & 0.035 & 0.029 & 4.000 & 4.937 & 0.100 & 0.192 \\
 & 5k & 0.035 & 0.029 & 3.900 & 4.957 & 0.090 & 0.175 \\
 & 25k & 0.035 & 0.029 & 3.923 & 4.941 & 0.070 & 0.158 \\
 & 125k & 0.035 & 0.029 & 4.500 & 4.903 & 0.064 & 0.148 \\
 & 625k & 0.035 & 0.029 & 4.263 & 4.905 & 0.098 & 0.141 \\
 & 1M & 0.035 & 0.029 & 4.474 & 4.905 & 0.093 & 0.141 \\
\hdashline
\multirow[c]{7}{*}{hu} & 1k & 0.109 & 0.094 & 2.800 & 15.589 & 0.330 & 0.424 \\
 & 2k & 0.109 & 0.094 & 3.300 & 15.703 & 0.330 & 0.425 \\
 & 3k & 0.108 & 0.094 & 3.400 & 15.618 & 0.330 & 0.424 \\
 & 4k & 0.108 & 0.093 & 3.500 & 15.564 & 0.329 & 0.424 \\
 & 5k & 0.108 & 0.094 & 3.455 & 15.607 & 0.299 & 0.400 \\
 & 25k & 0.108 & 0.093 & 4.083 & 15.689 & 0.274 & 0.370 \\
 & 125k & 0.108 & 0.093 & 3.789 & 15.635 & 0.262 & 0.327 \\
\hdashline
\multirow[c]{9}{*}{it} & 1k & 0.007 & 0.006 & 3.286 & 1.027 & 0.060 & 0.062 \\
 & 2k & 0.007 & 0.006 & 3.250 & 1.045 & 0.053 & 0.060 \\
 & 3k & 0.007 & 0.006 & 3.625 & 1.034 & 0.052 & 0.059 \\
 & 4k & 0.007 & 0.006 & 4.000 & 1.033 & 0.053 & 0.060 \\
 & 5k & 0.007 & 0.006 & 4.000 & 1.016 & 0.052 & 0.059 \\
 & 25k & 0.007 & 0.006 & 4.300 & 1.015 & 0.042 & 0.052 \\
 & 125k & 0.007 & 0.006 & 4.571 & 1.020 & 0.030 & 0.044 \\
 & 625k & 0.007 & 0.006 & 5.071 & 1.023 & 0.031 & 0.044 \\
 & 1M & 0.007 & 0.006 & 4.750 & 1.023 & 0.043 & 0.044 \\
\hdashline
\multirow[c]{7}{*}{lt} & 1k & 0.068 & 0.058 & 3.200 & 8.618 & 0.327 & 0.307 \\
 & 2k & 0.068 & 0.058 & 3.091 & 8.590 & 0.297 & 0.286 \\
 & 3k & 0.067 & 0.058 & 3.182 & 8.589 & 0.297 & 0.286 \\
 & 4k & 0.068 & 0.058 & 3.273 & 8.634 & 0.298 & 0.287 \\
 & 5k & 0.068 & 0.058 & 3.273 & 8.663 & 0.298 & 0.287 \\
 & 25k & 0.068 & 0.058 & 3.857 & 8.641 & 0.234 & 0.243 \\
 & 125k & 0.067 & 0.058 & 3.889 & 8.635 & 0.266 & 0.235 \\
\hdashline
\multirow[c]{7}{*}{lv} & 1k & 0.104 & 0.089 & 3.214 & 13.917 & 0.264 & 0.355 \\
 & 2k & 0.103 & 0.088 & 3.214 & 13.830 & 0.262 & 0.354 \\
 & 3k & 0.103 & 0.088 & 3.200 & 13.790 & 0.244 & 0.333 \\
 & 4k & 0.103 & 0.088 & 3.333 & 13.814 & 0.242 & 0.333 \\
 & 5k & 0.103 & 0.088 & 3.333 & 13.795 & 0.241 & 0.333 \\
 & 25k & 0.103 & 0.088 & 3.933 & 13.779 & 0.238 & 0.333 \\
 & 125k & 0.103 & 0.089 & 4.312 & 13.808 & 0.252 & 0.333 \\
\hdashline
\multirow[c]{9}{*}{nl} & 1k & 0.001 & 0.001 & 2.769 & 0.173 & 0.103 & 0.015 \\
 & 2k & 0.001 & 0.001 & 2.786 & 0.173 & 0.094 & 0.014 \\
 & 3k & 0.001 & 0.001 & 2.857 & 0.171 & 0.093 & 0.013 \\
 & 4k & 0.001 & 0.001 & 2.857 & 0.171 & 0.093 & 0.013 \\
 & 5k & 0.001 & 0.001 & 2.929 & 0.173 & 0.092 & 0.013 \\
 & 25k & 0.001 & 0.001 & 3.235 & 0.180 & 0.075 & 0.012 \\
 & 125k & 0.001 & 0.001 & 3.944 & 0.176 & 0.071 & 0.011 \\
 & 625k & 0.001 & 0.001 & 3.917 & 0.178 & 0.139 & 0.010 \\
 & 1M & 0.001 & 0.001 & 4.000 & 0.177 & 0.132 & 0.010 \\
\hdashline
\multirow[c]{7}{*}{pl} & 1k & 0.051 & 0.044 & 3.200 & 6.775 & 0.224 & 0.263 \\
 & 2k & 0.051 & 0.044 & 3.500 & 6.778 & 0.224 & 0.263 \\
 & 3k & 0.051 & 0.044 & 4.000 & 6.739 & 0.224 & 0.263 \\
 & 4k & 0.051 & 0.044 & 4.000 & 6.803 & 0.224 & 0.264 \\
 & 5k & 0.051 & 0.044 & 4.000 & 6.829 & 0.224 & 0.264 \\
 & 25k & 0.051 & 0.044 & 4.000 & 6.920 & 0.196 & 0.222 \\
 & 125k & 0.051 & 0.044 & 3.824 & 6.918 & 0.186 & 0.209 \\
\hdashline
\multirow[c]{8}{*}{pt} & 1k & 0.040 & 0.033 & 4.000 & 5.509 & 0.233 & 0.252 \\
 & 2k & 0.040 & 0.033 & 3.556 & 5.541 & 0.182 & 0.233 \\
 & 3k & 0.040 & 0.033 & 3.500 & 5.560 & 0.164 & 0.216 \\
 & 4k & 0.040 & 0.033 & 3.700 & 5.565 & 0.164 & 0.216 \\
 & 5k & 0.040 & 0.033 & 3.700 & 5.589 & 0.164 & 0.217 \\
 & 25k & 0.040 & 0.033 & 3.769 & 5.575 & 0.128 & 0.207 \\
 & 125k & 0.040 & 0.033 & 4.357 & 5.584 & 0.119 & 0.191 \\
 & 625k & 0.040 & 0.033 & 4.222 & 5.580 & 0.099 & 0.172 \\
\multirow[c]{1}{*}{pt} & 1M & 0.040 & 0.033 & 4.579 & 5.580 & 0.093 & 0.172 \\
\hdashline
\multirow[c]{7}{*}{ro} & 1k & 0.061 & 0.051 & 3.333 & 8.793 & 0.212 & 0.260 \\
 & 2k & 0.061 & 0.051 & 3.556 & 8.778 & 0.213 & 0.260 \\
 & 3k & 0.061 & 0.051 & 3.889 & 8.768 & 0.212 & 0.260 \\
 & 4k & 0.061 & 0.051 & 3.800 & 8.767 & 0.192 & 0.258 \\
 & 5k & 0.061 & 0.051 & 3.800 & 8.781 & 0.192 & 0.258 \\
 & 25k & 0.062 & 0.052 & 3.688 & 8.710 & 0.261 & 0.256 \\
 & 125k & 0.061 & 0.051 & 3.737 & 8.723 & 0.214 & 0.221 \\
\hdashline
\multirow[c]{7}{*}{sk} & 1k & 0.102 & 0.087 & 2.857 & 14.268 & 0.365 & 0.358 \\
 & 2k & 0.103 & 0.087 & 2.857 & 14.417 & 0.365 & 0.359 \\
 & 3k & 0.103 & 0.087 & 3.143 & 14.355 & 0.366 & 0.359 \\
 & 4k & 0.103 & 0.087 & 3.286 & 14.394 & 0.366 & 0.359 \\
 & 5k & 0.103 & 0.087 & 3.357 & 14.443 & 0.366 & 0.359 \\
 & 25k & 0.102 & 0.087 & 3.800 & 14.407 & 0.341 & 0.357 \\
 & 125k & 0.102 & 0.087 & 4.333 & 14.388 & 0.341 & 0.357 \\
\hdashline
\multirow[c]{6}{*}{sl} & 1k & 0.035 & 0.029 & 2.500 & 4.095 & 0.202 & 0.140 \\
 & 2k & 0.035 & 0.029 & 2.556 & 4.069 & 0.180 & 0.135 \\
 & 3k & 0.035 & 0.029 & 2.778 & 4.082 & 0.179 & 0.134 \\
 & 4k & 0.035 & 0.029 & 2.909 & 4.075 & 0.162 & 0.117 \\
 & 5k & 0.035 & 0.029 & 3.000 & 4.056 & 0.160 & 0.117 \\
 & 25k & 0.035 & 0.029 & 3.643 & 4.092 & 0.124 & 0.100 \\
\hdashline
\multirow[c]{9}{*}{sv} & 1k & 0.051 & 0.043 & 3.333 & 6.550 & 0.204 & 0.321 \\
 & 2k & 0.051 & 0.043 & 3.667 & 6.566 & 0.205 & 0.321 \\
 & 3k & 0.051 & 0.043 & 3.667 & 6.588 & 0.204 & 0.321 \\
 & 4k & 0.051 & 0.043 & 3.571 & 6.590 & 0.175 & 0.313 \\
 & 5k & 0.051 & 0.043 & 3.500 & 6.615 & 0.154 & 0.267 \\
 & 25k & 0.051 & 0.043 & 4.000 & 6.650 & 0.097 & 0.195 \\
 & 125k & 0.051 & 0.043 & 3.895 & 6.680 & 0.198 & 0.183 \\
 & 625k & 0.051 & 0.043 & 3.957 & 6.679 & 0.206 & 0.169 \\
 & 1M & 0.051 & 0.043 & 4.000 & 6.682 & 0.196 & 0.169 \\
\end{longtable}
\end{footnotesize}
\clearpage
\twocolumn

\end{document}